\documentclass{article}

\usepackage[final]{neurips_2025}

\usepackage{graphicx}
\usepackage{subcaption}
\usepackage{booktabs} % for professional tables
\usepackage{algorithm}
\usepackage{wrapfig}
\usepackage{algpseudocode}
\usepackage{tabularray}

\usepackage{hyperref}

\usepackage[utf8]{inputenc} % allow utf-8 input
\usepackage[T1]{fontenc}    % use 8-bit T1 fonts
\usepackage{hyperref}       % hyperlinks
\usepackage{url}            % simple URL typesetting
\usepackage{booktabs}       % professional-quality tables
\usepackage{amsfonts}       % blackboard math symbols
\usepackage{nicefrac}       % compact symbols for 1/2, etc.
\usepackage{microtype}      % microtypography
\usepackage{xcolor}         % colors
\usepackage{amsmath}
\usepackage{amssymb}
\usepackage{mathtools}
\usepackage{amsthm}
\usepackage{bbm}
\usepackage[capitalize,noabbrev]{cleveref}
\usepackage{float}

\theoremstyle{plain}
\newtheorem{theorem}{Theorem}[section]
\newtheorem{proposition}[theorem]{Proposition}
\newtheorem{lemma}[theorem]{Lemma}

\theoremstyle{definition}

\theoremstyle{remark}
\newtheorem{remark}[theorem]{Remark}

\DeclareMathOperator*{\argmax}{arg\,max}
\DeclareMathOperator*{\argmin}{arg\,min}

\title{Adversarial Diffusion \\
for Robust Reinforcement Learning}

\author{%
  Daniele Foffano\thanks{Corresponding author} \\
  Division of Decision and Control Systems\\
  KTH, Royal Institute of Technology\\
  \texttt{foffano@kth.se} \\
  \And
  Alessio Russo \\
  Faculty of Computing and Data Sciences \\
  Boston University \\
  \texttt{arusso2@bu.edu} \\
  \AND
  Alexandre Proutiere \\
  Division of Decision and Control Systems \\
  KTH, Royal Institute of Technology \\
  \texttt{alepro@kth.se} \\
}

\begin{document}

\maketitle

\begin{abstract}

%We investigate the problem of efficiently training robust reinforcement learning (RL) policies by leveraging diffusion models. These models are able to learn the underlying distribution of environment trajectories in an efficient way, and permit to sample entire trajectories “all at once”,  removing the necessity of simulating one state at a time. By leveraging this property, one can guide the diffusion process, allowing to sample from the $\alpha$-quantile of the most adverse trajectories in terms of value. In turn, these trajectories  can be used for training and robustifying a policy, effectively optimizing the Conditional Value at Risk (CVaR) of the total reward collected from an environment. This method stands in contrast to prior works in robust RL, which have mainly focused on costly zero-sum game training procedures or on other model-free methods that can be difficult to train.

Robustness to modeling errors and uncertainties remains a central challenge in reinforcement learning (RL). In this work, we address this challenge by leveraging diffusion models to train robust RL policies. Diffusion models have recently gained popularity in model-based RL due to their ability to generate full trajectories "all at once", mitigating the compounding errors typical of step-by-step transition models. Moreover, they can be conditioned to sample from specific distributions, making them highly flexible. We leverage conditional sampling to learn policies that are robust  to uncertainty in environment dynamics. Building on the established connection between Conditional Value at Risk (CVaR) optimization and robust RL, we introduce Adversarial Diffusion for Robust Reinforcement Learning ({\sc AD-RRL}). {\sc AD-RRL} guides the diffusion process to generate worst-case trajectories during training, effectively optimizing the CVaR of the cumulative return. Empirical results across standard benchmarks show that {\sc AD-RRL} achieves superior robustness and performance compared to existing robust RL methods.

\end{abstract}

\section{Introduction}

Reinforcement Learning (RL) has produced agents that surpass
human-level performance in various domains \cite{mirhoseini2021graph, vinyals2019grandmaster, silver2017mastering,
silver2016mastering, mnih2015human}. However, the same policies are notoriously sensitive to small
dynamics changes, sensor noise, or hardware mismatch, all of which can cause dramatic performance collapse. In safety–critical domains such as robotics, finance, or healthcare—where collecting new data is expensive, risky, or legally restricted—robustness to modeling errors is at least as important as maximizing nominal reward.

Model-based RL improves sample efficiency by learning a world-model and planning within it, but faces two key robustness obstacles: (i) compounding errors, which accumulate over long horizons \cite{wang2019benchmarking, clavera2018model}; and (ii) the Sim2Real gap, where controllers that succeed in simulation fail after minor real-world deviations \cite{rusu2017sim, christiano2016transfer}. Compounding error occurs in autoregressive models, where the model predicts one step ahead and then is fed its own prediction back: the state predicted at time $t$ is used to predict the state at $t{+}1$. Small one-step errors accumulate, the trajectory moves away from reality, and performance degrades. The Sim2Real gap arises because, even with a highly accurate simulator that minimizes unrealistic artifacts, simulated physics can never perfectly replicate reality. This discrepancy leads to reduced real-world performance due to inherent modeling inaccuracies. To overcome these challenges, RL algorithms should be made more robust by optimizing not only the expected return but also the performance under adverse or uncertain dynamics.

Recently, diffusion models have been proposed to mitigate compounding errors by generating entire trajectories rather than predicting one step at a time \cite{rigter2023world, janner2022planning}. While this reduces error accumulation, diffusion models remain imperfect: the trajectories they generate may deviate from real-world dynamics. As a result, transferring policies learned in simulation to the real world remains challenging. Despite recent progress, diffusion-based RL methods often struggle to maintain robustness when deployed in environments with unseen or perturbed dynamics.

Adversarial and risk-sensitive approaches have been explored to enhance robustness against model errors. These methods introduce worst-case perturbations during planning \cite{pinto2017robust, rajeswaran2016epopt}, or optimize Conditional Value at Risk (CVaR) objectives, which have been shown to improve resilience to both reward variability and model inaccuracies \cite{osogami2012robustness, chow2015risk}. In this work, we show that diffusion models and CVaR-based approaches can be seamlessly integrated to complement each other. We propose Adversarial Diffusion for Robust Reinforcement Learning (AD-RRL), a novel algorithm that combines the strengths of diffusion models and CVaR-based robustness. By leveraging trajectory-level generation to mitigate compounding errors and incorporating risk-aware objectives, AD-RRL enhances the adaptability and robustness of RL agents to modeling mismatches and environmental uncertainty. More precisely, we make the following contributions.

(a) We present Adversarial Diffusion (AD), a guided diffusion model that for a given policy, generates trajectories that are challenging for the agent and result in relatively low rewards. These trajectories are either rare in the current environment or originate from unexplored regions of the domain. We show that by learning from such adversarial scenarios, the agent can improve its robustness to modeling errors. To generate these trajectories, we leverage the CVaR framework, applied to trajectory rewards, and demonstrate how guided diffusion can be used to efficiently implement this objective. This mechanism forms the foundation of AD.

(b) Building on this, we introduce AD-RRL, our RL algorithm that integrates AD within the Advantage Actor-Critic (A2C) framework. AD-RRL significantly enhances the agent’s adaptability and robustness. We empirically evaluate AD-RRL across multiple environments from the Gym/MuJoCo suite, showing that it achieves superior robustness to modeling errors. In transfer scenarios involving variations in physics parameters, AD-RRL consistently outperforms state-of-the-art baselines. 

\begin{figure}
    \centering
    \includegraphics[width=\textwidth]{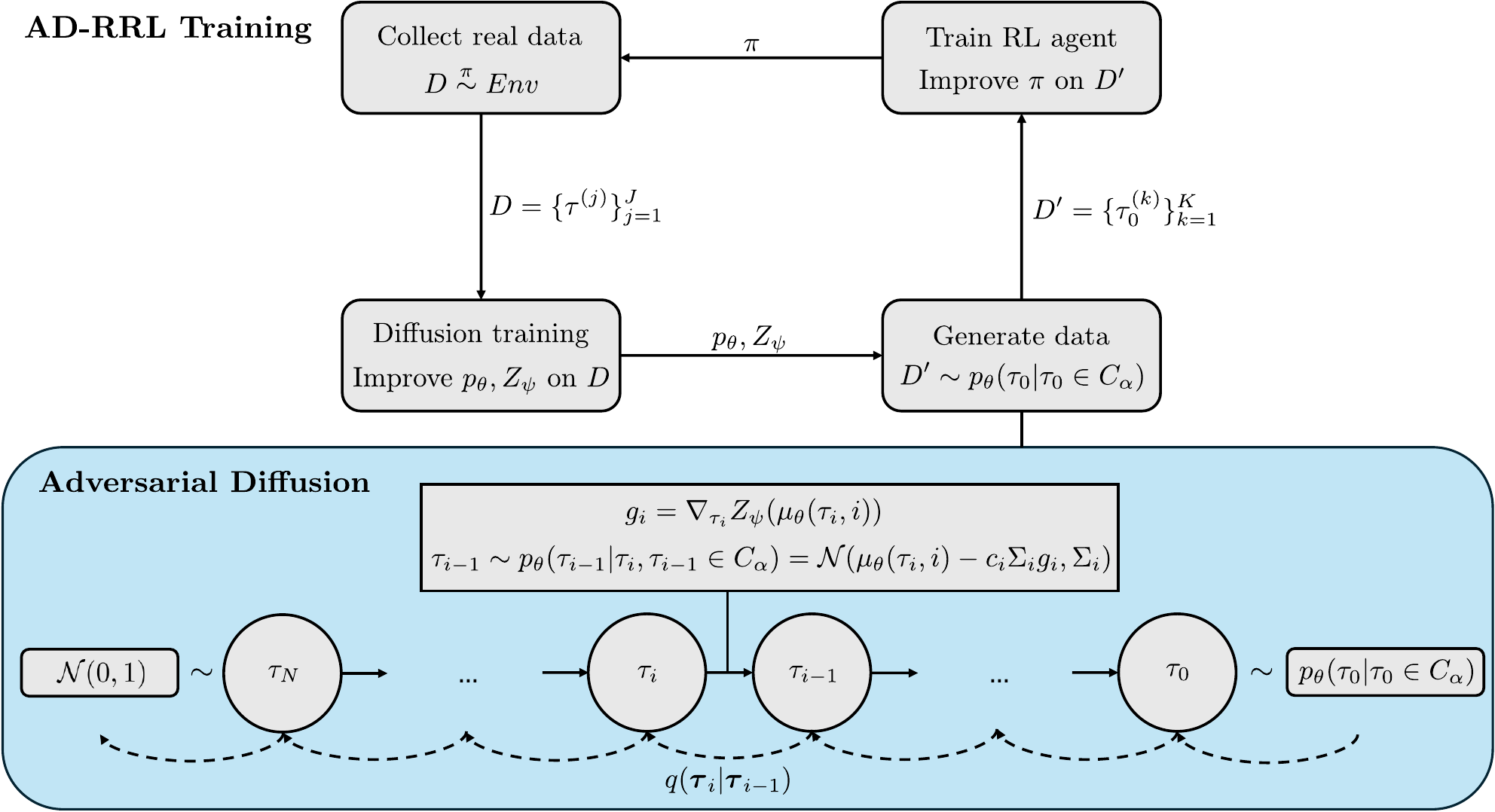}
    \caption{A high-level overview of AD-RRL. Following a Dyna-like structure \cite{sutton1991dyna}, the algorithm iteratively: samples trajectories from the real environment with a given policy $\pi$, improves the Diffusion models on the collected data, uses an Adversarially Guided Diffusion model (Section \ref{sec:adversarially_guided_diffusion}) to generate challenging synthetic trajectories which in turn are used to improve $\pi$. The loop is repeated until convergence to the optimal policy $\pi^\star$.}
    \label{fig:diagram}
\end{figure}

\section{Related Work}

\paragraph{Model-Based Reinforcement Learning.} In Model-Based Reinforcement Learning, the agent uses a model to generate new data, through which it is possible to plan further without interacting with the environment. This is essential for settings where collecting new data is impossible, illegal or dangerous. The parametric approach has received a lot of attention thanks to the constant improvements of function approximators, such as Deep Neural Networks \cite{nagabandi2018neural, chua2018deep, kaiser2019model, jafferjee2020hallucinating, van2020plannable}. Recently, Variational Auto Encoders \cite{kingma2013auto} and Transformers \cite{vaswani2017attention} have seen many successful applications as powerful models for environment dynamics \cite{micheli2022transformers, robine2023transformer, schubert2023generalist, hafner2020mastering, ha2018recurrent}, leading to state-of-the-art methods in terms of sample efficiency and performance \cite{hafner2023mastering}. These methods rely on bootstrapping to generate trajectory samples. The state prediction generated by the model is fed again as input to the model to predict the next state. As a result, these methods introduce two sources of error: one coming from an imperfect model and one from the input of the model always being wrong, except for the first timestep. The sum of these errors is commonly known as the Compounding Error problem of Model-Based methods. Multi-step prediction solutions have been proven effective even before the introduction of Diffusion models, for example by learning $H$ models to look $H$ steps into the future \cite{asadi2019combating}. It goes without saying that this approach results in a much higher learning complexity with respect to single-model approaches. Only recently, with Diffusion models becoming more popular, we have seen the rise of more efficient multi-step Model-Based RL methods \cite{rigter2023world, janner2022planning}.

\paragraph{Diffusion Models in Reinforcement Learning.}
Diffusion models are inspired by non-equilibrium thermodynamics \cite{sohl2015deep, ho2020denoising}, defining data generation as an iterative denoising process. Beyond being powerful function approximators, they also offer a natural way to condition the data generation process on labels \cite{dhariwal2021diffusion}. Recently, diffusion models have gained significant attention in the RL community. They have been used to model system dynamics, generating trajectory segments by predicting either states \cite{ajay2022conditional, zhu2023madiff}, actions \cite{chi2023diffusion, li2024crossway}, or both \cite{janner2022planning, liang2023adaptdiffuser}. Guidance techniques can further refine trajectory generation by conditioning the process on value estimates, promoting high-expected-reward sequences. Additionally, diffusion models have been employed for policy modeling \cite{wang2019benchmarking, hansen2023idql} and value function approximation \cite{mazoure2023value}. Most research on diffusion models in RL has focused on the offline setting. In this paper, we shift the focus to the online case, building on PolyGRAD—an online, Dyna-style Model-Based RL method that uses diffusion for modeling dynamics \cite{rigter2023world}. While prior work primarily uses conditioning to generate high-reward trajectories, we take the opposite approach. Our goal is to generate challenging trajectories—those that are underexplored or unlikely—so the agent can learn a more robust policy, better suited to handling changes in dynamics and modeling errors.

\paragraph{Robust Reinforcement Learning.} Classical RL can struggle to generalize when test environments deviate from training due to model errors or shifts. Robust RL addresses this by accounting for uncertainty in actions, states, and dynamics. One line of work regularizes transition probabilities within a defined uncertainty set \cite{derman2021twice, kumar2022efficient, liu2022distributionally}. This kind of methods, despite being theoretically sound and robust, do not scale well to more complex environment.

A well-known approach to tackling complex robust RL problems while maintaining theoretical guarantees is to frame the optimization problem as a two-player game \cite{morimoto2005robust}. In this framework, two players are trained iteratively to solve a maximin optimization problem: the primary agent aims to maximize the expected cumulative reward, while an adversarial agent attempts to minimize it by introducing disturbances. For instance, in Robust Adversarial Reinforcement Learning (RARL) \cite{pinto2017robust}, the adversary applies external forces to disturb the environment's dynamics. Max-min TD3 (M2TD3) \cite{tanabe2022max} follows a similar strategy, solving a maximin problem to maximize the expected reward under worst-case scenarios within an uncertainty set. In Noisy Action Robust MDPs \cite{tessler2019action}, the adversary perturbs the agent’s actions, while in State Adversarial MDPs \cite{stanton2021robust}, the adversary introduces perturbations to the state, resulting in a Partially Observable MDP formulation. 

Several Robust RL algorithms use CVaR to constrain their optimization problems \cite{ying2022towards}. For instance, CVaR-PPO is an extension of PPO \cite{schulman2017proximal} solving a risk-sensitive constrained optimization problem that constrains the CVaR to a given threshold.

Finally, we have algorithms using Domain Randomization (DR) \cite{tobin2017domain}, where the agent maximizes the expected return on average, over a predefined uncertainty set for some given environment parameters. These classes of methods have been proven very effective in domains such as robotics \cite{li2024reinforcement}. However, they do not aim to be robust to the worst-case scenarios, and might fall short when tested on environments outside of their training distribution. 

\section{Background and Problem Statement}

\subsection{Markov Decision Processes and Reinforcement Learning}
Consider a Markov Decision Process (MDP) $M = \langle S, A, P, r, \gamma \rangle$, where $S$ and $A$ are the state and action spaces, respectively, $P(\cdot |s,a)$ is the transition probability function, $r(\cdot|s,a)$ is the reward function and $\gamma$ is the discount factor. By interacting with the MDP, a Reinforcement Learning agent is able to collect sequences of states, actions and rewards, forming trajectories $\boldsymbol \tau = (s_0, a_0, r_0, \dots, s_H, a_H, r_H)$. The objective of the Reinforcement Learning agent is to learn an optimal policy $\pi^\star$ maximizing the policy value $v_\pi(s) = \mathbb{E}_{\pi}[\sum_{i=0}^{\infty}\gamma^ir_{t+i+1}|s_t = s]$. In this paper, we consider a model-based RL setting, where we use a diffusion model to approximate the distribution of trajectories under a given policy. Specifically, if $p^\pi$ denotes the true distribution of the trajectories $\boldsymbol{\tau}$ under policy $\pi$, the diffusion model samples trajectories with distribution $p_{\boldsymbol{\theta}}$ close to $p^\pi$. We adopt a Dyna-style approach \cite{sutton1991dyna}, where the diffusion model and the policy are iteratively updated: the policy is improved using data collected from the model, while the model is improved using samples gathered from the target environment using the learned policy.

\subsection{Robust RL through the Conditional Value at Risk.}
We now discuss Conditional Value at Risk (CVaR) and its connection to Robust RL.
\paragraph{Conditional Value at Risk.} When learning policies robust to modeling errors, a framework commonly used is the one of Conditional Value at Risk. We define the return of a trajectory $\boldsymbol \tau$ by $Z(\boldsymbol \tau)= \sum_{t=0}^H \gamma^t r_t$, where $r_t$ is the reward obtained at time $t$ in this trajectory\footnote{To avoid cluttering, we write $Z$ instead of $Z(\boldsymbol \tau)$ unless it is required to avoid misunderstandings.}. Under a policy $\pi$, $Z(\boldsymbol \tau)$ is a random variable with cdf $F$. The Value-at-Risk (VaR) of $Z$ at confidence level $\alpha \in (0,1)$ corresponds to its $\alpha$ quantile:
\begin{align}
    \text{VaR}_\alpha^\pi(Z) = \max\{z|F(z) \leq \alpha\}.
\end{align}
The Conditional Value-at-Risk (CVaR) of $Z$ is then defined as the expected value of $Z$ on the lower $\alpha$-portion of its distribution
\begin{align}
    \text{CVaR}^\pi_{\alpha} (Z)= \mathbb{E}_{\pi}[Z | Z \leq \text{VaR}_\alpha^\pi(Z)].
    \label{eq:cvar}
\end{align}

\paragraph{CVaR dual formulation and its connection to robustness to modeling errors.} An alternative way of defining CVaR stems from its dual formulation \cite{artzner1999coherent, rockafellar2007coherent}:
\begin{align}
    \text{CVaR}^\pi_{\alpha} (Z) = \min_{\xi \in \mathcal{U}_{C\text{VaR}}^{\alpha, \pi}} \mathbb{E}_{\boldsymbol{\tau}\sim p^\pi}[\xi(\boldsymbol\tau) Z(\boldsymbol\tau)],
    \label{eq:dual_cvar}
\end{align}
where $\xi$ acts as a perturbation of the return $Z$. This perturbation belongs to the set $\mathcal{U}_{\text{CVaR}}^{\alpha, \pi}$, called the {\it risk envelope} and  defined as 
\begin{align}
    \mathcal{U}_{\text{CVaR}}^{\alpha,\pi} \coloneqq \left\{ \xi : \forall \boldsymbol\tau, \xi(\boldsymbol \tau) \in \left[0, \frac{1}{\alpha}\right], \mathbb{E}_{\boldsymbol{\tau}\sim p^\pi}[\xi(\boldsymbol \tau)] = 1\; \right\}.
    \label{eq:risk_envelope}
\end{align} 
(\ref{eq:dual_cvar}) states that the {\rm CVaR} of $Z$ can be defined as its expected value under a worst-case perturbed distribution. 

In RL, optimizing a CVaR objective introduces robustness to model misspecification. This is exactly because of the dual form of CVaR, where the trajectory distribution is distorted by an adversarial density $\xi(\boldsymbol\tau)$. CVaR optimization in this case equals maximizing the worst-case discounted reward when adversarial perturbations are budgeted over the whole trajectory rather than at each time step \cite{chow2015risk}. The connection between CVaR and robustness to modeling errors is well established in the RL field \cite{osogami2012robustness, rajeswaran2016epopt, pinto2017robust}, and the dual formulation is at the core of our method, as explained in Section \ref{sec:adversarially_guided_diffusion} and Appendix \ref{appendix:adv_pert}.

\subsection{Diffusion Models}\label{subsec:diff}

In this work, we adopt a model-based approach to learn robust policies. To achieve this, we harness the efficiency of diffusion processes to learn a parameterized model $p_{\boldsymbol \theta}$ of the trajectory distribution. This model allows us to sample trajectories $\boldsymbol{\tau}$ as if they were generated by the true MDP, enabling policy training on these synthetic trajectories.

Diffusion models generate data by progressively refining noisy inputs through an iterative denoising process, $p_{\boldsymbol{\theta}}(\boldsymbol{\tau}_{i-1}| \boldsymbol{\tau}_i)$. This process reverses the forward diffusion, $q(\boldsymbol{\tau}_{i}| \boldsymbol{\tau}_{i-1})$, which gradually corrupts real data by adding random noise. Each step of the denoising process is typically parameterized as a Gaussian distribution
\begin{align}\label{eq:gausdiff}
    p_{\boldsymbol{\theta}}(\boldsymbol{\tau}_{i-1} \vert \boldsymbol{\tau}_i) = \mathcal{N}(\mu_{\boldsymbol \theta}(\boldsymbol \tau_i, i), \boldsymbol \Sigma_i),
\end{align}
with learned mean and fixed covariance matrices, both depending on the diffusion step $i$.

The denoising process is formulated as 
\begin{align}
    p_{\boldsymbol{\theta}}(\boldsymbol{\tau}_{0:N}) = p(\boldsymbol{\tau}_N) \prod^N_{i=1} p_{\boldsymbol{\theta}}(\boldsymbol{\tau}_{i-1} \vert \boldsymbol{\tau}_i),
\end{align}
where $p(\boldsymbol \tau_N) \approx \mathcal{N}(\boldsymbol 0, \boldsymbol I)$ and $\boldsymbol \tau_0$ is the real (i.e., noiseless) trajectory. The parameters $\boldsymbol \theta$ are learned by optimizing the variational lower bound on the negative log likelihood:
\begin{align}
\label{eq:vlb}
    \boldsymbol \theta^\star = \argmin_{\boldsymbol \theta} \mathbb{E}_{\boldsymbol \tau_0}[-\log p_{\boldsymbol \theta}(\boldsymbol \tau_0)],
\end{align}
where $p_{\boldsymbol \theta}(\boldsymbol \tau_0) = \int p_{\boldsymbol{\theta}}(\boldsymbol{\tau}_{0:N})\text{d}\boldsymbol{\tau}_{1:N}$.

\paragraph{Guided diffusion.} A classifier $p(y|\boldsymbol \tau_0)$ adding information about the sample to be reconstructed (e.g., the optimality of the trajectory) can enhance the generative performance of the diffusion model \cite{dhariwal2021diffusion}
\begin{align}
\label{eq:guided_diff}
    p_{\boldsymbol\theta}(\boldsymbol\tau_0|y) \propto p_{\boldsymbol\theta}(\boldsymbol\tau_0)p(y|\boldsymbol \tau_0).
\end{align}
By leveraging the classifier’s gradient, we can guide the denoising process toward generating samples that align more closely with the classifier’s predictions. This method, called Classifier-Guided Diffusion, generates trajectory samples according to
\begin{align}
    p_{\boldsymbol{\theta}}(\boldsymbol{\tau}_{i-1}| \boldsymbol{\tau}_i, y) = \mathcal{N}(\mu_{\boldsymbol{\theta}}(\boldsymbol \tau_i,i) + \boldsymbol{\Sigma}_i \boldsymbol{g}_i, \boldsymbol{\Sigma}_i)
    \label{eq:ggg}
\end{align}
where $\boldsymbol{g}_i = \nabla_{\boldsymbol{\tau}}\log p(y|\boldsymbol{\tau})|_{\boldsymbol{\tau}=\mu_{\boldsymbol{\theta}}(\boldsymbol \tau_i,i)}$.

\subsection{Problem statement}
We consider a model-based Reinforcement Learning setting. For a given policy $\pi$, we learn a model $p_{\boldsymbol \theta}$ of the distribution of the corresponding trajectories. Our goal is to use $p_{\boldsymbol \theta}$ to improve the policy and its robustness to modeling errors. 

We formulate the problem of learning a robust policy using the following optimization problem:
\begin{align}
    \pi^\star_\alpha &= \argmax_{\pi} \text{CVaR}^\pi_{\alpha}(Z)\label{eq:cvar_obj}\\
    &=  \argmax_{\pi}\min_{\xi \in \mathcal{U}_{C\text{VaR}}^{\alpha, \pi}} \mathbb{E}_{\boldsymbol{\tau}\sim p^\pi}[\xi(\boldsymbol\tau) Z(\boldsymbol\tau)].\label{eq:dual_cvar_obj}
    %&= \argmax_{\pi} \mathbb{E}_{\pi}[Z | Z \leq \text{VaR}_\alpha^\pi(Z)]
\end{align}
(\ref{eq:cvar_obj}) describes the objective to obtain the policy that maximizes the return on the worst $\alpha$-percentile of the trajectories, in terms of cumulative return. However, directly sampling trajectories from this worst $\alpha$-percentile is challenging. In our approach, we leverage the dual definition of CVaR, solving instead the double optimization problem described in (\ref{eq:dual_cvar_obj}). The problem can be seen as a game where an adversarial agent $\xi$ is perturbing the trajectories distribution under a given policy $\pi$. We model this distribution via a diffusion model $p_{\boldsymbol{\theta}}$, which allows us to leverage guiding techniques. We introduce {\it adversarial guiding}, a method that steers the diffusion process toward sampling trajectories that minimize the expected return for the agent. Because the adversarial guide actively seeks to reduce return, the generated trajectories naturally fall within the worst $\alpha$-percentile. We formally demonstrate that the resulting adversarially guided diffusion models can be adapted to actually sample from the worst $\alpha$-percentile. We also empirically validate our approach.

\section{Adversarially Guided Diffusion Models}
\label{sec:adversarially_guided_diffusion}
In this section, we consider a fixed policy $\pi$, and for notational convenience, we drop the corresponding superscripts. We explain below how to efficiently generate adversarial trajectories, sampled from the set of trajectories $C_\alpha := \{\boldsymbol\tau : Z(\boldsymbol\tau) \leq \text{VaR}_\alpha(Z)\}$.

\paragraph{Sampling suboptimal trajectories.} To steer the diffusion process towards the set $C_\alpha$ we need to define the proper guidance classifier, as in (\ref{eq:guided_diff}). We start from the definition of $\text{CVaR}_{\alpha, p_{\boldsymbol\theta}}(Z)$ given in (\ref{eq:cvar}). The index $p_{\boldsymbol\theta}$ indicates that the trajectory $\boldsymbol{\tau}$ from which the return is computed is generated using the diffusion model $p_{\boldsymbol\theta}$. We have: 
\begin{align}
    \text{CVaR}_{\alpha, p_{\boldsymbol \theta}}(Z) &= \mathbb{E}_{\tau_0 \sim p_{\boldsymbol \theta}}[Z(\boldsymbol\tau_0) | \boldsymbol \tau_0 \in C_\alpha],\\
    %&= \mathbb{E}_{\tau_0 \sim p_{\boldsymbol \theta}}[Z(\boldsymbol\tau_0) | \bar O(\boldsymbol\tau_0) = 1],\\
    &= \int Z(\boldsymbol\tau_0) p_{\boldsymbol \theta}(\boldsymbol\tau_0|\boldsymbol\tau_0 \in C_\alpha)d\boldsymbol\tau_0,\\
    &=  \min_{\xi \in \mathcal{U}_{C\text{VaR}}^{\alpha, \pi}}\int Z(\boldsymbol\tau_0) \xi(\boldsymbol\tau_0)p_{\boldsymbol \theta}(\boldsymbol\tau_0)d\boldsymbol\tau_0,\label{eq:dual_opt_prob}
\end{align}
where the last equality follows from the dual definition of CVaR presented in (\ref{eq:dual_cvar}). To steer the generating process towards trajectories from $C_\alpha$, we can use the classifier $p_{\boldsymbol \theta}(\boldsymbol\tau_0 \in C_\alpha|\boldsymbol \tau_0)$, since $p_{\boldsymbol \theta}(\boldsymbol\tau_0|\boldsymbol\tau_0 \in C_\alpha) \propto p_{\boldsymbol \theta}(\boldsymbol\tau_0)p_{\boldsymbol \theta}(\boldsymbol\tau_0 \in C_\alpha|\boldsymbol \tau_0)$. 

Notice also that if we define $\xi^\star(\boldsymbol\tau_0)$ as the solution to the minimization problem in (\ref{eq:dual_opt_prob}), i.e., the bounded trajectory perturbation minimizing the cumulative reward under the dynamics $p_{\boldsymbol\theta}$, we have 
\begin{align*}
    \xi^\star(\boldsymbol\tau_0)p_{\boldsymbol \theta}(\boldsymbol\tau_0) = p_{\boldsymbol \theta}(\boldsymbol\tau_0|\boldsymbol\tau_0 \in C_\alpha) \propto p_{\boldsymbol \theta}(\boldsymbol\tau_0)p_{\boldsymbol \theta}(\boldsymbol\tau_0 \in C_\alpha|\boldsymbol \tau_0).
\end{align*}

In other words, weighting the distribution with the classifier $p_{\boldsymbol \theta}(\boldsymbol\tau_0 \in C_\alpha \mid \boldsymbol \tau_0)$ is equivalent, up to a proportionality constant, to weighting $p_\theta(\tau_0)$ according to an adversarial perturbation $\xi^\star$. We can hence think of applying a guided diffusion to implement this perturbation. However, the set $C_\alpha$ is not known. To address this limitation, and following the approach of \cite{levine2018reinforcement}, we introduce a smooth approximation of $p_{\boldsymbol \theta}(\boldsymbol \tau_0 \in C_\alpha \mid \boldsymbol \tau_0)$, namely $\exp (-c_0 \sum_{t = 1}^H \gamma^t r_t)$ for some constant $c_0 > 0$. This approximation is intuitively reasonable, as it biases the generation process toward trajectories with lower cumulative rewards.

In the following two subsections, we describe how this guided diffusion can be implemented and how it influences the diffusion process. We also discuss how to tune the guided diffusion to ensure that the resulting adversarial perturbation $\xi$ remains within the risk envelope $\mathcal{U}_{C\text{VaR}}^{\alpha, \pi}$ defined in (\ref{eq:risk_envelope}).

\subsection{Perturbed diffusion model}
\label{subsec:perturbed_diff_model}

We use the classifier ${p_{\boldsymbol{\theta}}}(\boldsymbol \tau_i \in C_\alpha | \boldsymbol \tau_i)$ so that the trajectories $\boldsymbol \tau_i$ generated at every step $i$ of the diffusion process belong to the set $C_\alpha$. We assume that ${p_{\boldsymbol{\theta}}}(\boldsymbol \tau_i \in C_\alpha | \boldsymbol \tau_i) \approx \exp{(-c_i\sum_{t = 1}^H \gamma^t r_t^{(i)})}$ for some value $c_i>0$ as we did for $\tau_0$. In the last approximation, the reward $r_t^{(i)}$ represents the reward collected at time $t$ in  trajectory $\boldsymbol{\tau}_i$ for the $i$-th step of the diffusion process.

As a slight extension of the guided diffusion principle presented in \cref{subsec:diff}, we establish the following result (essentially obtained by applying (\ref{eq:ggg}) with $y=\{\boldsymbol \tau_i \in C_\alpha\}$). 

\begin{lemma}
\label{lem:perturbed_f}
Assume that the denoising process is Gaussian, that is (\ref{eq:gausdiff}) holds. Assume that for all $i\in [N]$, the approximation ${p_{\boldsymbol{\theta}}}(\boldsymbol \tau_i \in C_\alpha | \boldsymbol \tau_i) = \exp{(-c_i\sum_{t = 1}^H \gamma^t r_t^{(i)})}$ holds. Then, we can sample trajectories from $p_\theta(\boldsymbol\tau_0|\boldsymbol \tau_0 \in C_\alpha)$ using diffusion steps of the form:
\begin{align}
\label{eq:adversarial_diffusion_model}
    p_{\boldsymbol{\theta}}(\boldsymbol{\tau}_{i-1}| \boldsymbol{\tau}_i, \boldsymbol \tau_{i-1} \in C_\alpha) = \mathcal{N}(\mu_{\boldsymbol{\theta}}(\boldsymbol \tau_i,i) - c_i\boldsymbol{\Sigma}_i \boldsymbol{g}_i, \boldsymbol{\Sigma}_i),
\end{align}
where $\boldsymbol{g}_i = \nabla_{\boldsymbol{\tau}} Z(\mu_{\boldsymbol{\theta}}(\boldsymbol \tau_i,i))$ for $i\in [N]$. 
\end{lemma}

The lemma is proved in  \cref{appendix:adg_diffusion} for completeness. The conditional sampling procedure induces the following perturbed model:
\begin{align}\label{eq:cond_sampling}
    \bar p_{\boldsymbol{\theta}}(\boldsymbol \tau_0) &= p_{\boldsymbol \theta}(\boldsymbol \tau_0 |\boldsymbol \tau_0 \in C_\alpha) \propto \int   \prod_{i=1}^Np_{\boldsymbol \theta}(\boldsymbol\tau_{i-1}|\boldsymbol\tau_i,\boldsymbol \tau_{i-1} \in C_\alpha) p(\boldsymbol \tau_N){\rm d}\boldsymbol\tau_{1:N}.
\end{align}
We refer to this sampling procedure as an {\it Adversarially Guided Diffusion Model}.

\subsection{Selecting $c_1,\ldots, c_N$} 

Note that the Adversarially Guided Diffusion Model depends on the constants $c_1, \ldots, c_N$, and recall that the resulting perturbation must lie within the risk envelope defined in (\ref{eq:risk_envelope}). In the following, we establish conditions on these constants to ensure this requirement is satisfied. To that end, we first show in \cref{appendix:adv_pert} that our model $\bar{p}_{\boldsymbol{\theta}}$ admits a product-form representation:

\begin{lemma}\label{lem:multiplicative_noise} The Adversarially Guided Diffusion Model can be expressed as $\bar p_{\boldsymbol{\theta}}(\boldsymbol \tau_0) = \xi(\boldsymbol \tau_0)p_{\boldsymbol{\theta}}(\boldsymbol \tau_0)$, where $\xi(\boldsymbol \tau_0) = \frac{\int \xi(\boldsymbol \tau_{0:N})p_{\boldsymbol \theta}(\boldsymbol{\tau}_{0:N})\text{d}\boldsymbol{\tau}_{1:N}}{p_{\boldsymbol\theta}(\boldsymbol \tau_0)}$ and where $  \xi(\boldsymbol \tau_{0:N}) \coloneqq\prod_{i=1}^N \xi(\boldsymbol{\tau}_{i}, \boldsymbol{\tau}_{i-1})$ with 
\begin{align}
        \xi(\boldsymbol \tau_{i}, \boldsymbol \tau_{i-1}) \coloneqq \exp{\left( -\frac{1}{2}(2c_i\boldsymbol D_i^T \boldsymbol g_i + c_i^2 \boldsymbol g_i^T \boldsymbol \Sigma \boldsymbol g_i)\right)},
\end{align}
and $\boldsymbol D_i \coloneqq (\boldsymbol \tau_{i-1} - \mu_{\boldsymbol{\theta}}(\boldsymbol \tau_i,i))$. 
 
\end{lemma}

Note that by definition (since $\bar{p}_{\boldsymbol{\theta}}$ is a distribution), we have that $\mathbb{E}_{\boldsymbol{\tau}\sim p_{\boldsymbol{\theta}}}[\xi(\boldsymbol \tau_0)]=1$. Hence, we just need to verify that $\xi(\boldsymbol{\tau}_0)\le 1/\alpha$ for all $\boldsymbol{\tau}_0$ to ensure that $\xi$ belongs to the risk envelope. We define $\rho$ such that the trajectories $\boldsymbol  \tau_i$ lie in a bounded space $C = \{ \boldsymbol \tau_i : ||\boldsymbol 
 \tau_i||_\infty \leq \rho \}$ and such that $||\mu_{\boldsymbol{\theta}}(\boldsymbol \tau_i,i)||_\infty < \rho$. In the following proposition, proved in \cref{appendix:constraint}, we provide conditions on $c_1,\ldots, c_N$ so that this holds.

\begin{proposition}
\label{prop:ci_constraint}
For all $i\in [N]$, let $\eta_i(\alpha,N)\geq 0$ such that $\prod_{i=1}^N\eta_i(\alpha, N) = \frac{1}{\alpha}$. \\
(a) When for all $i\in [N]$,
    \begin{align}
    c_i \le \min\left(\sqrt{\frac{2\log\eta_i(\alpha, N)}{\boldsymbol g_i^T \boldsymbol \Sigma_i \boldsymbol g_i}}, \frac{\rho - ||\mu_{\boldsymbol{\theta}}(\boldsymbol \tau_i,i)||_\infty}{||\boldsymbol \Sigma_i \boldsymbol g_i||_\infty}\right),
    \label{eq:ci}
\end{align}
then we have: for all $\boldsymbol \tau_0$, $\xi(\boldsymbol \tau_0)\leq 1/\alpha$. \\
(b) Let $i\in [N]$. Assume that $\boldsymbol \Sigma_i$ is a diagonal matrix with $(\boldsymbol \Sigma_i)_{jj} \in [0,1)$. Assume $\eta_i(\alpha, N) = \left(\frac{1}{\alpha}\right)^\frac{1}{N}$. Then, for $N$ large enough, (\ref{eq:ci}) holds as soon as $c_i \leq \sqrt{\frac{2\log\eta_i(\alpha, N)}{\boldsymbol g_i^T \boldsymbol \Sigma_i \boldsymbol g_i}}$.
\end{proposition}

Since our diffusion model uses a cosine noise schedule as in \cite{nichol2021improved}, we have that for all $i\in [N]$ $\boldsymbol \Sigma_i = \beta_i \boldsymbol I$ with $\beta_i \in [0,1)$, so we can set $c_i = \sqrt{\frac{2\log\eta_i(\alpha, N)}{\boldsymbol g_i^T \boldsymbol \Sigma_i \boldsymbol g_i}}$ to ensure that $\xi$ belongs to the risk envelope.

\begin{remark} Note that in our analysis, we have assumed for simplicity that the states and the actions were a one-dimensional vector, so that trajectories become Gaussian vectors. We can extend the analysis to the case where states and actions are multidimensional at the expense of considering trajectories as Gaussian matrices. Refer to \cref{appendix:matrix_dist} for details.
\end{remark}

\section{Algorithms}

We now introduce Adversarial Diffusion for Robust Reinforcement Learning (AD-RRL)\footnote{The official implementation for AD-RRL can be found on GitHub: \url{https://github.com/danielefoffano/AD-RRL}}, which alternates between model improvement and policy improvement steps (see Algorithm \ref{alg:MBRL_diff_adv}). AD-RRL leverages the adversarial conditional sampling discussed in the previous section to sample trajectories in the worst $\alpha$-percentile in terms of return. Our approach is primarily inspired by PolyGRAD \cite{rigter2023world} and Diffuser \cite{janner2022planning}.

We adopt the common assumption that the policy follows a Gaussian distribution over the action space, parameterized by $\mu_\omega(s)$ and $ \sigma_\omega(s)$. The policy is deployed in the real environment to collect new data, which is then used to train both the dynamics model $\bar{p}_{\boldsymbol \theta}$ and the cumulative reward function $Z_{\boldsymbol \phi}$. Following the standard approach in Dyna-like algorithms \cite{sutton1991dyna}, we generate synthetic trajectories using our learned models 
(\cref{alg:adversarial_sampling}). These trajectories are then used to train the policy via an on-policy Reinforcement Learning algorithm.

\begin{figure}[H]                     % ONE float that holds both algorithms
\vspace{-0.4cm}
  \centering
\begin{minipage}[t]{0.48\textwidth}
    \centering
\begin{algorithm}[H]
   \caption{Adversarial Diffusion for Robust Reinforcement Learning (AD-RRL)}
   \label{alg:MBRL_diff_adv}
\begin{algorithmic}[1]
\State {\bfseries Input:} environment, $E$; 
\State {\bfseries Initialize:} policy, $\pi_\omega$; adversarial denoising model, $\bar{p}_{\boldsymbol{\theta}}$; cumulative reward function $Z_{\boldsymbol \phi}$; data buffer, $\mathcal{D}$; training iterations $M$
\item[]
\For{$m=1, \ldots, M$}
\State Sample $ \boldsymbol{\tau}\sim E$ using $\pi_\omega$, add $\boldsymbol \tau$ to ${\cal D}$
% \STATE  $\mathcal{D}\text{.add}(\boldsymbol{\tau})$
\State  Improve $\bar{p}_{\boldsymbol{\theta}}$, $Z_{\boldsymbol \phi}$ on $\mathcal{D}$ $\hspace{0.2cm} \triangleright$ Algorithm \ref{alg:diffusion_training}
\State Sample $\{\hat{\boldsymbol{\tau}}\} \sim \bar{p}_{\boldsymbol{\theta}}$ $\hspace{0.75cm} \triangleright$ Algorithm \ref{alg:adversarial_sampling}
\State Improve $\pi_\omega$ on $\{\hat{\boldsymbol{\tau}}\}$ using RL
\EndFor
%\STATE $\hspace{0.4cm}$ Update action update scale $\delta$ as in \cite{rigter2023world}
\end{algorithmic}
\end{algorithm}
\end{minipage}\hfill
\begin{minipage}[t]{0.48\textwidth}
    \centering
  \begin{algorithm}[H]
   \caption{Adversarial Diffusion Trajectory Sampling}
   \label{alg:adversarial_sampling}
    \begin{algorithmic}[1]
      \State \textbf{Input:} adversarial denoising model
             $\bar p_{\boldsymbol\theta}$; \\reward model
             $Z_{\boldsymbol\phi}$; buffer $\mathcal D$; level $\alpha$
      \item[] 
      \State $\widehat{\boldsymbol\tau}_N \sim
             \mathcal N(\mathbf 0,\mathbf I)$
      \State $\boldsymbol s_0 \sim \mathcal D$
      \For{$i=N,\dots,1$}
        \State set $\hat{\boldsymbol s}_0\gets\boldsymbol s_0$
               in $\hat{\boldsymbol\tau}_i$
        \State $c_i = \sqrt{2\log\eta_i(\alpha, N)/\boldsymbol g_i^T \boldsymbol \Sigma_i \boldsymbol g_i}$ %$c_i$ with \eqref{eq:ci}
        \State $\hat{\boldsymbol\tau}_{i-1}\sim\mathcal N\!\bigl(
               \mu_{\boldsymbol\theta}(\hat{\boldsymbol\tau}_i,i)
               -c_i\boldsymbol\Sigma_i\boldsymbol g_i,\;
               \boldsymbol\Sigma_i\bigr)$
      \EndFor
      \State \Return $\hat{\boldsymbol\tau}_0$
    \end{algorithmic}
    \end{algorithm}
\end{minipage}
\vspace{-0.3cm}
\end{figure}
\cref{alg:diffusion_training} illustrates the training procedure for both our diffusion models and is presented in \cref{appendix:implementation_details}, alongside additional implementation details. The pseudocode provided is simplified. In reality, the diffusion model consists of a noise prediction function $\epsilon_{\boldsymbol \theta}(\hat{\boldsymbol \tau}_i, i)$ from which the mean is computed in closed form \cite{ho2020denoising}. This model is trained using the following objective function (derived from (\ref{eq:vlb}))
\begin{align*}
    \mathcal{L}(\boldsymbol\theta) = \mathbb{E}_{i, \epsilon, \boldsymbol \tau_0}[||\epsilon - \epsilon_{\boldsymbol \theta}(\boldsymbol \tau_i, i)||^2],
\end{align*}
where $i\sim {\cal U}(\{1,\dots, N\})$ is the diffusion process step, $\epsilon \sim \mathcal{N}(0, 1)$ is the target noise and $\boldsymbol \tau_i$ is the trajectory $\boldsymbol \tau_0\sim {\cal D}$ after $i$ steps of the \textit{forward} diffusion process adding noise $\epsilon$. We update $\boldsymbol \theta$ $K$ times, each time by randomly sampling the step $i$. The model $Z_{\boldsymbol \phi}$ is trained to predict the cumulative reward of the trajectory samples $\boldsymbol{\tau}_i$.

Both the adversarial diffusion model $\bar{p}_{\boldsymbol\theta}$ and the cumulative reward function $Z_{\boldsymbol\phi}$ are used to sample adversarially generated trajectories in the worst $\alpha-$percentile. At every step of the diffusion process, we perform inpainting by substituting a real starting state $\boldsymbol s_0$ into the generated noisy trajectory $\hat{\boldsymbol \tau}_i$. We then proceed to compute $c_i$ according to (\ref{eq:ci}) and the gradient $\boldsymbol g_i = \nabla_{\boldsymbol{s}} Z_\phi$. Notice that the gradient is taken with respect to $\boldsymbol s$, so we only adversarially corrupt the states of the trajectory. To ensure that the generated actions are consistent with the generated states, we use the PolyGRAD diffusion guidance method \cite{rigter2023world}, which generates a sequence of actions guided by the gradient of the policy $\pi_\omega$. 

\section{Experiments}

In this section, we empirically evaluate how robust our method is. During training, the agent interacts with a fixed instance of the environment. At test time, we alter key physics-related parameters and assess the agent’s performance against both robust and non-robust baselines. Our experiments are conducted on several optimal control tasks from the MuJoCo suite: InvertedPendulum, Reacher, Hopper, HalfCheetah, and Walker. All agents are trained in the default MuJoCo/OpenAI Gym environment (fixed physics), for $1.5$M steps. Additional results are provided in \cref{appendix:additional_results}.

\paragraph{Baseline methods.} We evaluate AD-RRL against several state-of-the-art baselines for robust reinforcement learning: 

(a) Domain Randomization (DR) \cite{tobin2017domain}, widely used in robotics \cite{li2024reinforcement, mehta2020active}, improves policy generalization by maximizing expected return over a distribution of dynamics. However, it does not explicitly account for worst-case or lower-percentile outcomes. We implement DR using PPO and refer to the resulting method as DR-PPO. 

(b) Max-Min TD3 (M2TD3) \cite{tanabe2022max} frames robustness as a minimax optimization problem, training an actor-critic model to maximize performance under the worst-case dynamics sampled from a predefined uncertainty set.

(c) CVaR-PPO (CPPO) \cite{ying2022towards} augments Proximal Policy Optimization with a CVaR constraint, leading to a policy-gradient algorithm that explicitly controls the policy's risk.

Additionally, we compare AD-RRL to other baselines in RL. 

(d) PolyGRAD \cite{rigter2023world}, a diffusion-based model that our work builds upon, generates synthetic trajectories via policy-guided diffusion and trains policies in an online model-based setting. It improves sample efficiency but lacks explicit robustness to adverse dynamics. 

(e) TRPO \cite{schulman2015trust} and PPO \cite{schulman2017proximal}, two strong model-free baselines, are also included for comparison. TRPO constrains policy updates using a KL-divergence trust region, while PPO employs a clipped surrogate objective for improved computational efficiency.

{\bf Robustness under varying physical parameters.} To verify the robustness of AD-RRL, we vary several physical parameters of the environment at test time. For Hopper and Cheetah, we vary body mass, ground friction and environment gravity. For Walker, we modify friction and mass. For Reacher, we vary all the actuators' gears (i.e., the torque produced by the actions). For InvertedPendulum, we change the cart mass, the pole mass and environment gravity.

\begin{figure}[htbp]
  \centering
  \vspace{-0.4\baselineskip}
%============= Row 1 ================
  \begin{subfigure}[b]{0.32\linewidth}
    \centering
    \includegraphics[width=\linewidth]{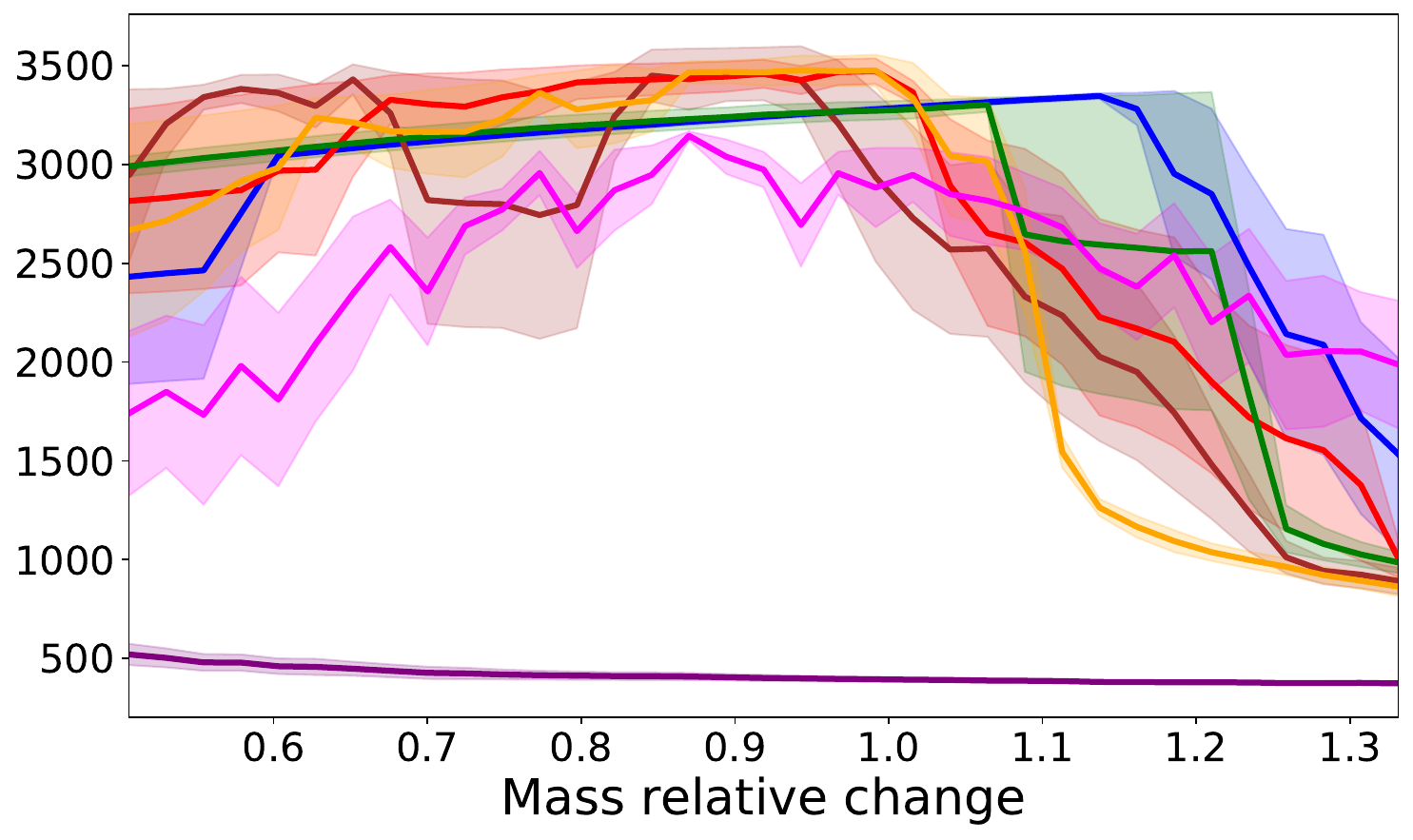}
    \subcaption{Hopper – Mass}\label{fig:Hopper_mass}
  \end{subfigure}\hfill
  \begin{subfigure}[b]{0.32\linewidth}
    \centering
    \includegraphics[width=\linewidth]{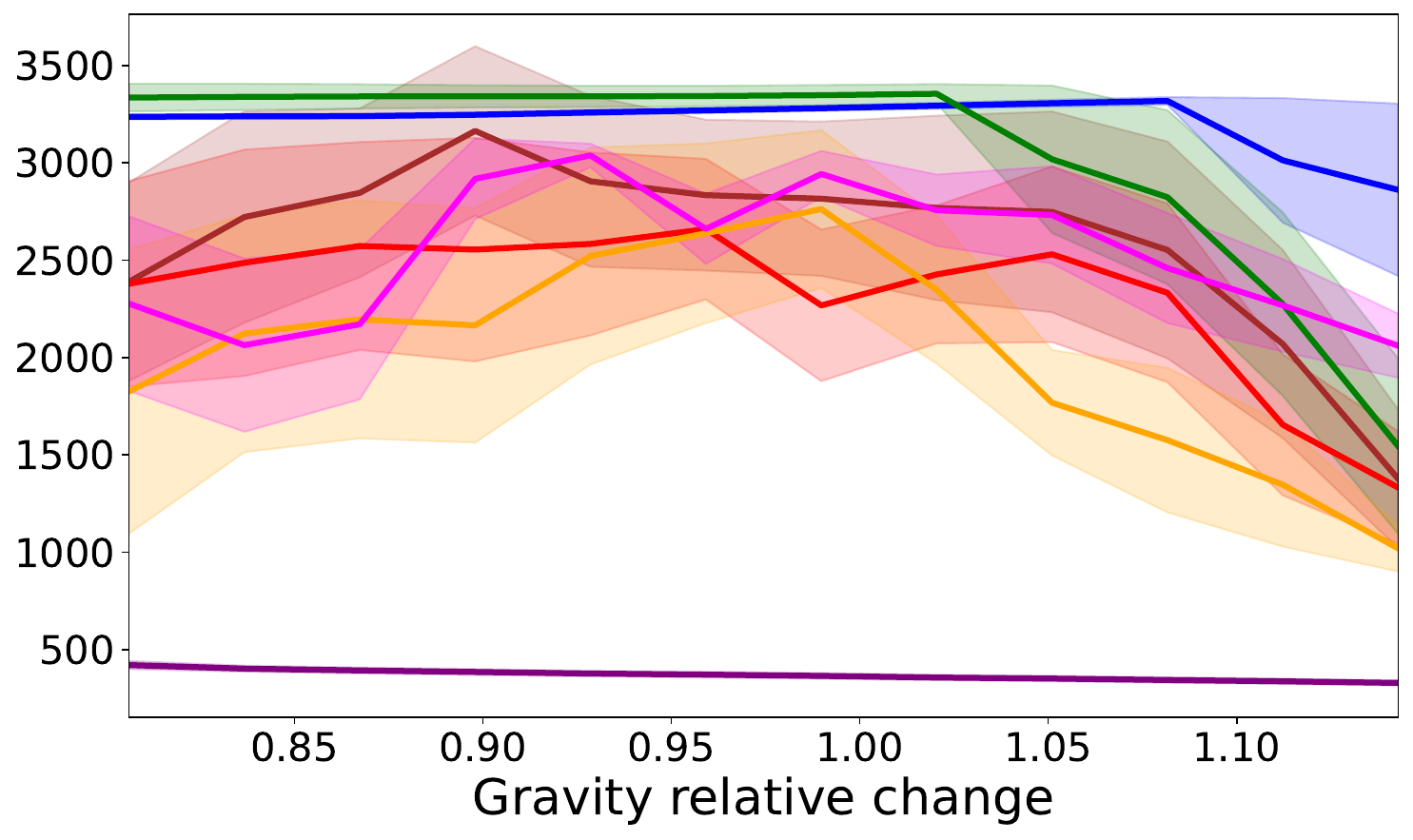}
    \subcaption{Hopper – Gravity}\label{fig:Hopper_gravity}
  \end{subfigure}\hfill
  \begin{subfigure}[b]{0.32\linewidth}
    \centering
    \includegraphics[width=\linewidth]{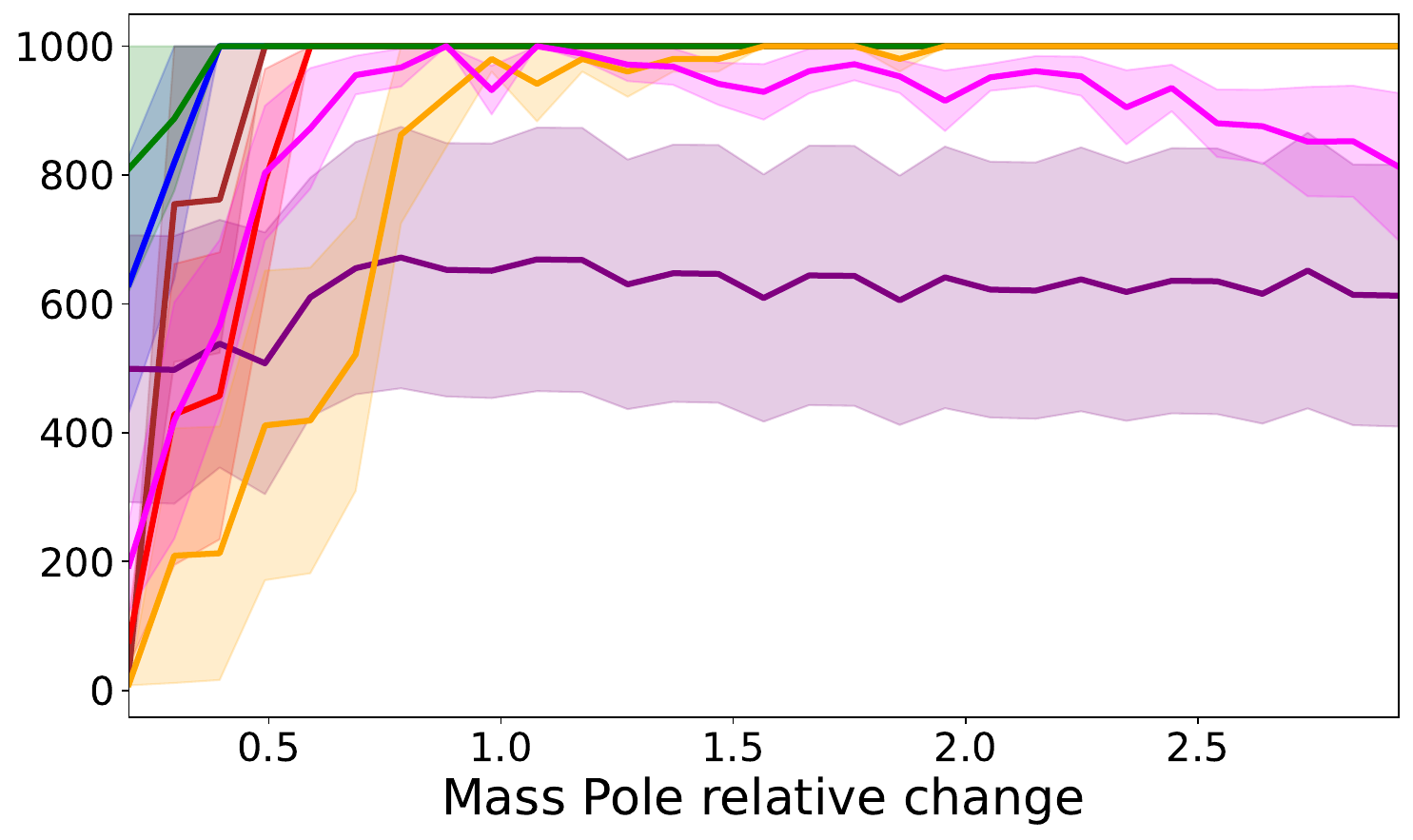}% <-- replace
    \subcaption{InvertedPendulum - Mass Pole}\label{fig:pendulum_masspole}
  \end{subfigure}

  \vspace{0.4\baselineskip}

%============= Row 2 ================
  \begin{subfigure}[b]{0.32\linewidth}
    \centering
    \includegraphics[width=\linewidth]{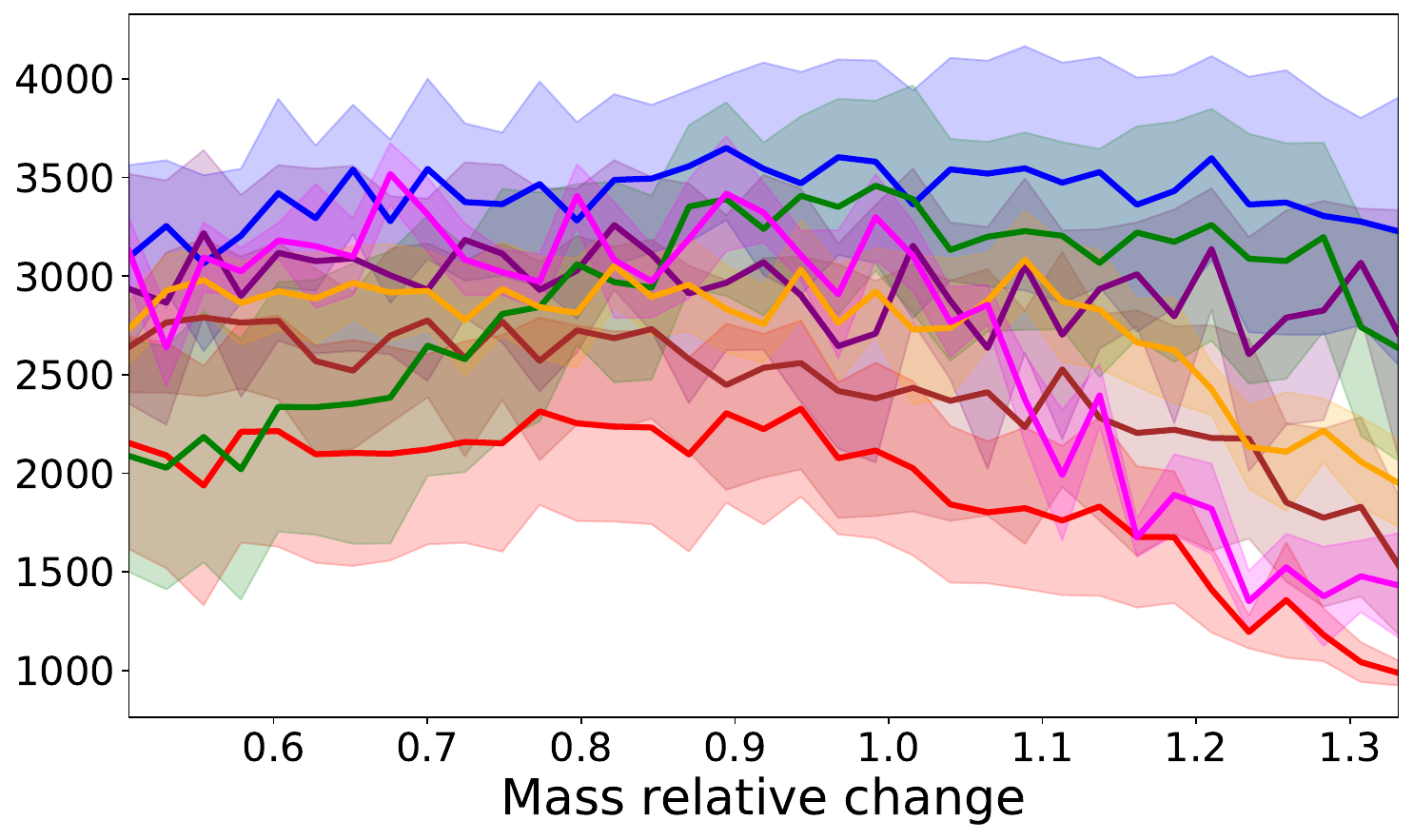}
    \subcaption{Walker – Mass}\label{fig:Walker_mass}
  \end{subfigure}\hfill
  \begin{subfigure}[b]{0.32\linewidth}
    \centering
    \includegraphics[width=\linewidth]{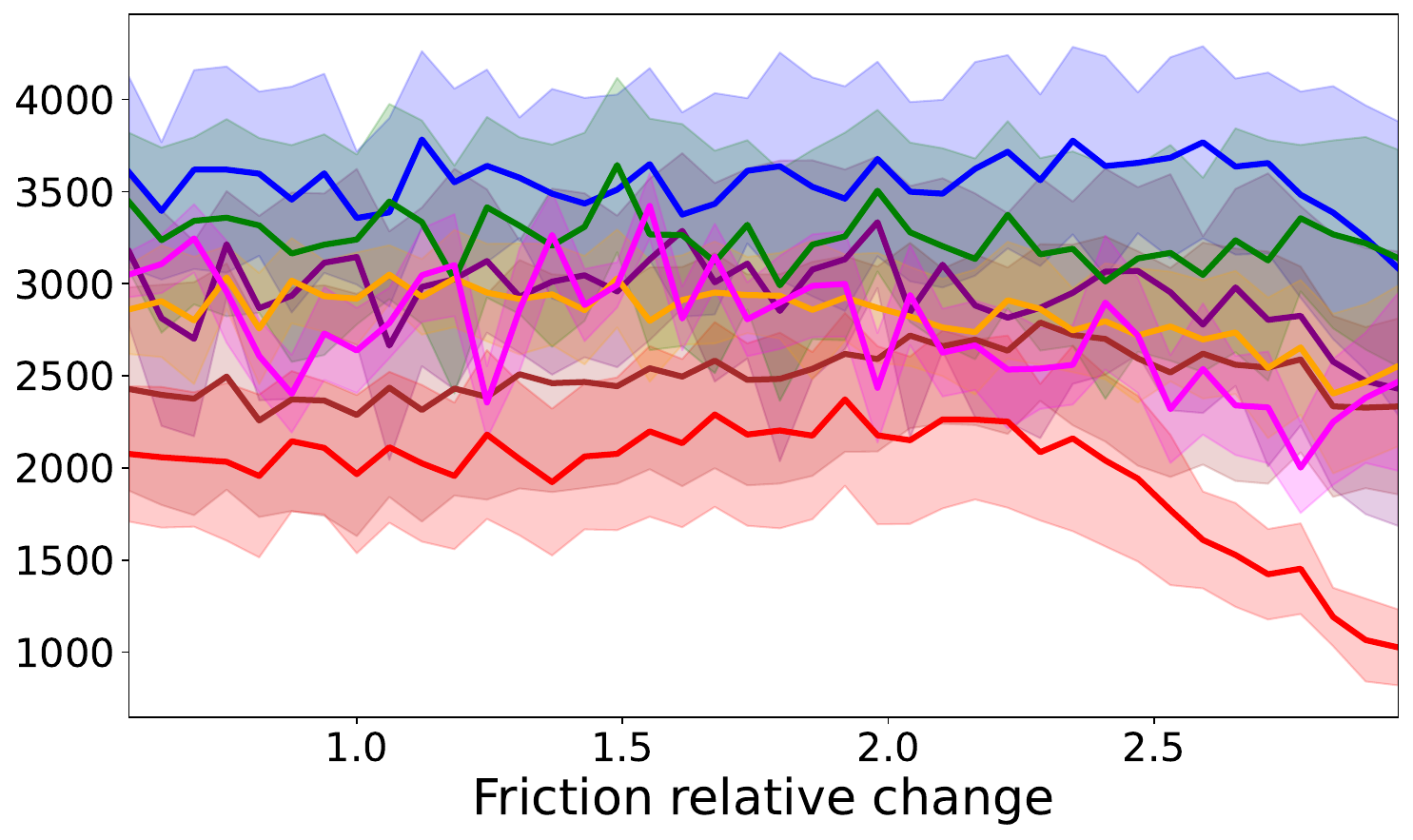}
    \subcaption{Walker – Friction}\label{fig:Walker_friction}
  \end{subfigure}\hfill
  \begin{subfigure}[b]{0.32\linewidth}
    \centering
    \includegraphics[width=\linewidth]{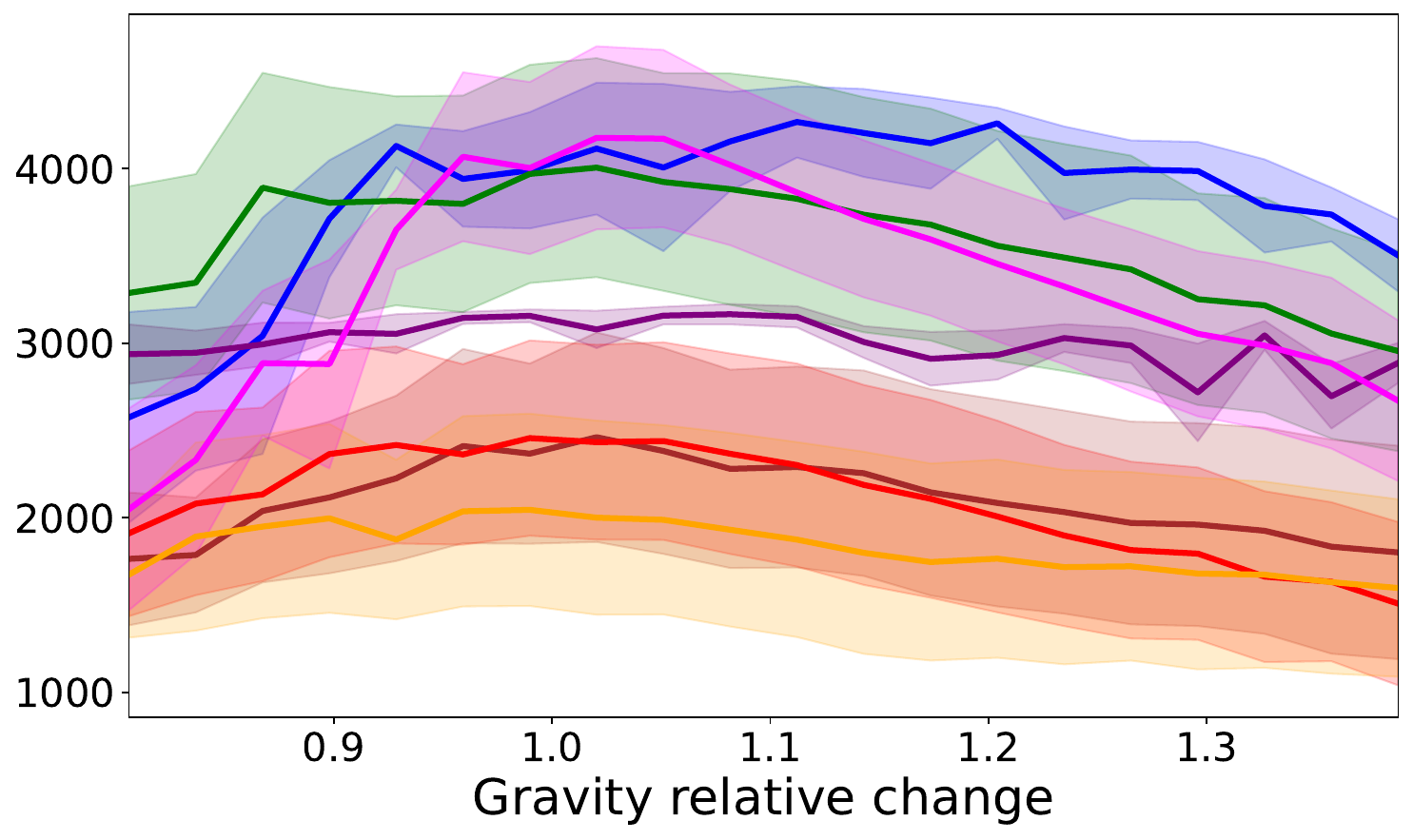}
    \subcaption{Cheetah - Gravity}\label{fig:cheetah_gravity}
  \end{subfigure}

  \vspace{0.4\baselineskip}

%============= Row 3 ================
  \begin{subfigure}[b]{0.32\linewidth}
    \centering
    \includegraphics[width=\linewidth]{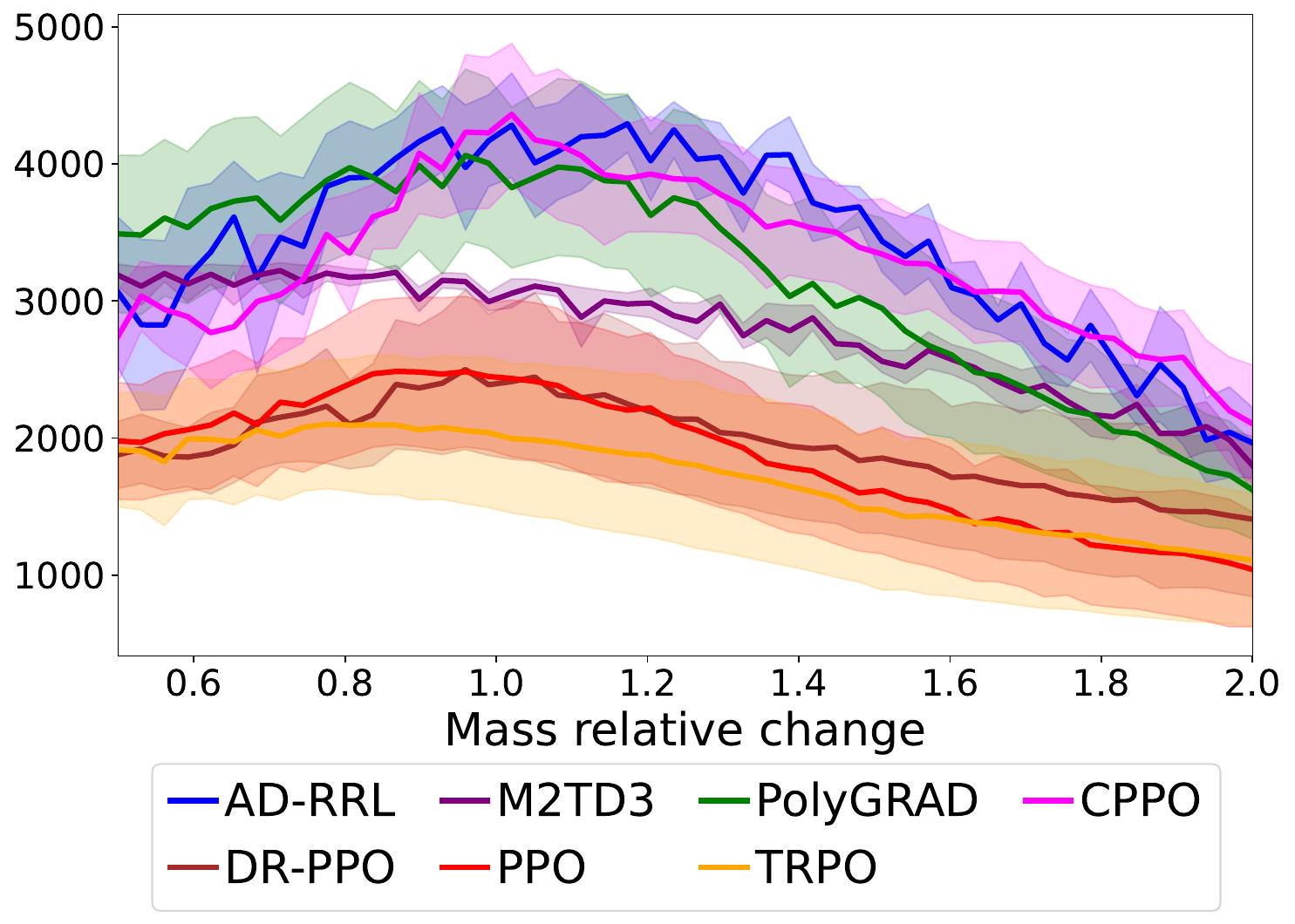}
    \subcaption{Cheetah – Mass}\label{fig:Cheetah_mass}
  \end{subfigure}\hfill
  \begin{subfigure}[b]{0.32\linewidth}
    \centering
    \includegraphics[width=\linewidth]{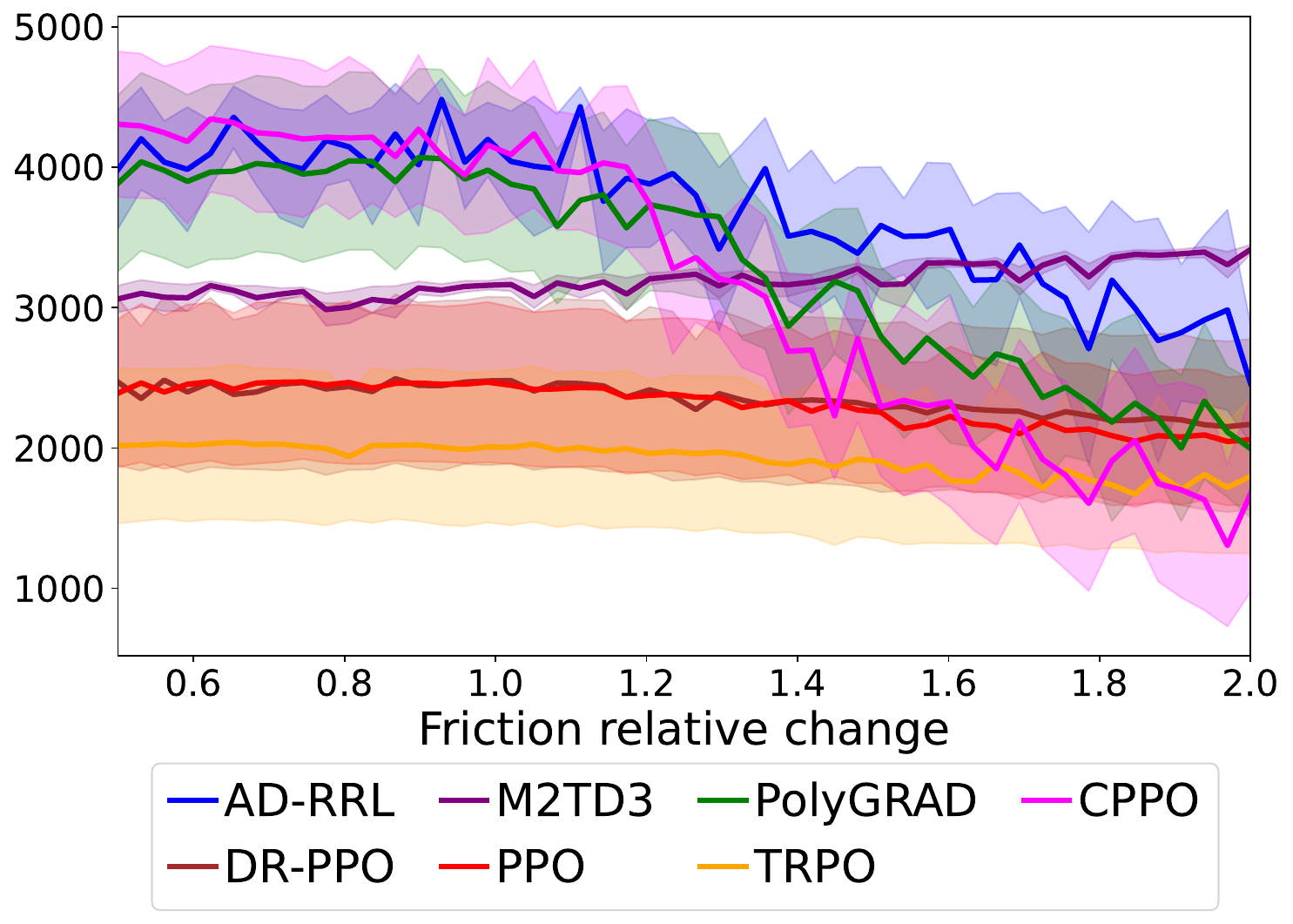}
    \subcaption{Cheetah – Friction}\label{fig:cheetah_friction}
  \end{subfigure}\hfill
  \begin{subfigure}[b]{0.32\linewidth}
    \centering
    \includegraphics[width=\linewidth]{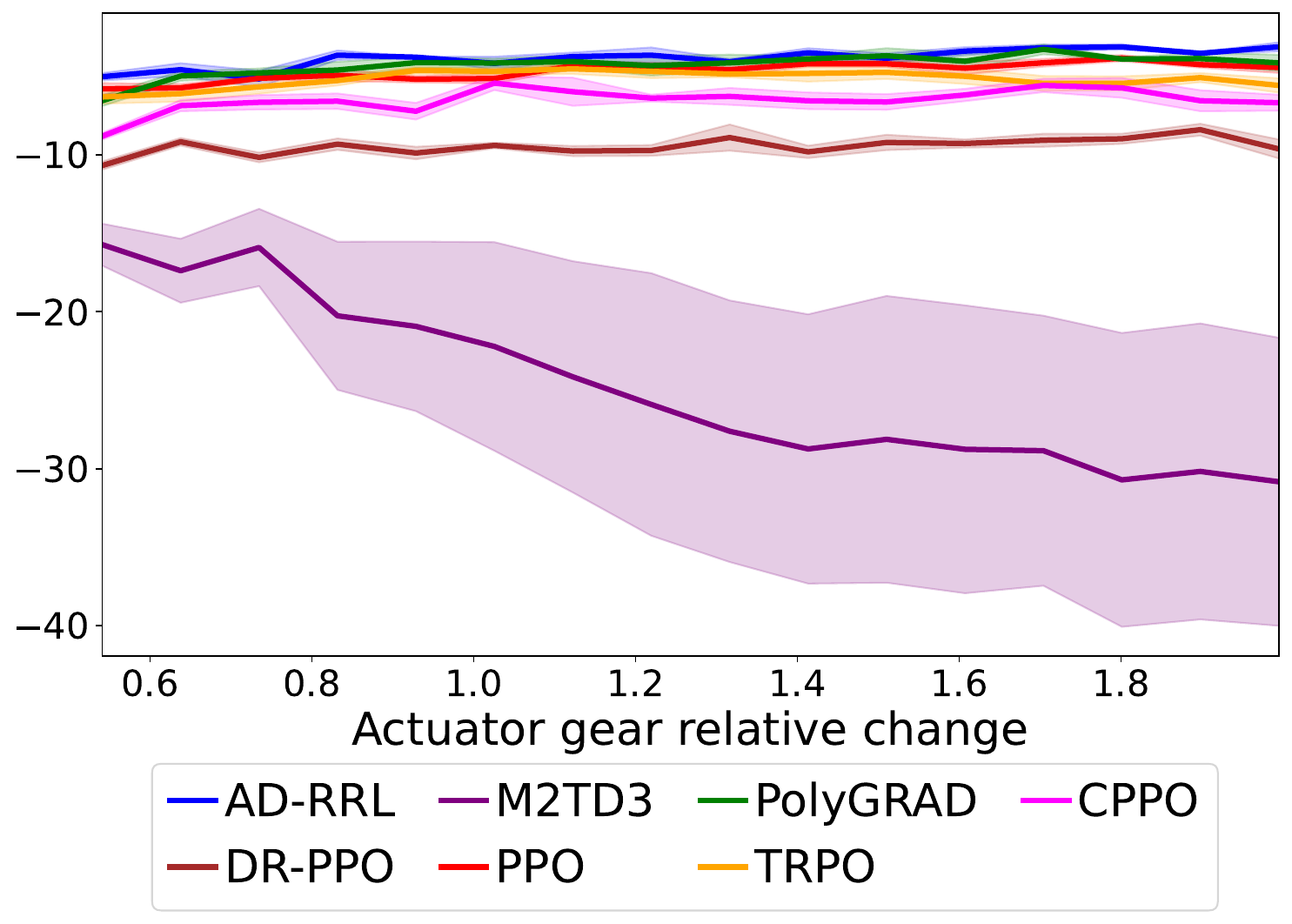}
    \subcaption{Reacher – Actuator Gear}\label{fig:Reacher_gear}
  \end{subfigure}

  \caption{Average return across variations in selected physics parameters. Shaded regions indicate $\pm$ one standard error.}
  \label{fig:robust_plots}
\end{figure}

In \cref{fig:robust_plots}, we plot the return under the different algorithms and for selected environments and varying parameters. Additional plots are provided in \cref{appendix:varying_parameters}. In most environments, AD-RRL consistently outperforms both robust and non-robust baselines. PPO and TRPO appear surprisingly stable, which is likely a consequence of the well-tuned Stable-Baselines3 implementations—but are still matched or surpassed by AD-RRL. At the same time, AD-RRL consistently outperforms both DR-PPO and M2TD3, demonstrating greater stability and achieving higher cumulative rewards across all environments.

Furthermore, a direct comparison with PolyGRAD (the foundation of our algorithm) highlights that our modifications significantly improve performance under diverse test-time conditions, enhancing robustness to large parameter shifts and model misspecifications. This can be clearly seen in \cref{fig:Walker_mass} or \cref{fig:Cheetah_mass}. In the Reacher environment (\cref{fig:Reacher_gear}) the difference in performance is less evident, but our model still performs consistently better or on par with the baselines. It is also clear from \cref{fig:Hopper_mass,fig:Hopper_gravity} that while PolyGRAD achieves slightly better performance on the nominal environment (as observed in \cref{tab:mujoco_results_se}), it sacrifices robustness under perturbed conditions.

For some environments—see for example \cref{fig:Hopper_mass} and Figure \ref{fig:pendulum_masscart} (presented in Appendix \ref{appendix:additional_results}), AD-RRL performance degrades for extreme changes in the modified parameter (but it remains better than other algorithms). We hypothesize that this is because our model is generating challenging trajectories which are nonetheless plausible under the agent policy and environment dynamics. Extreme changes in the environment physics do not reflect these constraints, and relevant trajectories might not be generated often.

{\bf Performance on the nominal (training) environments.}
Table~\ref{tab:mujoco_results_se} reports the final episode returns (mean $\pm$ one standard error) for five MuJoCo continuous-control tasks. Best results are highlighted in bold. The results are obtained on the training environment, with the nominal physics parameters. AD-RRL attains the best mean return on four of the five domains, %—HalfCheetah, Walker2d, Reacher, and Inverted Pendulum—
substantially outperforming the other baselines, showing that our risk-sensitive training does not trade nominal optimality for robustness. Only on Hopper, PolyGRAD performs better than AD-RRL, but the margin falls within overlapping confidence intervals. On the easy Inverted Pendulum task, multiple methods (AD-RRL, DR-PPO, PPO) reach the maximum score of~$1000$, as expected.

\begin{table}[h!]
\centering
\resizebox{\linewidth}{!}{
\begin{tabular}{@{}lrrrrr@{}}
\toprule
 & \textbf{Hopper} & \textbf{Cheetah} & \textbf{Walker} & \textbf{Reacher} & \textbf{InvertedPendulum} \\ \midrule
AD\textminus RRL &
$3280.23 \pm 13.83$ &
\textbf{4126.11 $\boldsymbol\pm$ 246.96} &
\textbf{4357.80 $\pm$ 187.37} &
$\boldsymbol{-}$\textbf{3.97 $\pm$ 0.13} &
\textbf{1000.00 $\pm$ 0.00} \\
PolyGRAD &
\textbf{3346.99 $\pm$ 52.39} &
$3879.16 \pm 626.40$ &
$3489.48 \pm 456.70$ &
$-4.48 \pm 0.13$ &
\textbf{1000.00 $\pm$ 0.00} \\
M2TD3 &
$361.73 \pm 13.71$ &
$3117.16 \pm 55.34$ &
$2948.03 \pm 598.77$ &
$-21.28 \pm 5.75$ &
$634.76 \pm 192.46$ \\
CPPO &
$2595.64 \pm 298.35$ &
$2173.30 \pm 422.97$ &
$2164.30 \pm 510.60$ &
$-6.06 \pm 0.32$ &
$979.85 \pm 20.15$ \\
DR\textminus PPO &
$2315.90 \pm 482.17$ &
$2429.46 \pm 558.93$ &
$2385.19 \pm 589.49$ &
$-15.80 \pm 1.44$ &
\textbf{1000.00 $\pm$ 0.00} \\
PPO &
$2998.90 \pm 432.28$ &
$2408.20 \pm 546.33$ &
$1894.03 \pm 349.06$ &
$-5.17 \pm 0.57$ &
\textbf{1000.00 $\pm$ 0.00} \\
TRPO &
$3270.27 \pm 273.04$ &
$2014.91 \pm 539.64$ &
$3090.80 \pm 267.79$ &
$-6.22 \pm 0.85$ &
$960.60 \pm 39.40$ \\ \bottomrule
\end{tabular}}
\vspace{0.1cm}
\caption{Return on the training environment (nominal physics parameters) for MuJoCo continuous-control tasks.}
\label{tab:mujoco_results_se}
\vspace{-0.5cm}
\end{table}

{\bf Sample efficiency.} The learning curves in \cref{fig:training_curves} (see Appendix \ref{appendix:training_curves}) show that AD-RRL reaches higher or matching final performance with the same number of samples as the baselines, and in several cases converges faster. Hence, our adversarially guided diffusion not only preserves (or improves) performance on the training environment (with nominal physics parameters), but also matches the sample efficiency of state-of-the-art model-based and model-free alternatives. In \cref{sec:ablation_study}, we perform an ablation study comparing AD-RRL trained on $1.5$M samples against the Model-Free Baselines trained on $3$M samples. This way we allow the model-free methods to compensate for their lower sample efficiency by leveraging their computational efficiency. Overall, \cref{fig:training_curves3m,fig:robust_plots3M} show that even when model-free baselines are given twice as many samples, AD-RRL remains on average better or on par across the proposed tasks.

\section{Conclusion and Future Work}

In this work we introduced AD-RRL, a novel approach to robust RL. AD-RRL is based on Adversarial Diffusion (AD), a diffusion model that can sample adversarial trajectories by leveraging the Conditional Value at Risk (CVaR) framework. AD enables agents to learn from adversarial scenarios that are either rare or unexplored in the environment. We demonstrated that AD-RRL, based on this diffusion model, significantly enhances the robustness of RL agents in the presence of modeling errors. Through empirical evaluation on multiple Gym/MuJoCo environments, we showed that AD-RRL outperforms current state-of-the-art robust RL methods.

AD relies on a specific strategy for guiding the diffusion process, and exploring alternative guidance methods presents a promising avenue for future work. Potential directions include (i) modifying the overall diffusion objective beyond the current CVaR framework, and (ii) enhancing the diffusion model architecture or algorithms to reduce computational overhead.

\newpage

\section{Acknowledgments}
The computations were enabled by resources provided by the National Academic Infrastructure for Supercomputing in Sweden (NAISS), partially funded by the Swedish Research Council through grant agreement no. 2022-06725. We also thank Adam Miksits for the invaluable support provided in the development of the numerical results.

\bibliography{example_paper}
\bibliographystyle{abbrv}

\newpage
\tableofcontents
\newpage

\appendix

\section{Adversarially Guided Diffusion Models: Proofs of results from Section \ref{subsec:perturbed_diff_model}}
\label{appendix:adg_diffusion}

\begin{proof}[Proof of \cref{lem:perturbed_f}]
For a sufficiently smooth function $r$, the conditional distribution $\bar p_{\boldsymbol\theta}(\boldsymbol\tau_i |\boldsymbol\tau_{i+1})$ can be approximated using a Gaussian. Following \cite[Appendix C]{sohl2015deep} (see also \cite{dhariwal2021diffusion}) we know that
\begin{align*}
    p_{\boldsymbol{\theta}}(\boldsymbol{\tau}_{i-1}| \boldsymbol{\tau}_i, \boldsymbol\tau_{i-1} \in C_\alpha) \approx \mathcal{N}(\mu_{\boldsymbol{\theta}}(\boldsymbol \tau_i,i) + \boldsymbol{\Sigma}_i \boldsymbol{h}_i, \boldsymbol{\Sigma}_i)
\end{align*}
where $\boldsymbol{h}_i = \nabla_{\boldsymbol{\tau}}\log p_{\boldsymbol\theta}(\boldsymbol\tau \in C_\alpha|\boldsymbol{\tau})|_{\boldsymbol{\tau}=\mu_{\boldsymbol{\theta}}(\boldsymbol \tau_i,i)}$.

Since we assume that the approximation ${p_{\boldsymbol{\theta}}}(\boldsymbol \tau_i \in C_\alpha | \boldsymbol \tau_i) = \exp{(-c_i\sum_{t = 1}^H \gamma^t r_t^{(i)})}$ holds, we get that
\begin{align}
    \boldsymbol{h}_i &= \nabla_{\boldsymbol{\tau}}\log p(\boldsymbol\tau \in C_\alpha |\boldsymbol \tau)|_{\boldsymbol{\tau}=\mu_{\boldsymbol{\theta}}(\boldsymbol \tau_i,ti)} \nonumber\\
    &=  -c_i\sum_{t=1}^T \nabla_{\boldsymbol{s_t, a_t}} r(s_t, a_t)|_{(s_t, a_t)=\mu^t_{\boldsymbol{\theta}}(\boldsymbol \tau_i,i)} \nonumber\\
    &= -c_i\nabla_{\boldsymbol{\tau}} Z(\boldsymbol \tau)|_{\boldsymbol{\tau}=\mu_{\boldsymbol{\theta}}(\boldsymbol \tau_i,i)}, \nonumber
\end{align}
where $\mu^t_{\boldsymbol{\theta}}(\boldsymbol \tau_i,i)$ is the $t$-th state-action pair of $\mu_{\boldsymbol{\theta}}(\boldsymbol \tau_i,i)$. 
Substituting, we get 
\begin{align}
    p_{\boldsymbol{\theta}}(\boldsymbol{\tau}_{i-1}| \boldsymbol{\tau}_i, \boldsymbol\tau_{i-1} \in C_\alpha) = \mathcal{N}(\mu_{\boldsymbol{\theta}}(\boldsymbol \tau_i,i) - c_i\boldsymbol{\Sigma}_i \boldsymbol{g}_i, \boldsymbol{\Sigma}_i),
\end{align}
where $\boldsymbol{g}_i = \nabla_{\boldsymbol{\tau}} Z(\boldsymbol \tau)|_{\boldsymbol{\tau}=\mu_{\boldsymbol{\theta}}(\boldsymbol \tau_i,i)}$. 
\end{proof}

\begin{proof}[Proof of \cref{eq:cond_sampling}]

As mentioned earlier, to sample from $p_{\boldsymbol \theta}(\boldsymbol\tau_0|  \boldsymbol\tau_0 \in C_\alpha)$, we multiply each intermediate distribution in the diffusion process by $r_i(\boldsymbol\tau_i)$, with $r_i(\boldsymbol\tau_i)=\exp(-c_i \sum_{t=1}^H \gamma^t r_t^{(i)})$, where the notation $r_t^{(i)}$ refers to the $t$-th reward in $\boldsymbol\tau_i$ for the $i$-th diffusion step. This means that the corresponding modified distribution $\bar p_{\boldsymbol \theta}$ satisfies in the intermediate diffusion step $i$:
\begin{equation}\label{eq:ddd}
\bar p_{\boldsymbol \theta}(\boldsymbol\tau_{i}) = \frac{1}{\bar Z_i} r_i(\boldsymbol\tau_i) p_{\boldsymbol \theta}(\boldsymbol\tau_i),
\end{equation}
where $\tilde Z_i$ is the normalizing constant. Next, we use the same strategy as that used in \cite{sohl2015deep} to determine the diffusion process $\bar p_{\boldsymbol \theta}(\boldsymbol\tau_{i} | \boldsymbol\tau_{i+1})$. Note first that:  
\[
\bar p_{\boldsymbol \theta}(\boldsymbol\tau_{i}) = \int \bar p_{\boldsymbol \theta}(\boldsymbol\tau_{i} | \boldsymbol\tau_{i+1})\bar p_{\boldsymbol \theta}(\boldsymbol\tau_{i+1}) {\rm d} \boldsymbol\tau_{i+1}.
\]
Plugging (\ref{eq:ddd}), the previous condition can be rewritten as
\begin{equation}\label{eq:dd1}
p_{\boldsymbol \theta}(\boldsymbol\tau_i) =\int \bar p_{\boldsymbol \theta}(\boldsymbol\tau_{i} | \boldsymbol\tau_{i+1}) \frac{\bar Z_i}{\bar Z_{i+1}}  \frac{r_{i+1}(\boldsymbol\tau_{i+1})}{ r_i(\boldsymbol\tau_i) } p_{\boldsymbol \theta}(\boldsymbol\tau_{i+1}) {\rm d} \boldsymbol\tau_{i+1}.
\end{equation}
However, we know that $p_{\boldsymbol \theta}$ also satisfies:
\[
p_{\boldsymbol \theta}(\boldsymbol\tau_i) =\int p_{\boldsymbol \theta}(\boldsymbol\tau_{i} | \boldsymbol\tau_{i+1})p_{\boldsymbol \theta}(\boldsymbol\tau_{i+1}) {\rm d} \boldsymbol\tau_{i+1}.
\]
This implies that (\ref{eq:dd1}) holds if: 
\[
 \bar p_{\boldsymbol \theta}(\boldsymbol\tau_{i} | \boldsymbol\tau_{i+1}) = p_{\boldsymbol \theta}(\boldsymbol\tau_{i} | \boldsymbol\tau_{i+1}) \frac{\bar Z_{i+1}r_i(\boldsymbol\tau_i)}{\bar Z_{i}r_{i+1}(\boldsymbol\tau_{i+1})}.
\]
Now defining the normalization constant $\bar Z_{i}(\boldsymbol\tau_{i+1}) = \frac{\bar Z_{i+1}}{\bar Z_{i}r_{i+1}(\boldsymbol\tau_{i+1})}$, we get
\[
 \bar p_{\boldsymbol \theta}(\boldsymbol\tau_{i} | \boldsymbol\tau_{i+1}) =\frac{1}{\bar Z_{i}(\boldsymbol\tau_{i+1})} p_{\boldsymbol \theta}(\boldsymbol\tau_{i} | \boldsymbol\tau_{i+1})r_i(\boldsymbol\tau_i).
\]
We conclude that $p_{\boldsymbol \theta}(\boldsymbol\tau_i | \boldsymbol\tau_{i+1},\boldsymbol\tau_i \in C_\alpha) \propto \tilde p_{\boldsymbol \theta}(\boldsymbol\tau_i |\boldsymbol\tau_{i+1})$.
Therefore, we have shown that:
\begin{align*}
p_{\boldsymbol \theta}(\boldsymbol\tau_0|  \boldsymbol\tau_0 \in C_\alpha) &= \bar p_{\boldsymbol \theta}(\boldsymbol\tau_{0}),\\
&= \int  \bar p_{\boldsymbol \theta}(\boldsymbol\tau_{0}|\boldsymbol\tau_1) \bar p_{\boldsymbol \theta}(\boldsymbol\tau_1) {\rm d}\boldsymbol\tau_1,\\
&= \int  \bar p_{\boldsymbol \theta}(\boldsymbol\tau_{0}|\boldsymbol\tau_1)\cdots \bar p_{\boldsymbol \theta}(\boldsymbol\tau_{N-1}|\boldsymbol\tau_N)  p_{\boldsymbol \theta}(\boldsymbol\tau_N) {\rm d}\boldsymbol\tau_1,\dots,\boldsymbol\tau_N,\\
&\propto \int   p_{\boldsymbol \theta}(\boldsymbol\tau_{0}|\boldsymbol\tau_1, \boldsymbol\tau_0 \in C_\alpha)\cdots  p_{\boldsymbol \theta}(\boldsymbol\tau_{N-1}|\boldsymbol\tau_N, \boldsymbol\tau_{N-1} \in C_\alpha) p_{\boldsymbol \theta}(\boldsymbol\tau_N) {\rm d}\boldsymbol\tau_1,\dots,\boldsymbol\tau_N.
\end{align*}
\end{proof}

\section{Adversarial guide as a multiplicative perturbation: Proof of \cref{lem:multiplicative_noise}}
\label{appendix:adv_pert}

\begin{proof}[Proof of \cref{lem:multiplicative_noise}]
Let's define a denoising diffusion model $ p_{\boldsymbol{\theta}}(\boldsymbol{\tau}_{i-1}| \boldsymbol{\tau}_i)$, and a perturbed denoising step of the form $p_{\boldsymbol{\theta}}(\boldsymbol{\tau}_{i-1}| \boldsymbol{\tau}_
i, \boldsymbol\tau_{i-1} \in C_\alpha ) = \mathcal{N}(\mu_{\boldsymbol{\theta}}(\boldsymbol \tau_i,i) - c_i\boldsymbol{\Sigma}_i \boldsymbol{g}_i, \boldsymbol{\Sigma}_i)$. Since the two distributions are Gaussians with known mean and covariance matrices, we have
\begin{align}
    p_{\boldsymbol{\theta}}(\boldsymbol{\tau}_{i-1}| \boldsymbol{\tau}_
i, \boldsymbol\tau_{i-1} \in C_\alpha ) &= \mathcal{N}(\mu_{\boldsymbol{\theta}}(\boldsymbol \tau_i,i) - c_i\boldsymbol{\Sigma}_i \boldsymbol{g}_i, \boldsymbol{\Sigma}_i) \nonumber \\
    &= K\exp{\left(-\frac{1}{2}(\boldsymbol D_i + c_i\boldsymbol \Sigma_i \boldsymbol g_i)^T \boldsymbol \Sigma_i^{-1}(\boldsymbol D_i + c_i\boldsymbol \Sigma_i \boldsymbol g_i)\right)} \nonumber\\
    &= K\exp{\left(-\frac{1}{2}(\boldsymbol D_i^T \boldsymbol \Sigma_i \boldsymbol D_i + 2c_i \boldsymbol D_i^T \boldsymbol g_i +  c_i^2\boldsymbol g_i^T \boldsymbol \Sigma_i \boldsymbol g_i)\right)} \nonumber\\
    &= K\exp{\left(-\frac{1}{2}\boldsymbol D_i^T \boldsymbol \Sigma_i \boldsymbol D_i\right)} \exp{\left( -\frac{1}{2}(2c_i\boldsymbol D_i^T \boldsymbol g_i + c_i^2\boldsymbol g_i^T \boldsymbol \Sigma_i \boldsymbol g_i)\right)} \nonumber \\
    &= {\cal N}(\boldsymbol \tau_{i-1}|\mu_{\boldsymbol{\theta}}(\boldsymbol \tau_i,i), \boldsymbol \Sigma_i) \exp{\left( -\frac{1}{2}(2c_i\boldsymbol D_i^T \boldsymbol g_i + c_i^2\boldsymbol g_i^T \boldsymbol \Sigma_i \boldsymbol g_i)\right)} \nonumber\\
    &= \xi(\boldsymbol \tau_{i}, \boldsymbol \tau_{i-1})p_{\boldsymbol{\theta}}(\boldsymbol{\tau}_{i-1}| \boldsymbol{\tau}_i)\nonumber
\end{align}
where $K =  \frac{1}{(2\pi)^{d/2}|\boldsymbol \Sigma_i|^{1/2}}$, $\boldsymbol D_i = (\boldsymbol \tau_{i-1} - \mu_{\boldsymbol{\theta}}(\boldsymbol \tau_i,i))$, and $\xi(\boldsymbol \tau_{i}, \boldsymbol \tau_{i-1}) = \exp{\left( -\frac{1}{2}(2c_i\boldsymbol D_i^T \boldsymbol g_i + c_i^2 \boldsymbol g_i^T \boldsymbol \Sigma \boldsymbol g_i)\right)}$. ${\cal N}(\boldsymbol \tau_{i-1}|\mu_{\boldsymbol{\theta}}(\boldsymbol \tau_i,i), \boldsymbol \Sigma_i)$ is the density of the Gaussian distribution of $\boldsymbol \tau_{i-1}$, with mean $\mu_{\boldsymbol{\theta}}(\boldsymbol \tau_i,i)$ and covariance matrix $\boldsymbol \Sigma_i$.

If we define $\xi(\boldsymbol \tau_{0:N}) = \prod_{i=1}^N \xi(\boldsymbol{\tau}_{i}, \boldsymbol{\tau}_{i-1})$, we have
\begin{align}
    p_{\boldsymbol \theta}(\boldsymbol{\tau}_{0:N}|\boldsymbol\tau_0 \in C_\alpha) = \xi(\boldsymbol \tau_{0:N})p(\boldsymbol{\tau}_{0:N})\nonumber
\end{align}
So we can define
\begin{align}
    \bar p_{\boldsymbol{\theta}}(\boldsymbol \tau_0) = p_{\boldsymbol \theta}(\boldsymbol \tau_0 |\boldsymbol\tau_0 \in C_\alpha) &= \int p_{\boldsymbol \theta}(\boldsymbol{\tau}_{0:N}|\boldsymbol\tau_0 \in C_\alpha) = 1)\text{d}\boldsymbol{\tau}_{1:N} \nonumber\\
    &= \int \xi(\boldsymbol \tau_{0:N})p(\boldsymbol{\tau}_{0:N})\text{d}\boldsymbol{\tau}_{1:N} \nonumber\\
    &= \frac{\int \xi(\boldsymbol \tau_{0:N})p(\boldsymbol{\tau}_{0:N})\text{d}\boldsymbol{\tau}_{1:N}}{P(\boldsymbol \tau_0)} P(\boldsymbol \tau_0) \nonumber\\
    &= \xi(\boldsymbol \tau_0) P(\boldsymbol \tau_0)\nonumber
\end{align}
with $\xi(\boldsymbol \tau_0) = \frac{\int \xi(\boldsymbol \tau_{0:N})p(\boldsymbol{\tau}_{0:N})\text{d}\boldsymbol{\tau}_{1:N}}{P(\boldsymbol \tau_0)}$.
\end{proof}

\section{Attaining the duality constraint on $\xi(\boldsymbol \tau_0)$: Proof of \cref{prop:ci_constraint}}
\label{appendix:constraint}

\begin{proof}[Proof of \cref{prop:ci_constraint}]\hfill

\textit{Proof of (a)}: From \cref{lem:multiplicative_noise} we know that $\xi(\boldsymbol \tau_0) = \frac{\int \xi(\boldsymbol \tau_{0:N})p(\boldsymbol{\tau}_{0:N})\text{d}\boldsymbol{\tau}_{1:N}}{P(\boldsymbol \tau_0)}$. 
We want to have  $  \xi(\boldsymbol{\tau}_{0}) \leq \frac{1}{\alpha}$, this is equivalent to
\begin{align}
    %\xi(\boldsymbol{\tau}_{0}) &\leq \frac{1}{\alpha}, \nonumber\\
    \frac{\int \xi(\boldsymbol \tau_{0:N})p(\boldsymbol{\tau}_{0:N})\text{d}\boldsymbol{\tau}_{1:N}}{P(\boldsymbol \tau_0)} &\leq \frac{1}{\alpha}, \nonumber\\
     \int \xi(\boldsymbol \tau_{0:N})p(\boldsymbol{\tau}_{0:N})\text{d}\boldsymbol{\tau}_{1:N}&\leq \frac{1}{\alpha}\int p(\boldsymbol{\tau}_{0:N})\text{d}\boldsymbol{\tau}_{1:N}. \nonumber
\end{align}
One way to achieve this is to impose $\xi(\boldsymbol \tau_{0:N}) = \prod_{i=1}^N \xi(\boldsymbol{\tau}_{i}, \boldsymbol{\tau}_{i-1}) \leq \frac{1}{\alpha}$. This is satisfied also by constraining the single terms of the product using $\eta_i(\alpha, N)$ such that $\xi(\boldsymbol{\tau}_{i}, \boldsymbol{\tau}_{i-1}) \leq \eta_i(\alpha, N)$ and $\prod_{i=1}^N \eta_i(\alpha, N) = \frac{1}{\alpha}$.

However, $\boldsymbol \tau_{i-1}$ is a random quantity to which we do not have access at step $i$ of the diffusion process. Therefore, to satisfy the constraints on the single terms we impose 
\begin{align}
    \max_{\boldsymbol\tau_{i-1}} \xi(\boldsymbol\tau_{i-1}, \boldsymbol\tau_i) \leq \eta_i(\alpha, N).\nonumber
\end{align}

$\xi(\boldsymbol\tau_{i-1}, \boldsymbol\tau_i)$ is maximized for $(\boldsymbol \tau_{i-1} - \mu_{\boldsymbol{\theta}}(\boldsymbol \tau_i,i))^T \boldsymbol g < 0$, so we want $(\boldsymbol \tau_{i-1} - \mu_{\boldsymbol{\theta}}(\boldsymbol \tau_i,i))^T$ to be a vector opposite to $\boldsymbol g$. We can take $\boldsymbol \tau_{i-1} = \mu_{\boldsymbol{\theta}}(\boldsymbol \tau_i,i) - c_i\boldsymbol \Sigma_i \boldsymbol g_i$.

We assume that the trajectories $\boldsymbol  \tau_i$ lie in a bounded space $C = \{ \boldsymbol \tau_i : ||\boldsymbol 
 \tau_i||_\infty \leq \rho \}$, where $||\boldsymbol 
 \tau_i||_\infty = \max_{\boldsymbol  s \in \boldsymbol \tau_i} ||\boldsymbol  s||_\infty$. From this assumption it follows that
\begin{align}
    ||\boldsymbol \tau_{i-1}||_\infty &\leq \rho \nonumber\\
    ||\mu_{\boldsymbol{\theta}}(\boldsymbol \tau_i,i) - c_i \boldsymbol \Sigma_i \boldsymbol g_i||_\infty &\leq \rho \nonumber\\
    ||\mu_{\boldsymbol{\theta}}(\boldsymbol \tau_i,i)||_\infty + c_i||\boldsymbol \Sigma_i \boldsymbol g_i||_\infty &\leq \rho \nonumber\\
     c_i &\leq \frac{(\rho - ||\mu_{\boldsymbol{\theta}}(\boldsymbol \tau_i,i)||_\infty)}{||\boldsymbol \Sigma_i \boldsymbol g_i||_\infty}. \nonumber
\end{align}

Substituting $\boldsymbol \tau_{i-1} = \mu_{\boldsymbol{\theta}}(\boldsymbol \tau_i,i) - c_i\boldsymbol \Sigma_i \boldsymbol g_i$ into $\xi(\boldsymbol\tau_{i-1}, \boldsymbol\tau_i) \leq \eta_i(\alpha, N)$ and developing we get
\begin{align}
    -\frac{1}{2}(-2 c_i^2\boldsymbol g_i^T \boldsymbol \Sigma_i \boldsymbol g_i + c_i^2 \boldsymbol g_i^T \boldsymbol \Sigma_i \boldsymbol g_i) &\leq \log\eta_i(\alpha, N) \nonumber\\
    c_i  &\leq \sqrt{\frac{2\log\eta_i(\alpha, N)}{\boldsymbol g_i^T \boldsymbol \Sigma_i \boldsymbol g_i}}. \nonumber
\end{align}

So combining the two inequalities we can take

\begin{align}
    c_i \leq \min\left(\sqrt{\frac{2\log\eta_i(\alpha, N)}{\boldsymbol g_i^T \boldsymbol \Sigma_i \boldsymbol g_i}}, \frac{\rho - ||\mu_{\boldsymbol{\theta}}(\boldsymbol \tau_i,i)||_\infty}{||\boldsymbol \Sigma_i \boldsymbol g_i||_\infty}\right).
    \label{eq:ci_min}
\end{align}

\textit{Proof of (b)}: To find the minimum between the terms in \cref{eq:ci_min}, we analyze the denominators and numerators. For the denominators, $\sqrt{\boldsymbol g_i^T \boldsymbol \Sigma_i \boldsymbol g_i}$ and $||\boldsymbol \Sigma_i \boldsymbol g_i||_\infty = \max_j |(\boldsymbol \Sigma_i \boldsymbol g_i)_j|$, it is equivalent to compare $\boldsymbol g_i^T \boldsymbol \Sigma_i \boldsymbol g_i$ and $||\boldsymbol \Sigma_i \boldsymbol g_i||^2_\infty$. 

Since our diffusion model adopts a cosine scheduler for the covariance matrix, $\Sigma_i$ is diagonal with elements $(\boldsymbol\Sigma_i)_{jj} \in [0,1)$. We can write the $j-$th element of $\boldsymbol\Sigma_i \boldsymbol g_i$ as $(\boldsymbol\Sigma_i \boldsymbol g_i)_j = \boldsymbol e_j^T \boldsymbol \Sigma_i \boldsymbol g_i$, where $\boldsymbol e_j$ is a basis vector. Then using Cauchy-Schwarz inequality we get
\begin{align}
	|(\boldsymbol \Sigma_i \boldsymbol g_i)_j|^2 &= |\boldsymbol e_j^T \boldsymbol \Sigma_i \boldsymbol g_i|^2 \nonumber\\
	&\leq (\boldsymbol e_j^T \boldsymbol\Sigma_i \boldsymbol e_j)(\boldsymbol g_i^T \boldsymbol \Sigma_i \boldsymbol g_i), \nonumber
\end{align}
with $\boldsymbol e_j^T \boldsymbol\Sigma_i \boldsymbol e_j \leq 1$ by definition of $\boldsymbol\Sigma_i$. It follows that $|(\boldsymbol \Sigma_i \boldsymbol g_i)_j|^2 \leq \boldsymbol g_i^T \boldsymbol \Sigma_i \boldsymbol g_i$, and since this is true for all $j$ we can conclude that
\begin{align}
	\max_j |(\boldsymbol \Sigma_i \boldsymbol g_i)_j|^2 &\leq \boldsymbol g_i^T \boldsymbol \Sigma_i \boldsymbol g_i \nonumber\\
	||\boldsymbol \Sigma_i \boldsymbol g_i||_\infty \leq \sqrt{\boldsymbol g_i^T \boldsymbol \Sigma_i \boldsymbol g_i} \nonumber
\end{align}

When comparing the numerators of both terms in \cref{eq:ci_min}, since $\log \eta_i(\alpha, N) = \frac{1}{N} \log\left(\frac{1}{\alpha}\right)$, for $N$ large enough, $\sqrt{2\log\eta_i(\alpha, N)} < \rho - ||\mu_{\boldsymbol{\theta}}(\boldsymbol \tau_i,i)||_\infty$.

So $c_i \leq \sqrt{\frac{2\log\eta_i(\alpha, N)}{\boldsymbol g_i^T \boldsymbol \Sigma_i \boldsymbol g_i}}$ satisfies the dual CVaR constraints.
\end{proof}

\section{Adaptation to Matrix Normal Distribution}
\label{appendix:matrix_dist}

Here we extend the analysis to the case where states and actions are multidimensional, and we consider trajectories as Gaussian matrices.

We define $p(\boldsymbol \tau_{i-1} |\boldsymbol \tau_i)$ as
\begin{align}
    p(\boldsymbol \tau_{i-1} |\boldsymbol \tau_i) &= \mathcal{M}N(\boldsymbol \tau_{i-1} | M_{\boldsymbol \theta}(\boldsymbol \tau_i, i), \boldsymbol U_i, \boldsymbol V_i) \nonumber
\end{align}
where $\mathcal{M}N(\boldsymbol \tau_{i-1} | M_{\boldsymbol \theta}(\boldsymbol \tau_i, i), \boldsymbol U_i, \boldsymbol V_i)$ is a Matrix Normal Distribution with mean $ M_{\boldsymbol \theta}(\boldsymbol \tau_i, i) \in \mathbb{R}^{n \times p}$, row and column covariances $\boldsymbol U_i \in \mathbb{R}^{n \times n}$ and $\boldsymbol V_i \in \mathbb{R}^{p \times p}$. 

The probability density function of this Matrix Normal Distribution is defined as
\begin{align}
    \mathcal{M}N(\boldsymbol \tau_{i-1} | M_{\boldsymbol \theta}(\boldsymbol \tau_i, i), \boldsymbol U_i, \boldsymbol V_i)\coloneqq K_i\exp\left(  -\frac{1}{2}\text{Tr}[\boldsymbol V_i^{-1}(\boldsymbol \tau_{i-1}- M_{\boldsymbol \theta}(\boldsymbol \tau_i, i))^T \boldsymbol U_i^{-1}(\boldsymbol \tau_{i-1}- M_{\boldsymbol \theta}(\boldsymbol \tau_i, i))]\right) \nonumber
\end{align}

where $\text{Tr}[\cdot]$ is the trace operator, $K_i = \frac{1}{(2\pi)^{np/2} |\boldsymbol V_i|^{n/2}|\boldsymbol U_i|^{n/2}}$ and $|\cdot|$ is the determinant of a matrix. 

Define $\boldsymbol G_i \in \mathbb{R}^{n\times p}$ as the gradient $\nabla_{\boldsymbol{\tau}} Z$ with respect to the second order tensor representing the trajectory $\boldsymbol \tau \in \mathbb{R}^{n\times p}$ evaluated at $ M_{\boldsymbol \theta}(\boldsymbol \tau_i, i)$. Also define $\boldsymbol \Gamma_i = \boldsymbol U_i \boldsymbol G_i \boldsymbol V_i$ for notation convenience. Consider the perturbed distribution with a mean $ M_{\boldsymbol\theta}(\boldsymbol \tau_i, i) - c_i \boldsymbol \Gamma_i$ , we get
\begin{align}
    &p_{\boldsymbol{\theta}}(\boldsymbol{\tau}_{i-1}| \boldsymbol{\tau}_
i, \boldsymbol\tau_{i-1} \in C_\alpha ) \nonumber\\
&= \exp\left(  -\frac{1}{2}\text{Tr}[\boldsymbol V^{-1}(\boldsymbol \tau_{i-1}- M_{\boldsymbol \theta}(\boldsymbol \tau_i, i) + c_i \boldsymbol \Gamma_i)^T \boldsymbol U^{-1}(\boldsymbol \tau_{i-1}- M_{\boldsymbol \theta}(\boldsymbol \tau_i, i) + c_i \boldsymbol \Gamma_i)]\right) \nonumber\\
&= K\exp\left(  -\frac{1}{2}\text{Tr}[\boldsymbol V^{-1}(\boldsymbol \tau_{i-1}- M_{\boldsymbol \theta}(\boldsymbol \tau_i, i))^T \boldsymbol U^{-1}(\boldsymbol \tau_{i-1}- M_{\boldsymbol \theta}(\boldsymbol \tau_i, i))]\right)\xi(\boldsymbol \tau_{i}, \boldsymbol \tau_{i-1}) \nonumber
\end{align}
with
\begin{align}\xi(\boldsymbol \tau_{i}, \boldsymbol \tau_{i-1}) = \exp\left(  -\frac{1}{2}\text{Tr}[\boldsymbol V^{-1}_i(2c_i(\boldsymbol \tau_{i-1}- M_{\boldsymbol\theta}(\boldsymbol \tau_i, i))^T\boldsymbol U^{-1}_i\boldsymbol \Gamma_i + c_i^2\boldsymbol \Gamma_i^T\boldsymbol U^{-1}_i\boldsymbol \Gamma_i)]\right). \nonumber
\end{align}

As we did in Appendix \ref{appendix:constraint}, we take $\xi(\boldsymbol{\tau}_{i}, \boldsymbol{\tau}_{i-1}) \leq \eta_i(\alpha, N)$. We can take $\boldsymbol \tau_{i-1} = M_{\boldsymbol\theta}(\boldsymbol \tau_i, i) -  c_i\boldsymbol \Gamma_i$ and get
\begin{align}
\exp\left(  -\frac{1}{2}\text{Tr}[\boldsymbol V^{-1}_i(2c_i(\boldsymbol \tau_{i-1}- M_{\boldsymbol\theta}(\boldsymbol \tau_i, i))^T\boldsymbol U^{-1}_i\boldsymbol \Gamma_i + c_i^2\boldsymbol \Gamma_i^T\boldsymbol U^{-1}_i\boldsymbol \Gamma_i)]\right) &\leq \eta_i(\alpha, N) \nonumber\\
-\frac{1}{2}\text{Tr}[\boldsymbol V^{-1}_i(-2c_i^2\boldsymbol \Gamma_i^T\boldsymbol U^{-1}_i\boldsymbol \Gamma_i + c^2_i\boldsymbol \Gamma_i^T\boldsymbol U^{-1}_i\boldsymbol \Gamma_i)] &\leq \log\eta_i(\alpha, N) \nonumber\\
     \text{Tr}[\boldsymbol V^{-1}_i(\boldsymbol \Gamma_i^T\boldsymbol U^{-1}_i\boldsymbol \Gamma_i)c^2_i)] &\leq 2\log\eta_i(\alpha, N), \nonumber
\end{align}
giving 
\begin{align}
    c_i &\leq \sqrt{\frac{2\log\eta_i(\alpha, N)}{\text{Tr}[\boldsymbol V^{-1}_i(\boldsymbol U_i\boldsymbol G_i\boldsymbol V_i)^T\boldsymbol U^{-1}_i(\boldsymbol U_i\boldsymbol G_i\boldsymbol V_i))]}}. \nonumber
\end{align}

Under the same assumptions of \cref{appendix:constraint}, we get that
\begin{align}
    ||\boldsymbol \tau_{i-1}||_\infty &\leq \rho \nonumber\\
    ||M_{\boldsymbol\theta}(\boldsymbol \tau_i, i) - c_i\boldsymbol U_i\boldsymbol G_i\boldsymbol V_i||_\infty &\leq \rho \nonumber\\
    || M_{\boldsymbol\theta}(\boldsymbol \tau_i, i)||_\infty + c_i||\boldsymbol U_i\boldsymbol G_i\boldsymbol V_i||_\infty &\leq \rho \nonumber\\
     c_i &\leq \frac{(\rho - || M_{\boldsymbol\theta}(\boldsymbol \tau_i, i)||_\infty)}{||\boldsymbol U_i\boldsymbol G_i\boldsymbol V_i||_\infty}. \nonumber
\end{align}

So we can pick
\begin{align}
    c_i = \min\left( \sqrt{\frac{2\log\eta_i(\alpha, N)}{\text{Tr}[\boldsymbol V^{-1}_i(\boldsymbol U_i\boldsymbol G_i\boldsymbol V_i)^T\boldsymbol U^{-1}_i(\boldsymbol U_i\boldsymbol G_i\boldsymbol V_i))]}}, \frac{(\rho - || M_{\boldsymbol\theta}(\boldsymbol \tau_i, i)||_\infty)}{||\boldsymbol U_i\boldsymbol G_i\boldsymbol V_i||_\infty}\right) \nonumber
\end{align}

Following the same reasoning as in \cref{appendix:constraint}, if the covariance matrices $\boldsymbol U_i$ and $\boldsymbol V_i$ are diagonal with elements in $[0,1)$ we can pick
\begin{align}
    c_i = \sqrt{\frac{2\log\eta_i(\alpha, N)}{\text{Tr}[\boldsymbol V^{-1}_i(\boldsymbol U_i\boldsymbol G_i\boldsymbol V_i)^T\boldsymbol U^{-1}_i(\boldsymbol U_i\boldsymbol G_i\boldsymbol V_i))]}} \nonumber
\end{align}

\section{Implementation details}
\label{appendix:implementation_details}
Our method makes use of three MLPs: the policy $\pi_{\boldsymbol \omega}$, the adversarial denoising diffusion model $\bar p_{\boldsymbol \theta}$ and the learned cumulative reward function $Z_{\boldsymbol \phi}$.

\paragraph{Policy network and training.}
The policy $\pi_{\boldsymbol \omega}$ is parameterized in the same way as PolyGRAD \cite{rigter2023world}. We consider a Gaussian policy of the form $\pi_{\boldsymbol \omega} = \mathcal{N}(\mu_{\boldsymbol \omega}(s), \sigma_{\boldsymbol \omega})$, where $\boldsymbol \omega$ are the parameters of the MLP. The standard deviation of the policy is a single learnable parameter $\sigma_{\boldsymbol \omega}$, independent of the state.

The policy is trained using Advantage Actor Critic (A2C) with Generalised Advantage Estimation (GAE). The optimizer used is ADAM. The hyperparameters can be found in \cref{tab:hyperparameters_policy}.

\begin{table}[h!]
\centering
\begin{tabular}{@{}p{6cm}p{4cm}@{}}
\toprule
\textbf{Parameter} & \textbf{Value} \\ \midrule
Batch size & 512 \\
Synthetic trajectory length & 10 \\
GAE $\lambda$ & 0.9 \\
Critic learning rate & 3e-4 \\
Actor learning rate & 3e-5 \\
Discount factor, $\gamma$ & 0.99 \\
Entropy bonus weight & 1e-5 \\\\ \bottomrule
\end{tabular}
\caption{Hyperparameters for A2C training.}
\label{tab:hyperparameters_policy}
\end{table}

\paragraph{Adversarial Diffusion and Cumulative Reward models.}

Our implementation builds directly on top of PolyGRAD. We follow the same training procedure, summarized in Algorithm \ref{alg:diffusion_training}. For the Diffusion Model, we use the same MLP architecture as PolyGRAD, trained by minimizing the L2 loss with ADAM optimizer. The MDP has skip connections at every layer, and features a learneable embedding of the diffusion step $i$, which is common for Diffusion Architectures \cite{janner2022planning}. The hyperparameters are summarized in \cref{tab:hyperparameters_diffusion}.

\begin{table}[h!]
\centering
\begin{tabular}{@{}p{6cm}p{4cm}@{}}
\toprule
\textbf{Parameter} & \textbf{Value} \\ \midrule
Hidden size & 1024 \\
Length of generated trajectory & 10 \\
Batch size & 256 \\
Diffusion step embedding size & 128 \\
Number of layers & 6 \\
Learning rate & 3e-4 \\ 
\bottomrule
\end{tabular}
\caption{Hyperparameters for adversarial diffusion training.}
\label{tab:hyperparameters_diffusion}
\end{table}

When computing $c_i$ according to (\ref{eq:ci}), we chose $\rho = 3\sigma_i$, where $\sigma_i$ is the standard deviation of the diffusion process\footnote{In Denoising Diffusion Probabilistic Models, the standard deviation is fixed at every step $i$ according to a known scheduling rate \cite{ho2020denoising}.} at step $i$. In our implementation we choose $\eta_i(\alpha, N) = \left(\frac{1}{\alpha}\right)^\frac{1}{N}$. 

The cumulative reward model $Z_{\boldsymbol \phi}$ follows the same structure and hyperparameters of the Diffusion Model (also the step embedding), with a final linear layer producing a scalar output. It is optimized using the L2 loss and the ADAM optimizer.

\begin{algorithm}[H]
\caption{Diffusion model training}
\label{alg:diffusion_training}
\begin{algorithmic}[1]
\State {\bfseries Input:} adversarial denoising model $\bar{p}_{\boldsymbol{\theta}}$; cumulative reward function $Z_{\boldsymbol \phi}$; data buffer, $\mathcal{D}$; diffusion steps $N$; training iterations $K$
\For{$k=1, \ldots, K$}
\State  Improve $\bar{p}_{\boldsymbol{\theta}}$ (\ref{eq:vlb}) using $\{\boldsymbol{\tau}_0\}\sim \mathcal{D}$.
\State  Train $Z_{\boldsymbol \phi}$ to predict the reward $Z(\boldsymbol\tau_0)$.
\EndFor
\end{algorithmic}
\end{algorithm}

\paragraph{Baselines implementation}
The implementation of PolyGRAD was taken from the respective github repository \cite{rigter2023world}. The same was done for CPPO and M2TD3: for M2TD3, we created a new config file for Reacher, where we set the range of the actuator gear to $[50.0, 500.0]$. For TRPO and PPO we used the implementation from Stable-Baselines3 \cite{stable-baselines3}. We used Domain Randomization on top of the PPO baseline, training with values of the mass and parameters uniformly sampled according to the uncertainty intervals specified in \cref{tab:Uncertainty_sets_DR}. 

\begin{table}[h!]
\centering
\setlength{\tabcolsep}{6pt}    % tighten horizontal spacing a bit
\begin{tabular}{@{}lccccc@{}}
\toprule
\textbf{Environment} &
\textbf{Mass} &
\textbf{Friction} &
\textbf{Mass\,pole} &
\textbf{Mass\,cart} &
\textbf{Act.\ gear} \\ \midrule
Hopper       & {[0.5,\,6.5]} & {[0.1,\,3.0]}  & — & — & — \\
HalfCheetah  & {[3.5,\,9.5]} & {[0.2,\,0.8]}  & — & — & — \\
Walker       & {[0.5,\,6.5]} & {[0.5,\,2.0]}  & — & — & — \\ 
Cartpole       & — & — & {[2.5,\,10.0]} & {[5.0,\,20.0]} & — \\
Reacher       & — & — & — & — & {[50.0,\,500.0]}  \\
\bottomrule
\end{tabular}
\caption{Uncertainty sets used for domain randomization.}
\label{tab:Uncertainty_sets_DR}
\end{table}

\paragraph{Computational resources}
The training of AD-RRL and Polygrad was performed on three different machines. On a cluster node with one A100 GPU, Icelake CPU and 256 GB of RAM.

The remaining model-free baselines were trained on a laptop with an Intel i7-1185G7 CPU, Mesa Intel Xe Graphics GPU and 32 GB of RAM.

In table \ref{tab:training_time} we report the wall-clock training time for each algorithm . As it is expected, the model-based algorithms (AD-RRL and PolyGRAD) are slower than the model-free ones. This is a well-known shortcoming of Model-Based RL methods, even more so when using Diffusion Models, known for their longer training times when compared to standard MLPs. AD-RRL is slower than PolyGRAD since it employes an additional Diffusion Model to approximate the cumulative reward of a trajectory.

\begin{table}[h!]
\centering
%       Alg. | Hop | Halfch | Walk | InvPend | Reacher
\begin{tabular}{@{}lccccc@{}}
\toprule
\textbf{Algorithm} &
\textbf{Hopper} &
\textbf{Halfcheetah} &
\textbf{Walker} &
\textbf{InvertedPendulum} &
\textbf{Reacher} \\ \midrule
AD\textminus RRL (ours)$^\dagger$ & 3-20-00 & 3-20-00 & 3-20-00 & 3-20-00 & 3-20-00 \\
Polygrad$^\dagger$                & 2-14-00 & 2-14-00 & 2-14-00 & 2-14-00 & 2-14-00 \\
M2TD3$^\ddagger$                   & 0-02-00 & 0-03-30 & 0-02-45 & 0-02-45 & 0-02-45 \\
CPPO$^\ddagger$                      & 0-00-30 & 0-00-30 & 0-00-30 & 0-00-30 & 0-00-30 \\
PPO$^\ddagger$                      & 0-00-30 & 0-00-30 & 0-00-30 & 0-00-30 & 0-00-30 \\
TRPO$^\ddagger$                     & 0-00-30 & 0-00-30 & 0-00-30 & 0-00-30 & 0-00-30 \\
DR\textminus PPO$^\ddagger$         & 0-00-30 & 0-00-30 & 0-00-30 & 0-00-30 & 0-00-30 \\ \bottomrule
\end{tabular}
\caption{Wall-clock training time (days–hours–minutes) needed to reach the reported performance on the MuJoCo tasks. Times are rounded up to the nearest quarter hour. $^\dagger$Trained on cluster node. $^\ddagger$Trained on laptop.}
\label{tab:training_time}
\end{table}

\newpage

\section{Additional Results}
\label{appendix:additional_results}
In this section, we provide additional plots to support our conclusions. 

\subsection{Training curves}
\label{appendix:training_curves}

\begin{figure}[H]
  \centering
  % ---------- Row 1 ----------
  \begin{subfigure}[b]{0.48\linewidth}
    \centering
    \includegraphics[width=\linewidth]{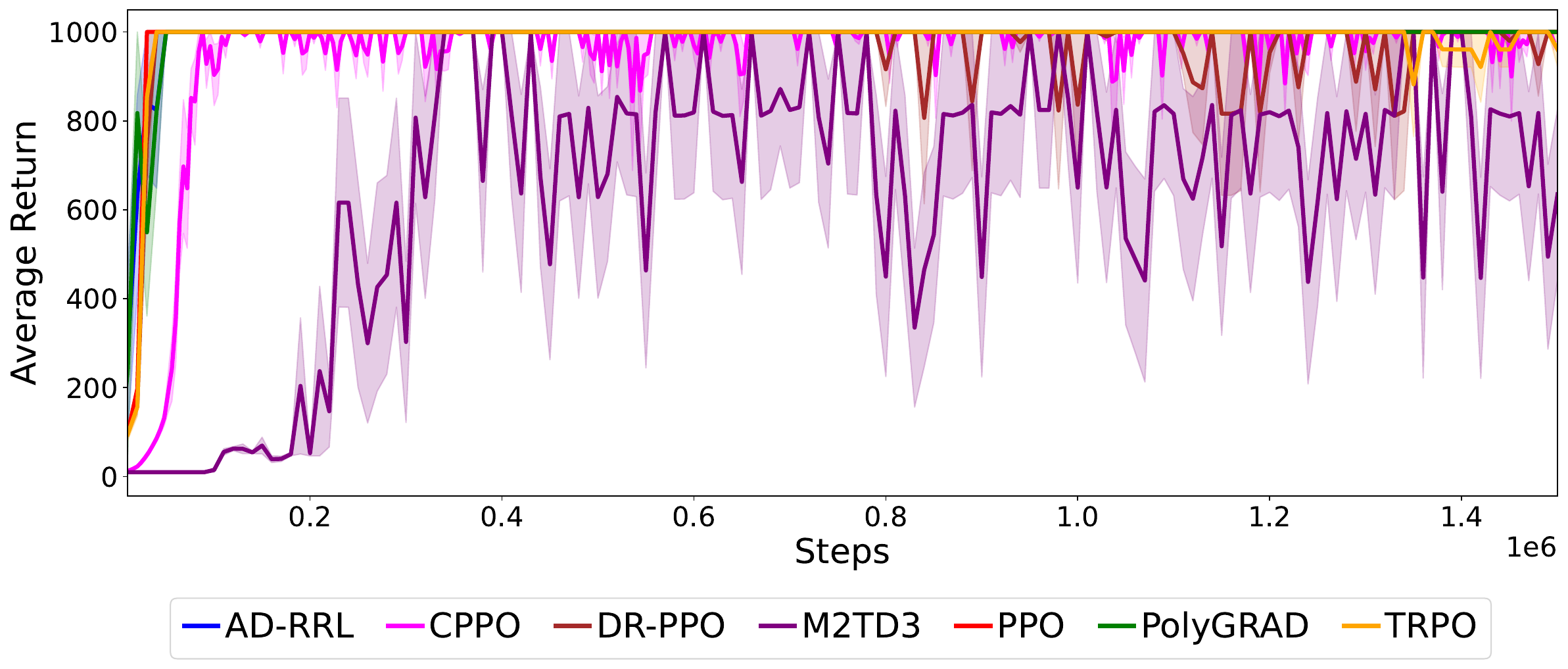}
    \subcaption{InvertedPendulum}\label{fig:training_pendulum}
  \end{subfigure}\hfill
  \begin{subfigure}[b]{0.48\linewidth}
    \centering
    \includegraphics[width=\linewidth]{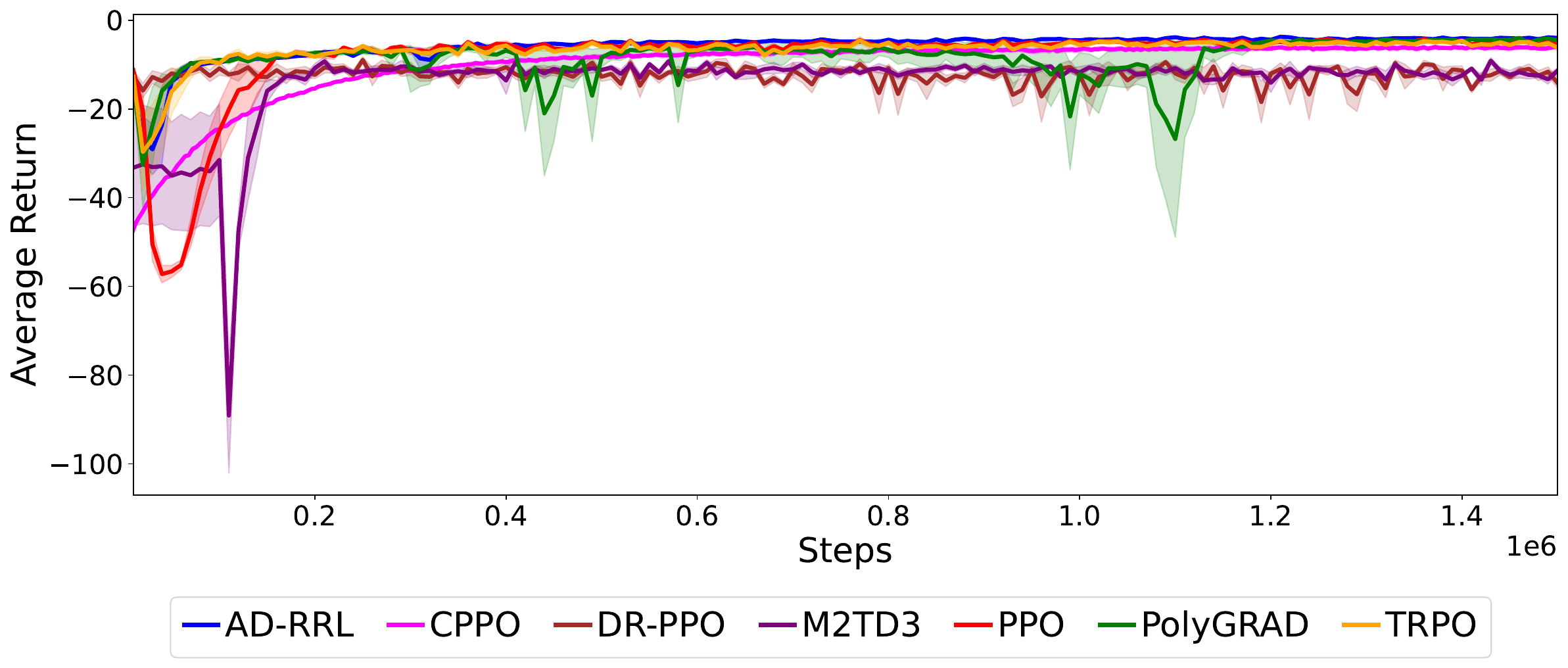}
    \subcaption{Reacher}\label{fig:training_reacher}
  \end{subfigure}

  \vspace{0.3\baselineskip}

  % ---------- Row 2 ----------
  \begin{subfigure}[b]{0.48\linewidth}
    \centering
    \includegraphics[width=\linewidth]{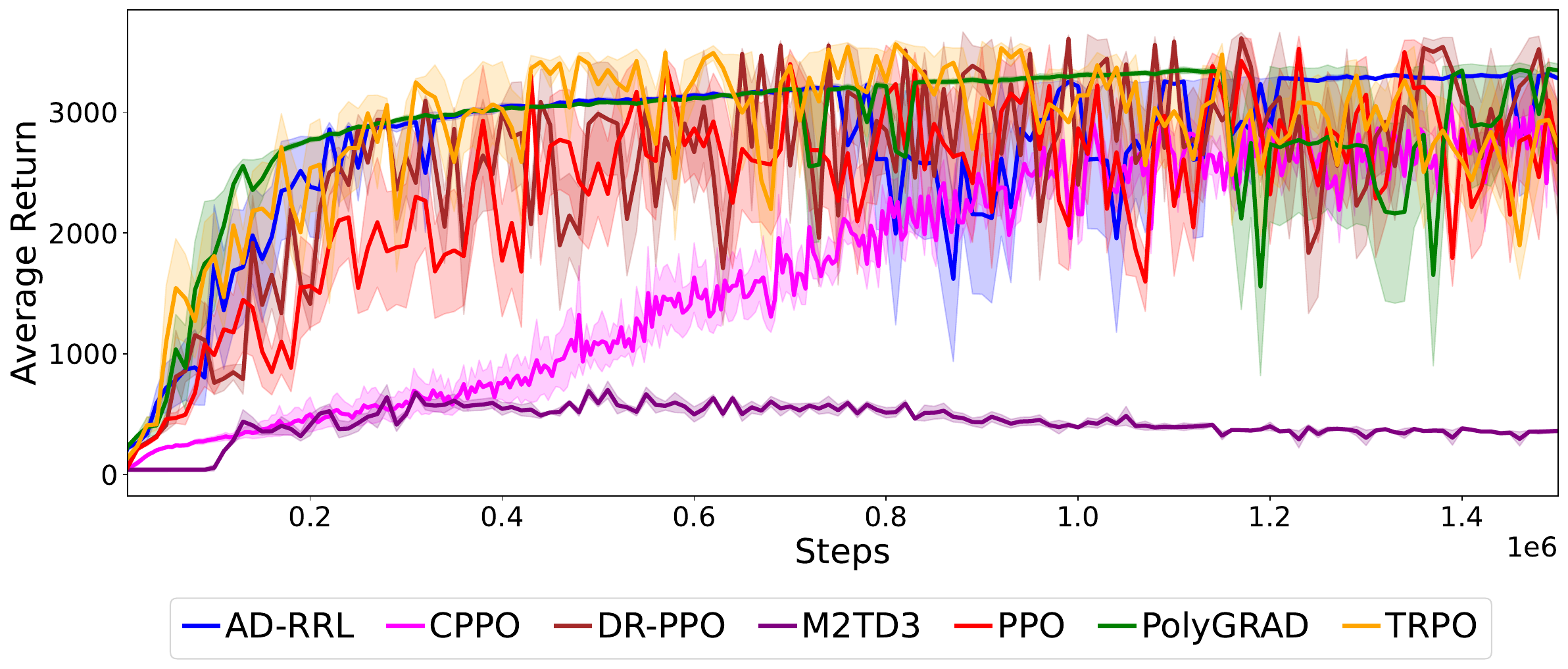}
    \subcaption{Hopper}\label{fig:training_hopper}
  \end{subfigure}\hfill
  \begin{subfigure}[b]{0.48\linewidth}
    \centering
    \includegraphics[width=\linewidth]{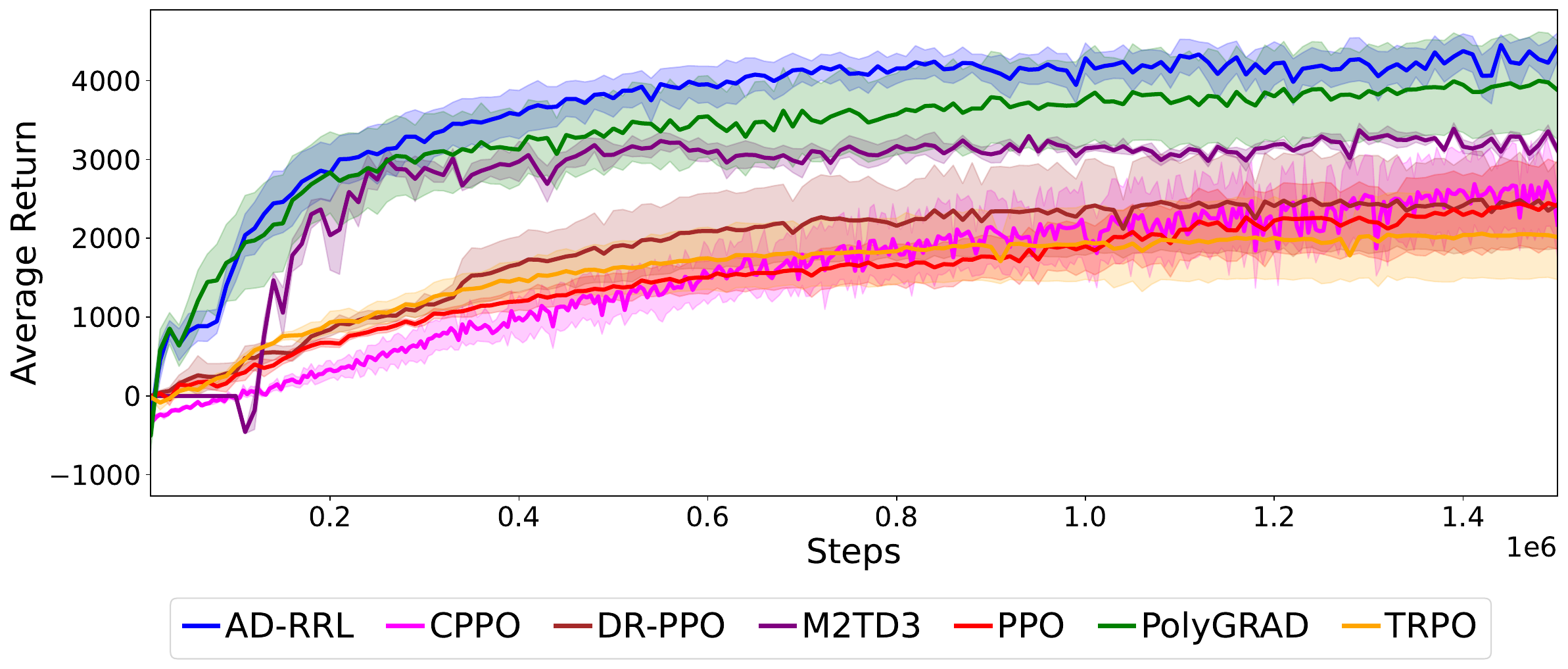}
    \subcaption{Cheetah}\label{fig:training_cheetah}
  \end{subfigure}

  \vspace{0.3\baselineskip}

  % ---------- Row 3 ----------
  \begin{subfigure}[b]{0.48\linewidth}   % narrower than full width looks nicer
    \centering
    \includegraphics[width=\linewidth]{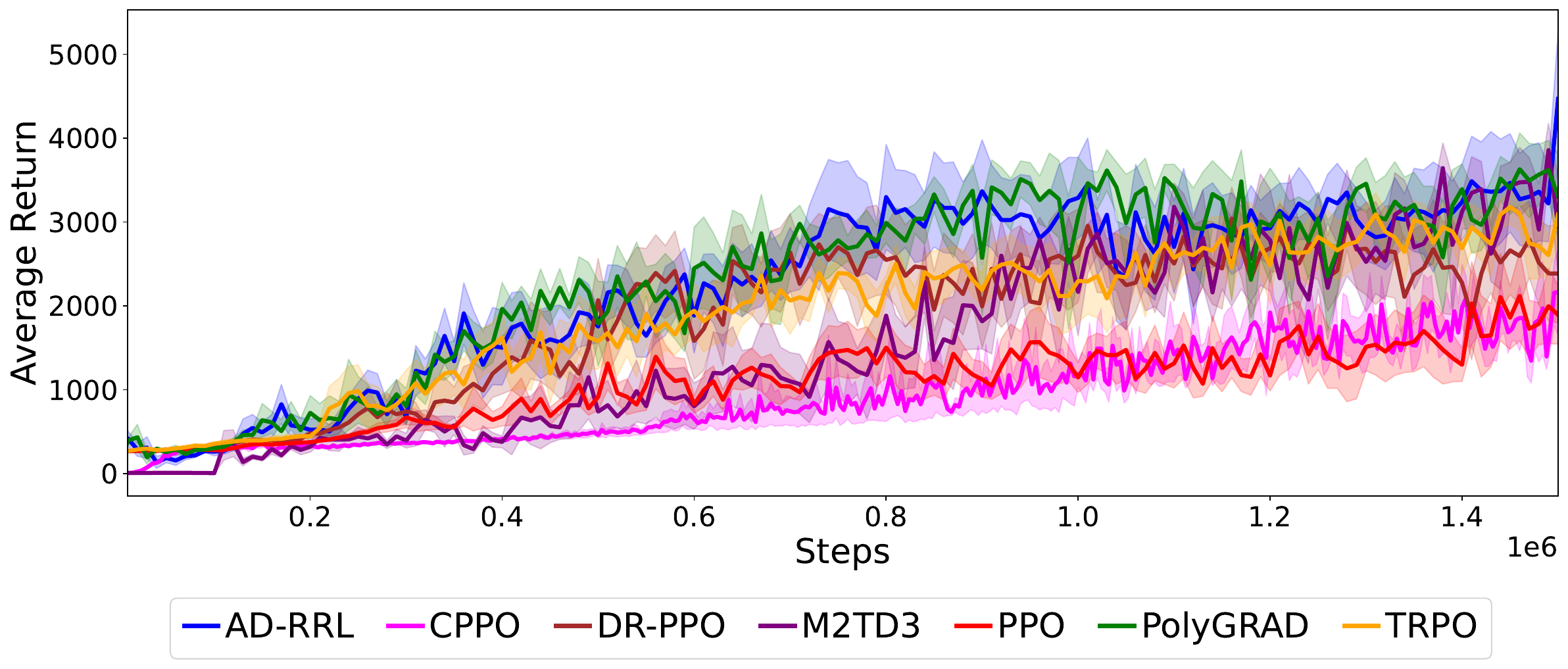}
    \subcaption{Walker}\label{fig:training_walker}
  \end{subfigure}

  \caption{Training-return curves on the nominal environment for five MuJoCo tasks.  % global caption
  Shaded areas represent one standard error over 5 runs.}
  \label{fig:training_curves}
\end{figure}

\cref{fig:training_curves} shows the learning curves
for AD-RRL and all the baselines for the considered MuJoCo tasks. Across seeds, AD-RRL reaches its final performance at least as quickly as the other methods. AD-RRL also achieves a final score matching or surpassing that of the baselines. This is particularly clear for the Cheetah and Walker environments, presented in \cref{fig:training_cheetah,fig:training_walker}. These results confirm that our method is more robust to modeling errors but does not sacrifice optimality in the training environment or learning speed.

% --------------------------------------

\newpage

\subsection{Varying parameters}
\begin{figure}[H]
  \centering

  \begin{subfigure}[b]{0.48\linewidth}
    \centering
    \includegraphics[width=\linewidth]{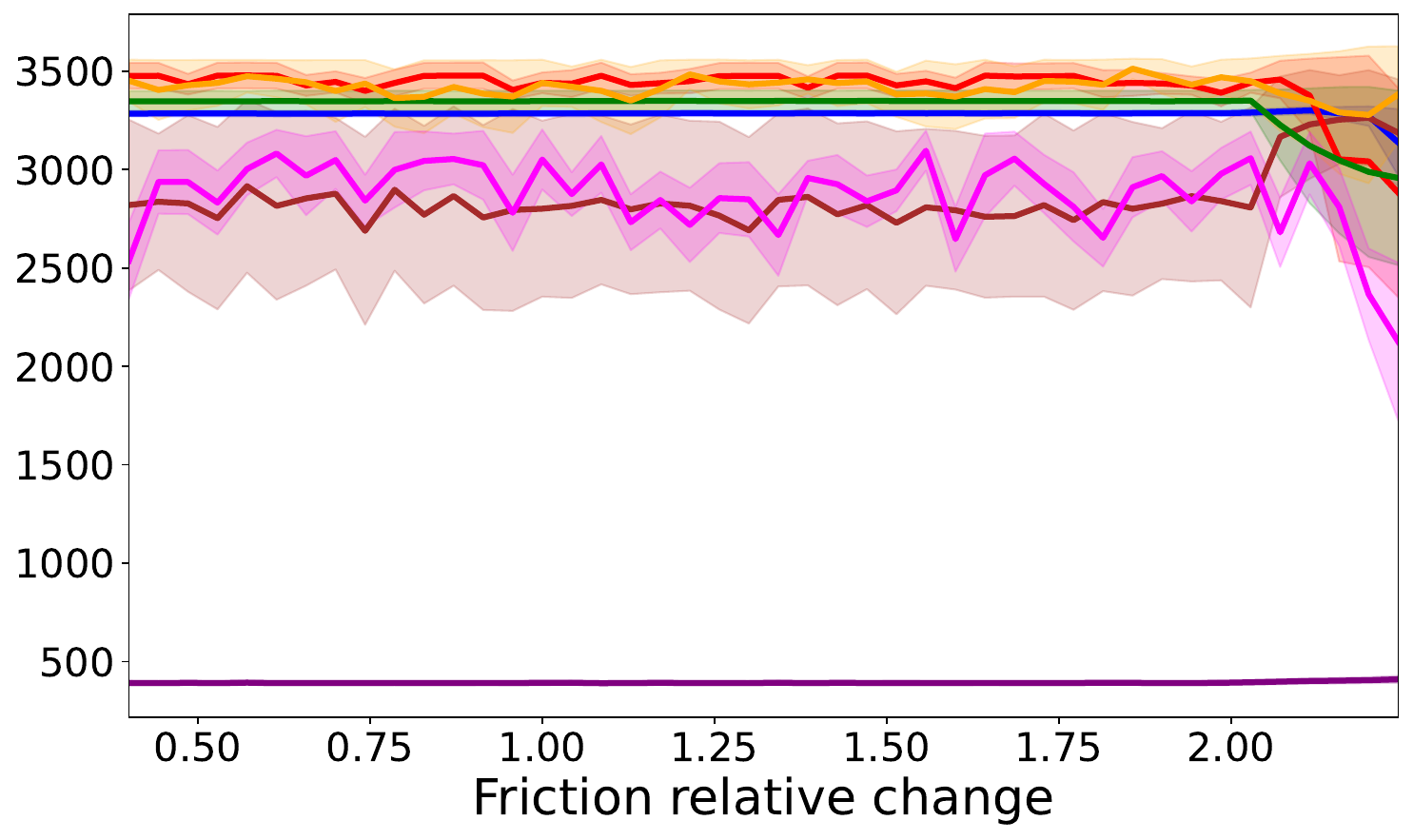}
    \subcaption{Hopper - Friction}\label{fig:hopper_friction}
  \end{subfigure}\hfill
  
  \vspace{0.3\baselineskip}

  % ---------- Row 3 ----------
  \begin{subfigure}[b]{0.48\linewidth}
    \centering
    \includegraphics[width=\linewidth]{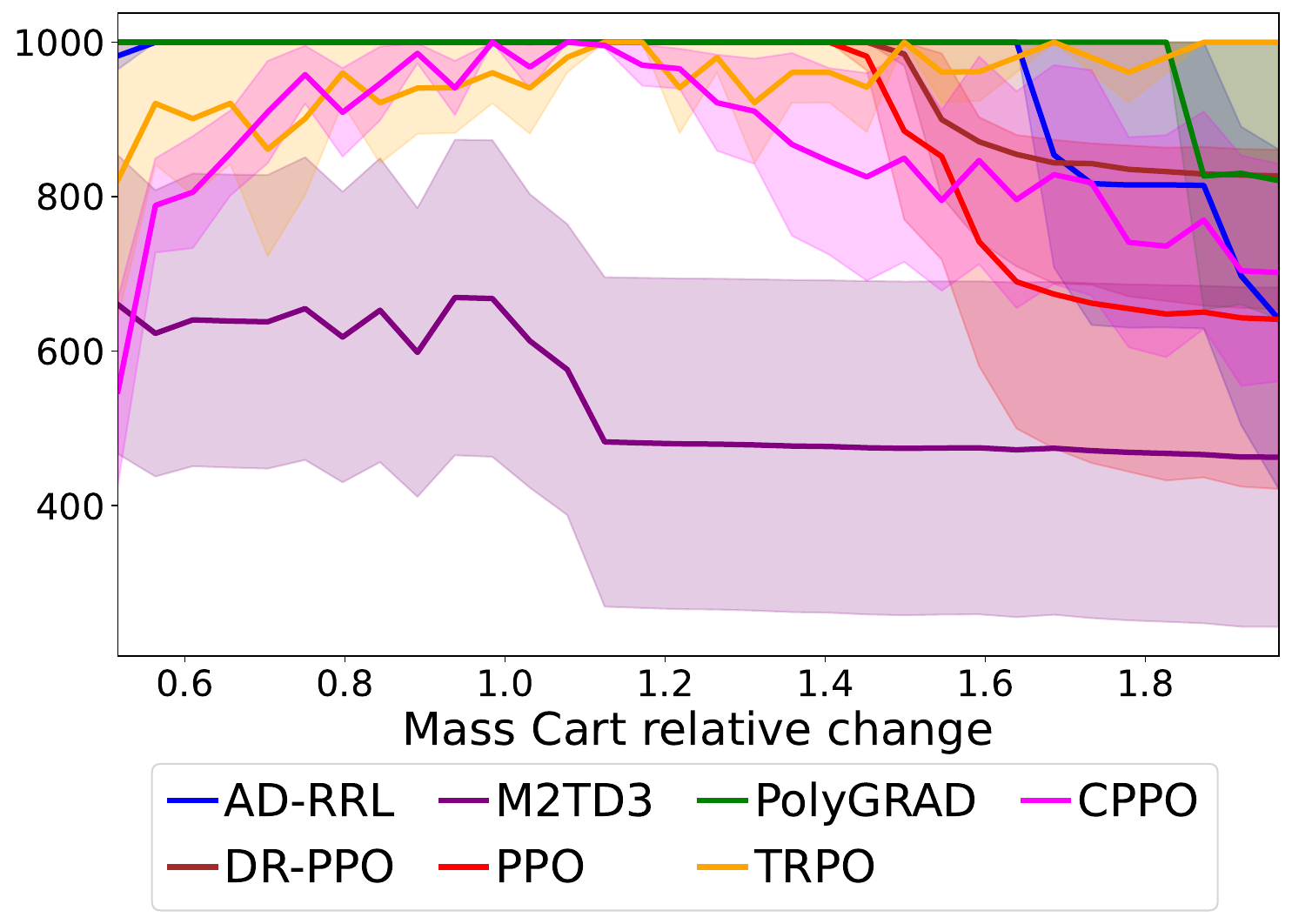}
    \subcaption{InvertedPendulum - Mass Cart}\label{fig:pendulum_masscart}
  \end{subfigure}\hfill
  \begin{subfigure}[b]{0.48\linewidth}   % narrower than full width looks nicer
    \centering
    \includegraphics[width=\linewidth]{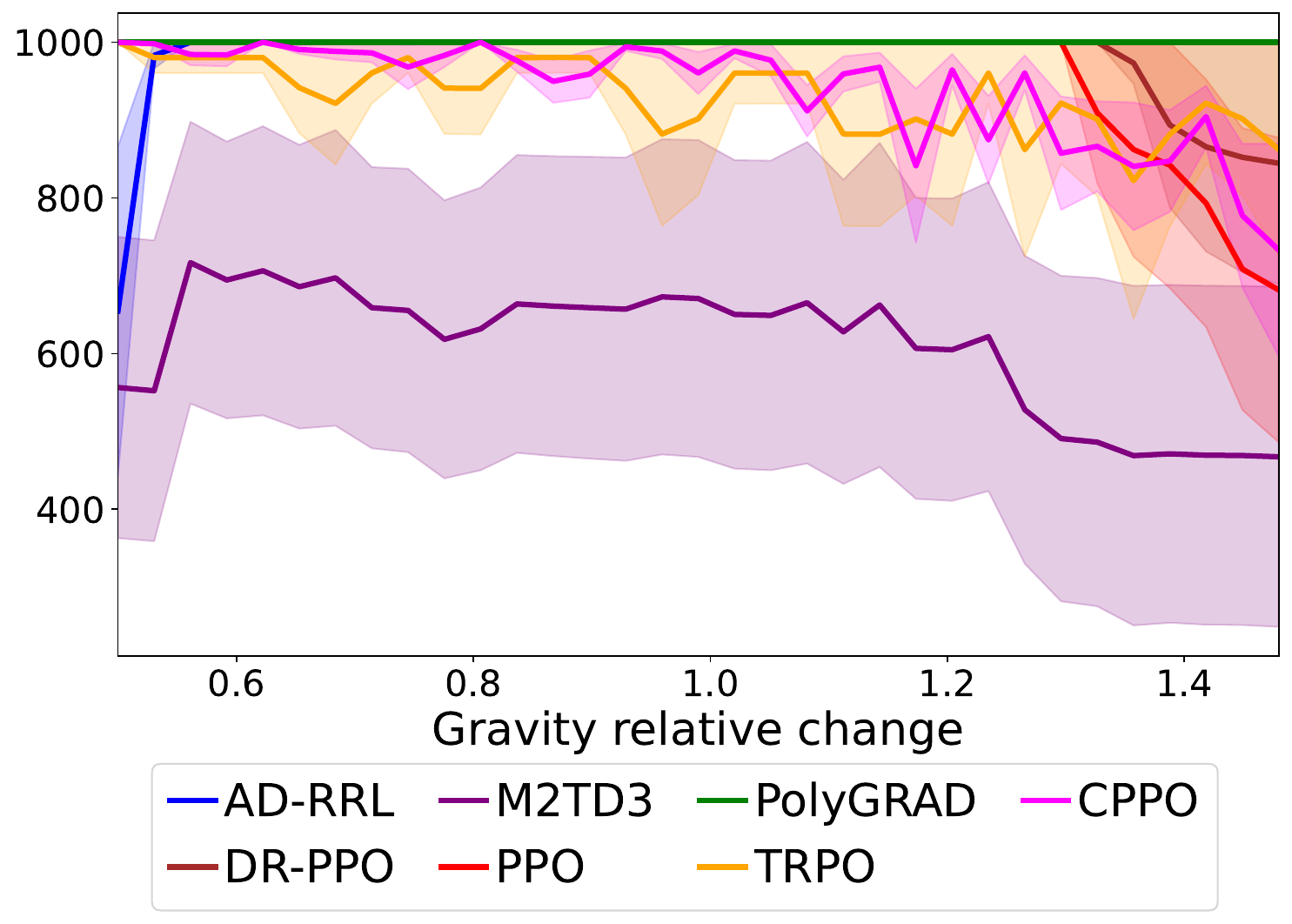}\subcaption{InvertedPendulum - Gravity}\label{fig:pendulum_gravity}
  \end{subfigure}

  \caption{Average Return for varying physical parameters. Shaded areas represent one standard error over 5 runs.}
  \label{fig:additional_results_parameters}
\end{figure}

In \cref{fig:additional_results_parameters} we present additional parameters variations for the InvertedPendulum and Cartpole environment. The pattern is consistent with the plots presented in \cref{fig:robust_plots}: AD-RRL achieves on par or higher returns than both robust and non-robust
baselines as the dynamics deviate from nominal values. The only exception appears to be for higher variations of the cart mass, in the Inverted Pendulum environment (\cref{fig:pendulum_masscart}), where the additional inertia pushes most methods toward failure and AD-RRL similarly shows a performance decline.
\label{appendix:varying_parameters}

\subsection{Ablation study}
\label{sec:ablation_study}
\begin{figure}[H]
  \centering
  % ---------- Row 1 ----------
  \begin{subfigure}[b]{0.48\linewidth}
    \centering
    \includegraphics[width=\linewidth]{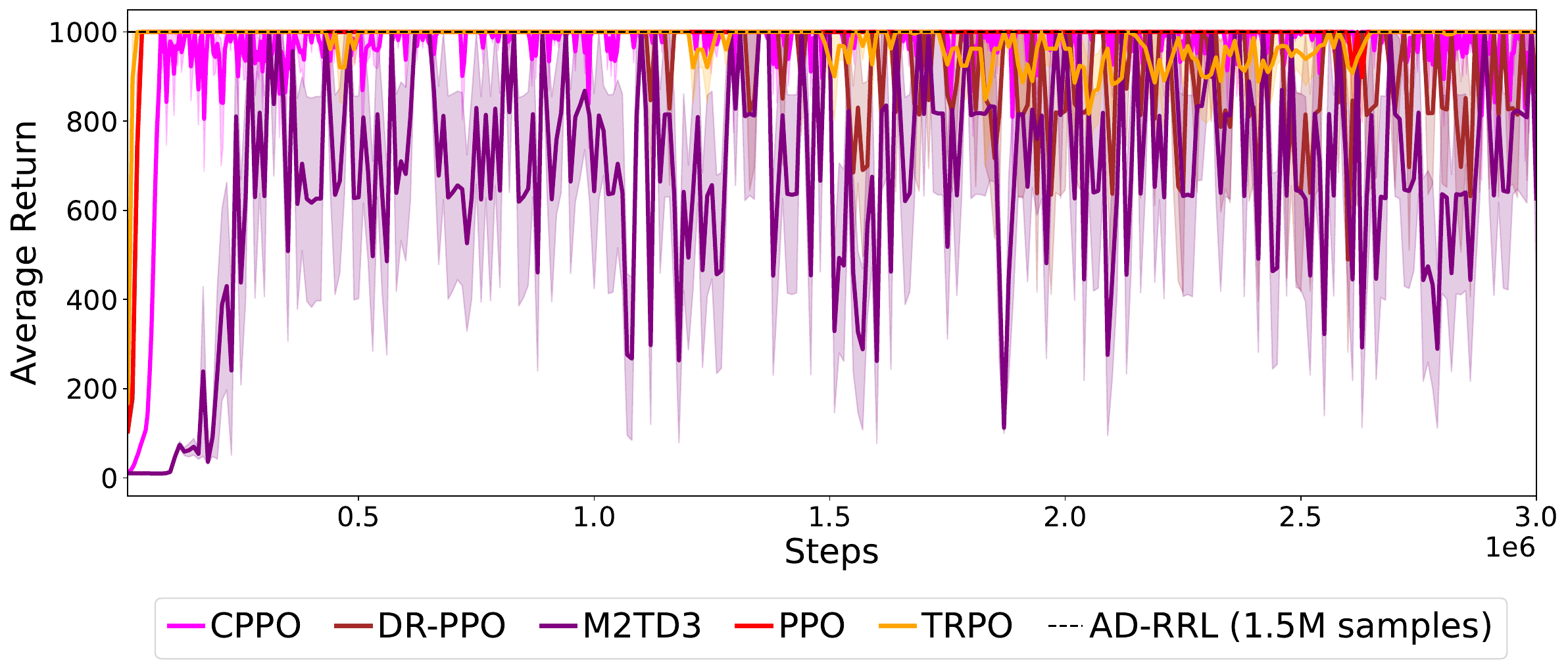}
    \subcaption{InvertedPendulum}\label{fig:training_pendulum3m}
  \end{subfigure}\hfill
  \begin{subfigure}[b]{0.48\linewidth}
    \centering
    \includegraphics[width=\linewidth]{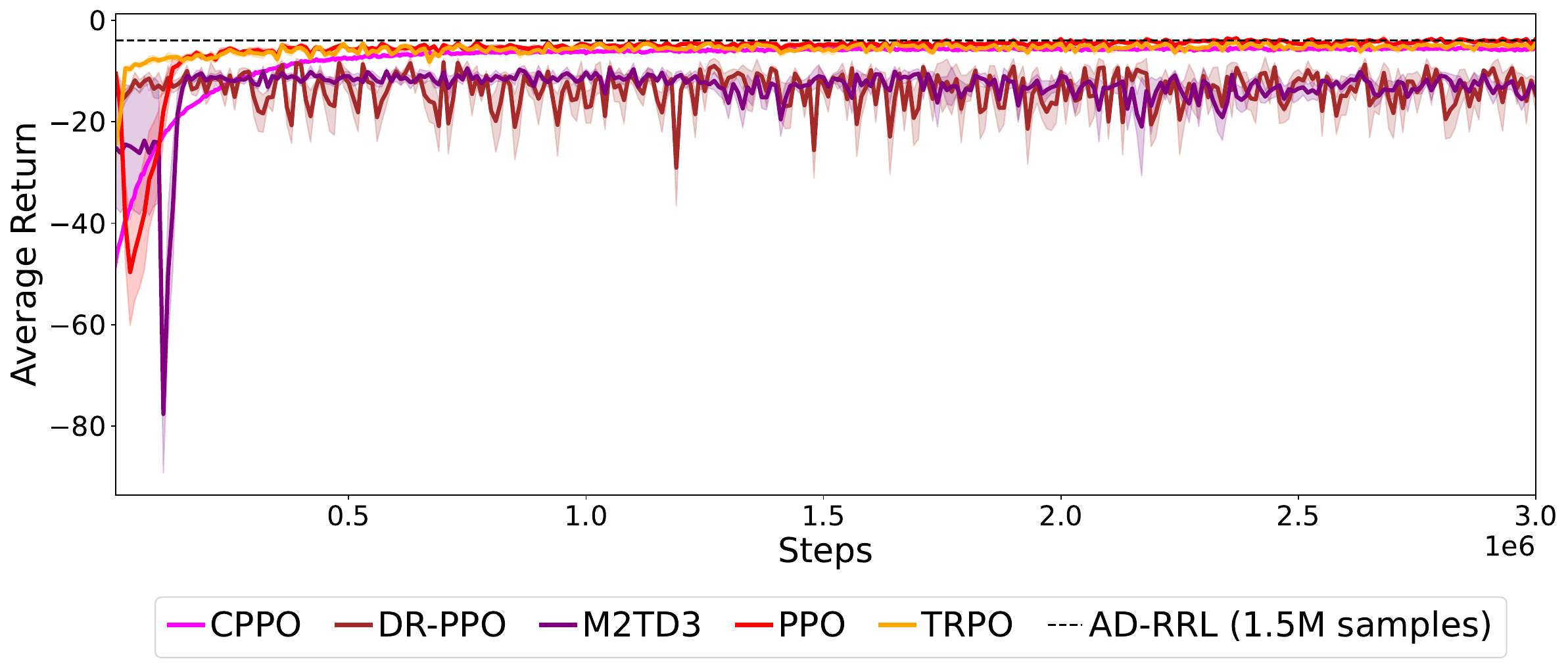}
    \subcaption{Reacher}\label{fig:training_reacher3m}
  \end{subfigure}

  \vspace{0.3\baselineskip}

  % ---------- Row 2 ----------
  \begin{subfigure}[b]{0.48\linewidth}
    \centering
    \includegraphics[width=\linewidth]{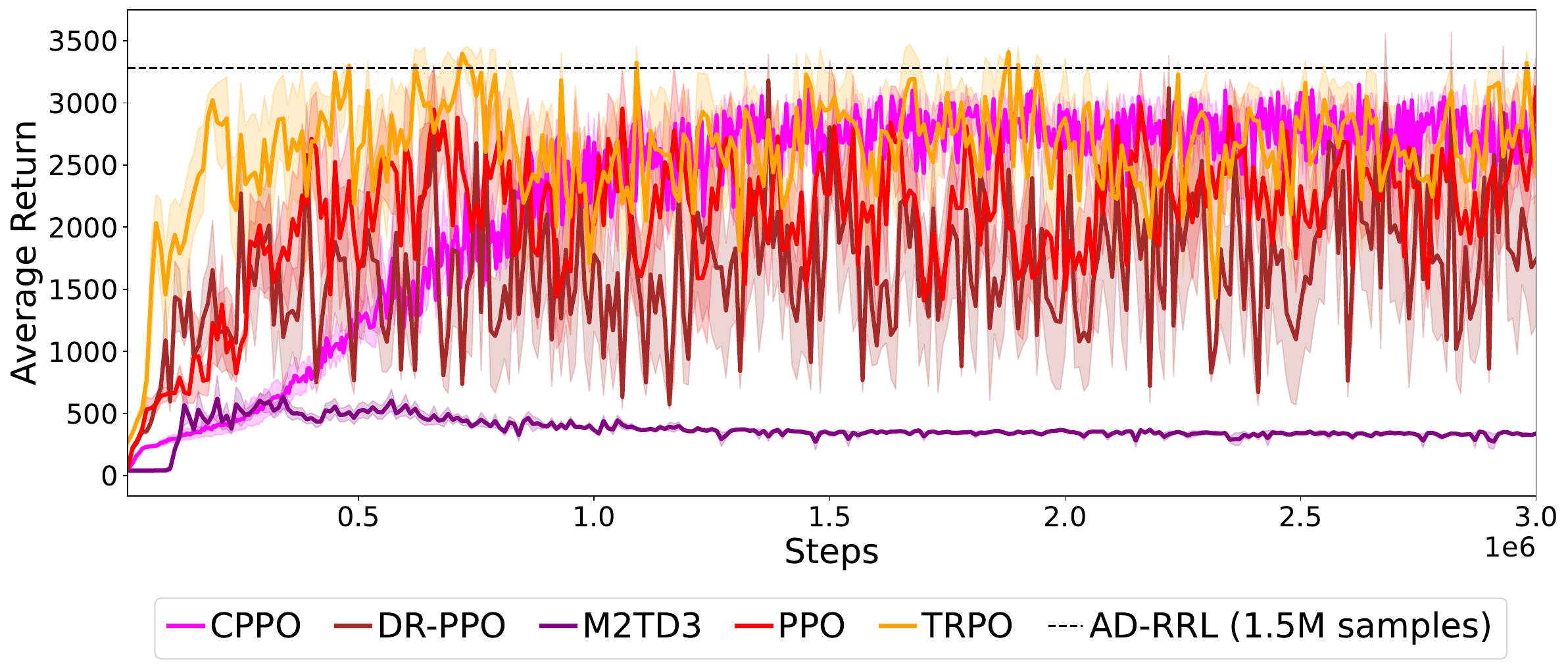}
    \subcaption{Hopper}\label{fig:training_hopper3m}
  \end{subfigure}\hfill
  \begin{subfigure}[b]{0.48\linewidth}
    \centering
    \includegraphics[width=\linewidth]{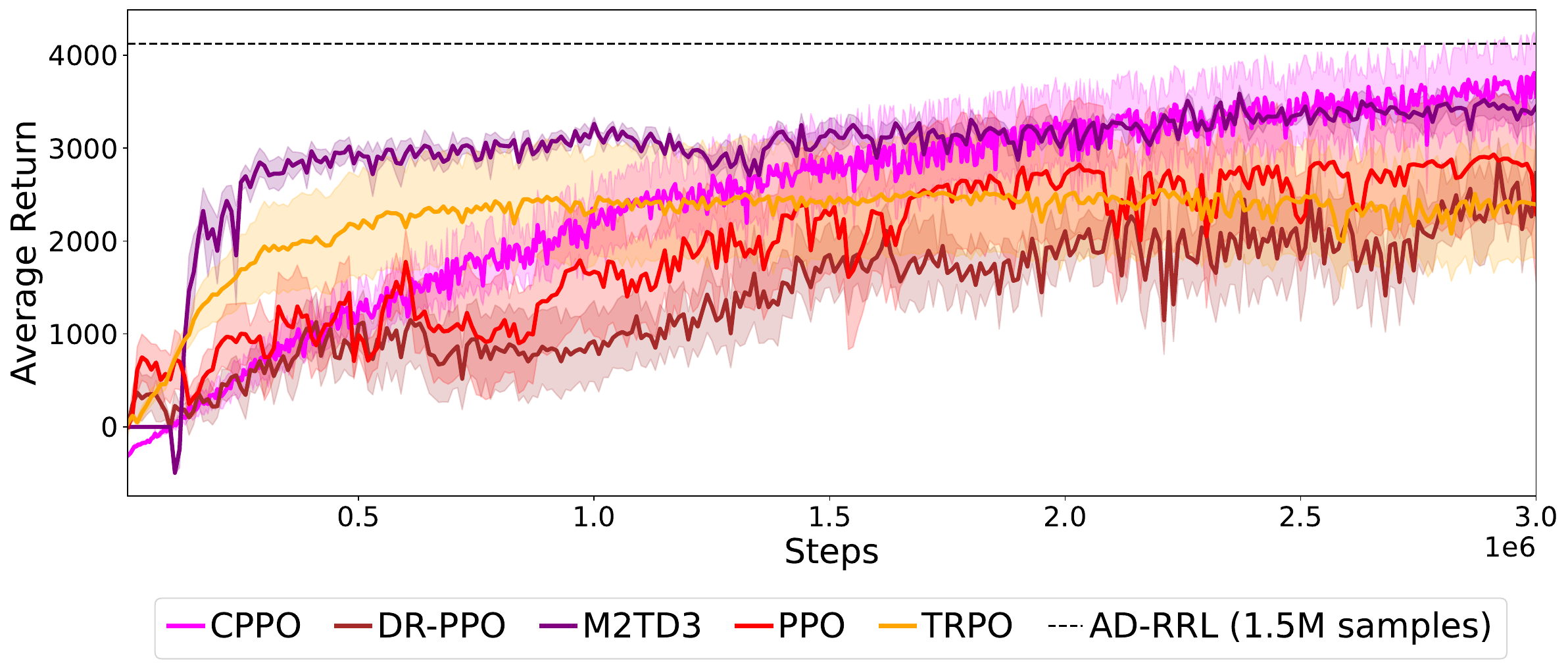}
    \subcaption{Cheetah}\label{fig:training_cheetah3m}
  \end{subfigure}

  \vspace{0.3\baselineskip}

  % ---------- Row 3 ----------
  \begin{subfigure}[b]{0.48\linewidth}   % narrower than full width looks nicer
    \centering
    \includegraphics[width=\linewidth]{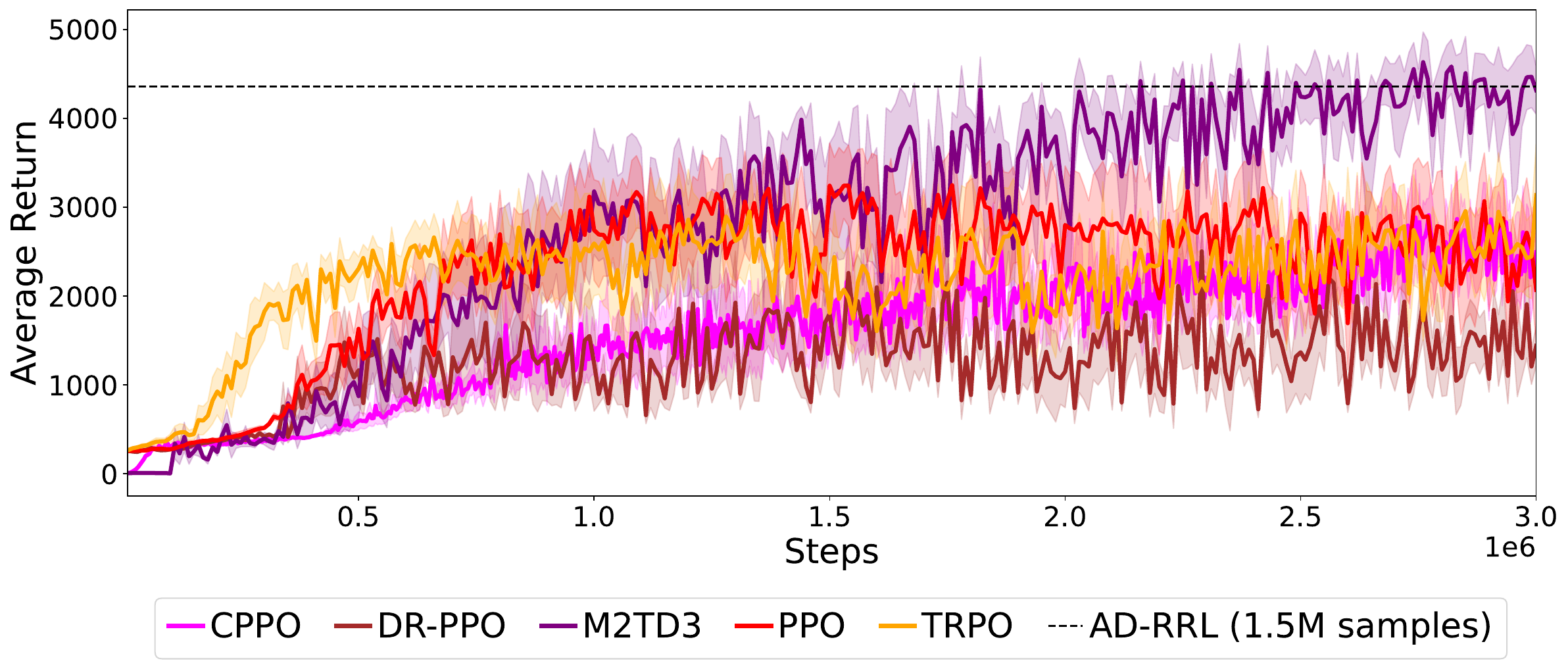}
    \subcaption{Walker}\label{fig:training_walker3m}
  \end{subfigure}

  \caption{Training-return curves on the nominal environment for the Model-Free baslines trained on $3$M samples.
  Shaded areas represent one standard error over 5 runs. The dashed line represents the reference final cumulative reward achieved by AD-RRL trained on $1.5$M samples.}
  \label{fig:training_curves3m}
\end{figure}

\begin{figure}[htbp]
  \centering
  \vspace{-0.4\baselineskip}
%============= Row 1 ================
  \begin{subfigure}[b]{0.32\linewidth}
    \centering
    \includegraphics[width=\linewidth]{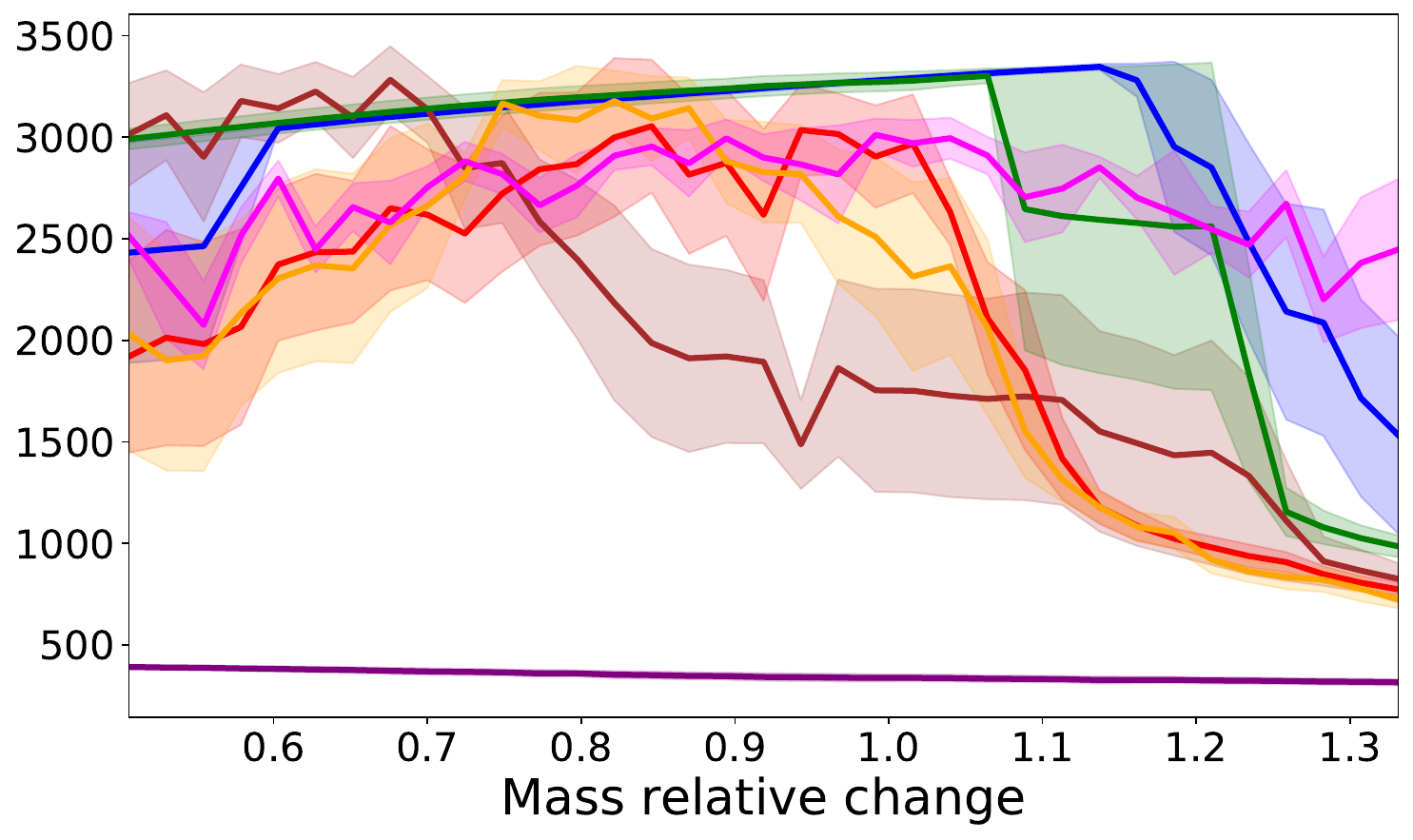}
    \subcaption{Hopper – Mass}\label{fig:Hopper_mass3M}
  \end{subfigure}\hfill
  \begin{subfigure}[b]{0.32\linewidth}
    \centering
    \includegraphics[width=\linewidth]{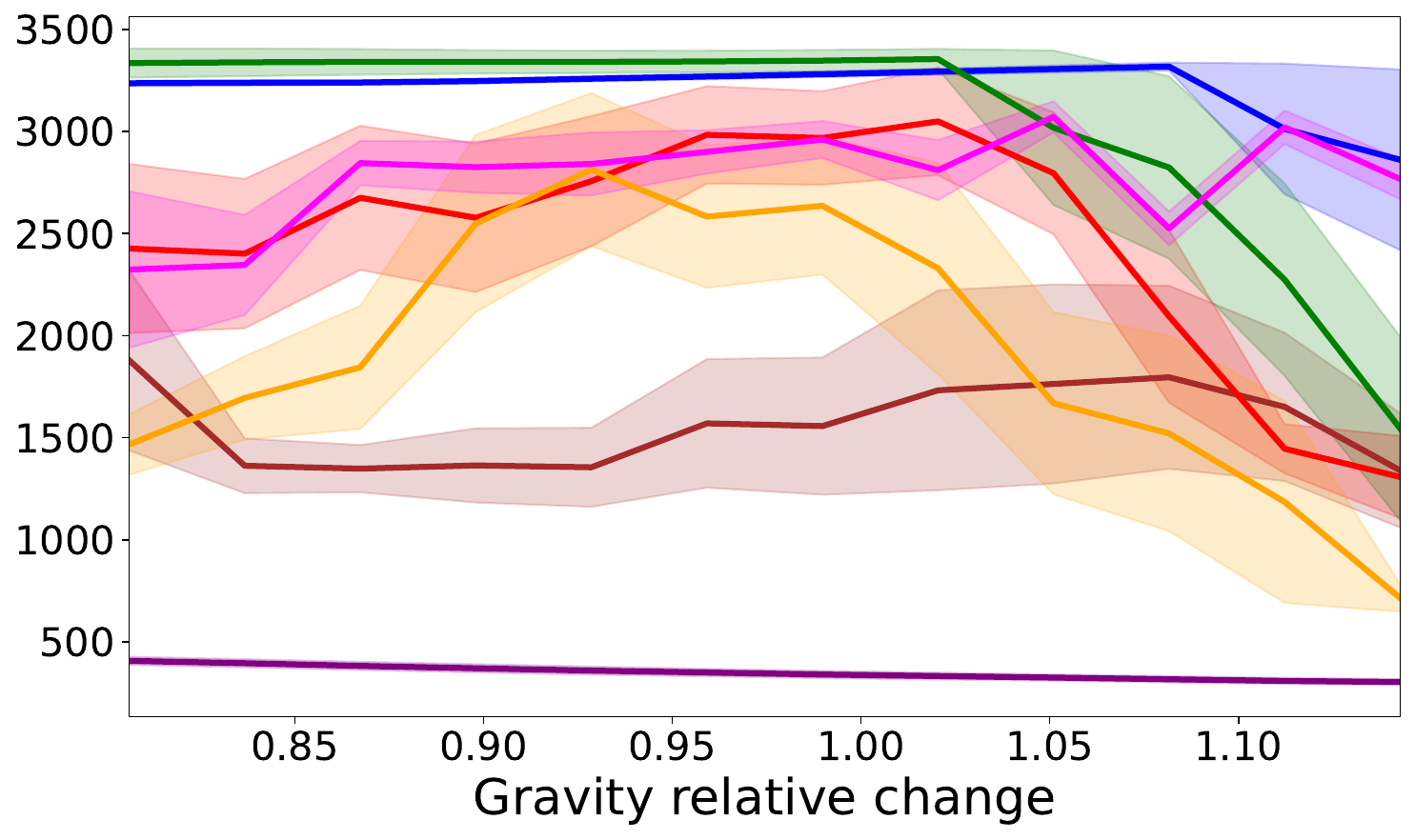}
    \subcaption{Hopper – Gravity}\label{fig:Hopper_gravity3M}
  \end{subfigure}\hfill
  \begin{subfigure}[b]{0.32\linewidth}
    \centering
    \includegraphics[width=\linewidth]{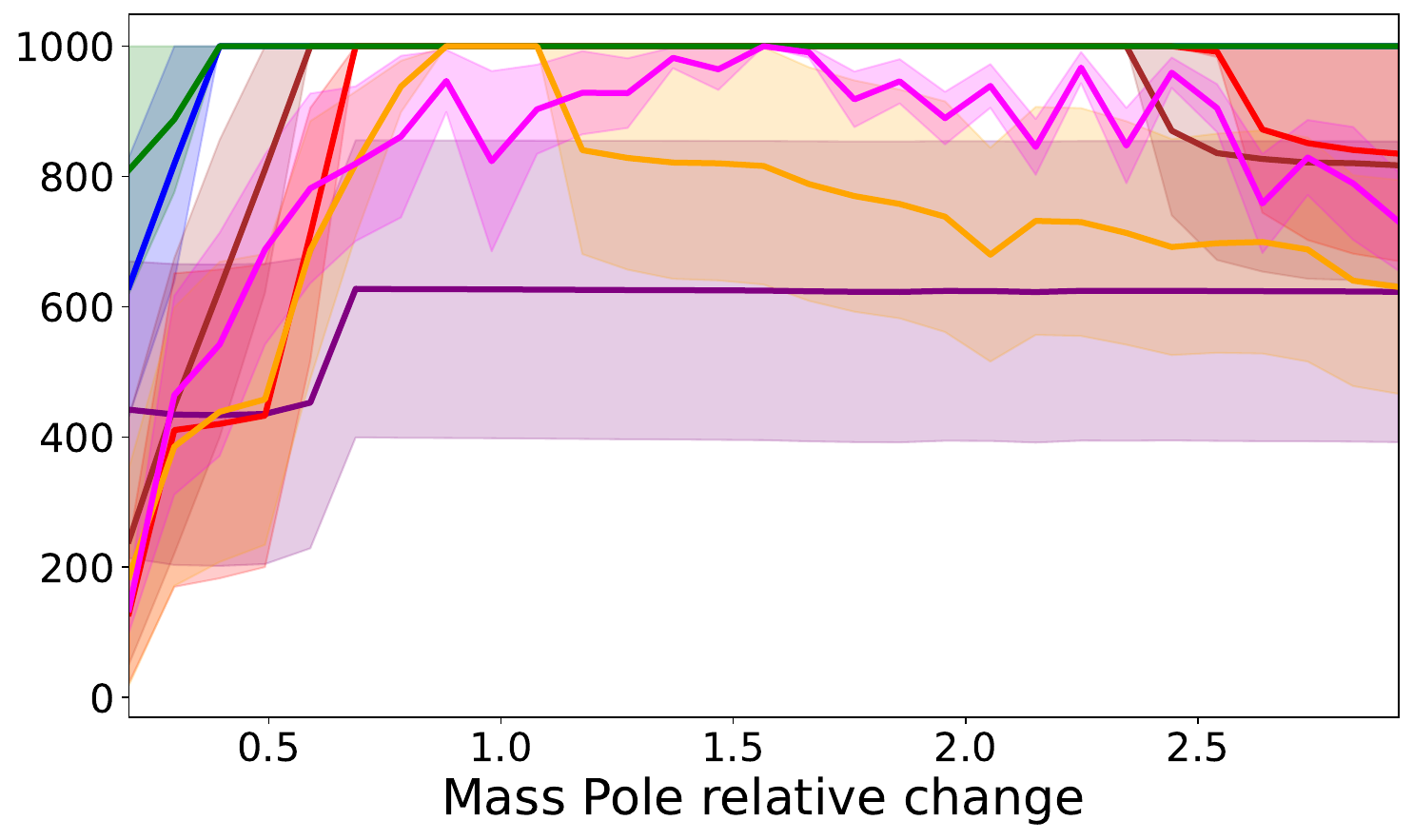}% <-- replace
    \subcaption{InvertedPendulum - Mass Pole}\label{fig:pendulum_masspole3M}
  \end{subfigure}

  \vspace{0.4\baselineskip}

%============= Row 2 ================
  \begin{subfigure}[b]{0.32\linewidth}
    \centering
    \includegraphics[width=\linewidth]{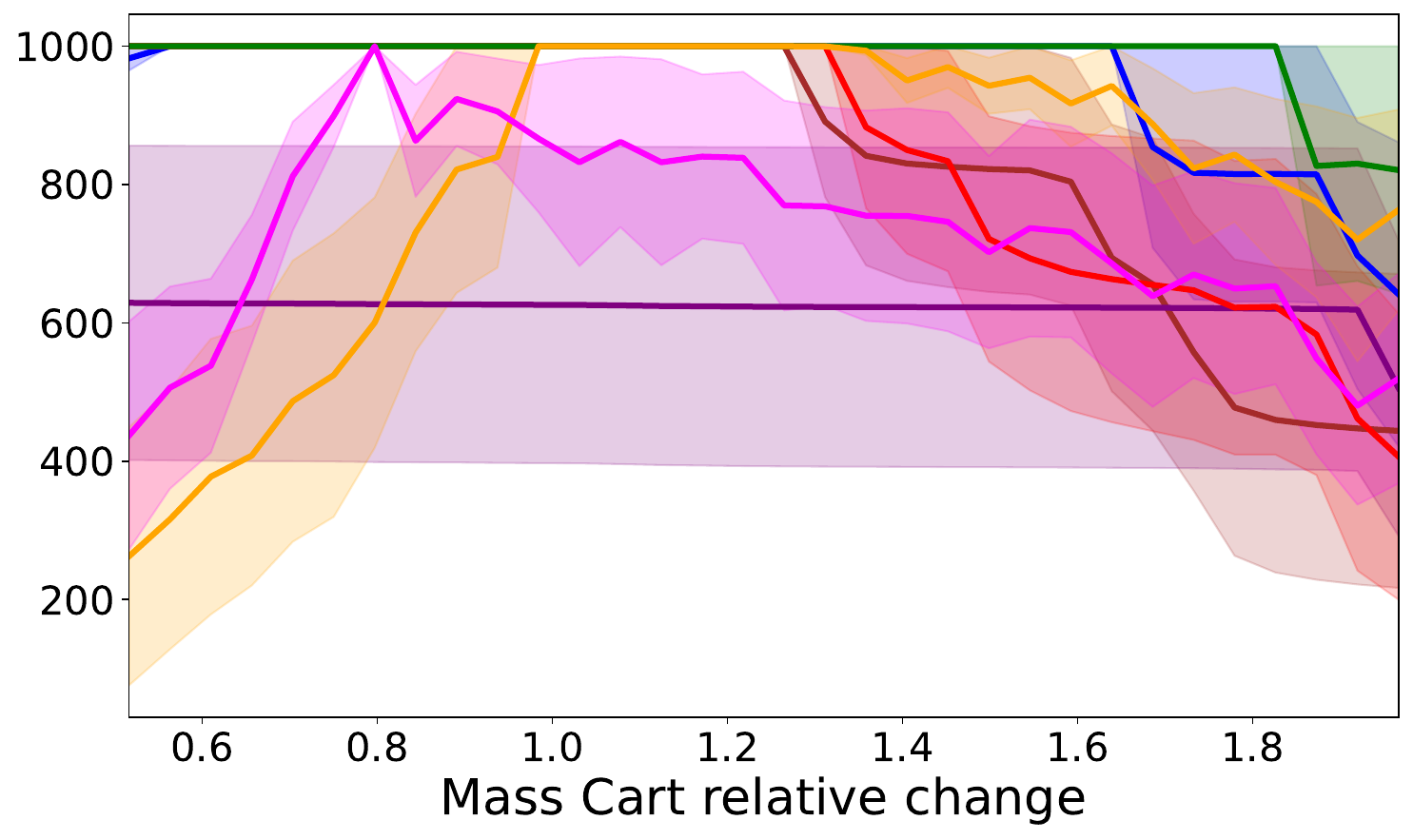}
    \subcaption{InvertedPendulum – Mass Cart}\label{fig:pendulum_mass_cart3m}
  \end{subfigure}\hfill
  \begin{subfigure}[b]{0.32\linewidth}
    \centering
    \includegraphics[width=\linewidth]{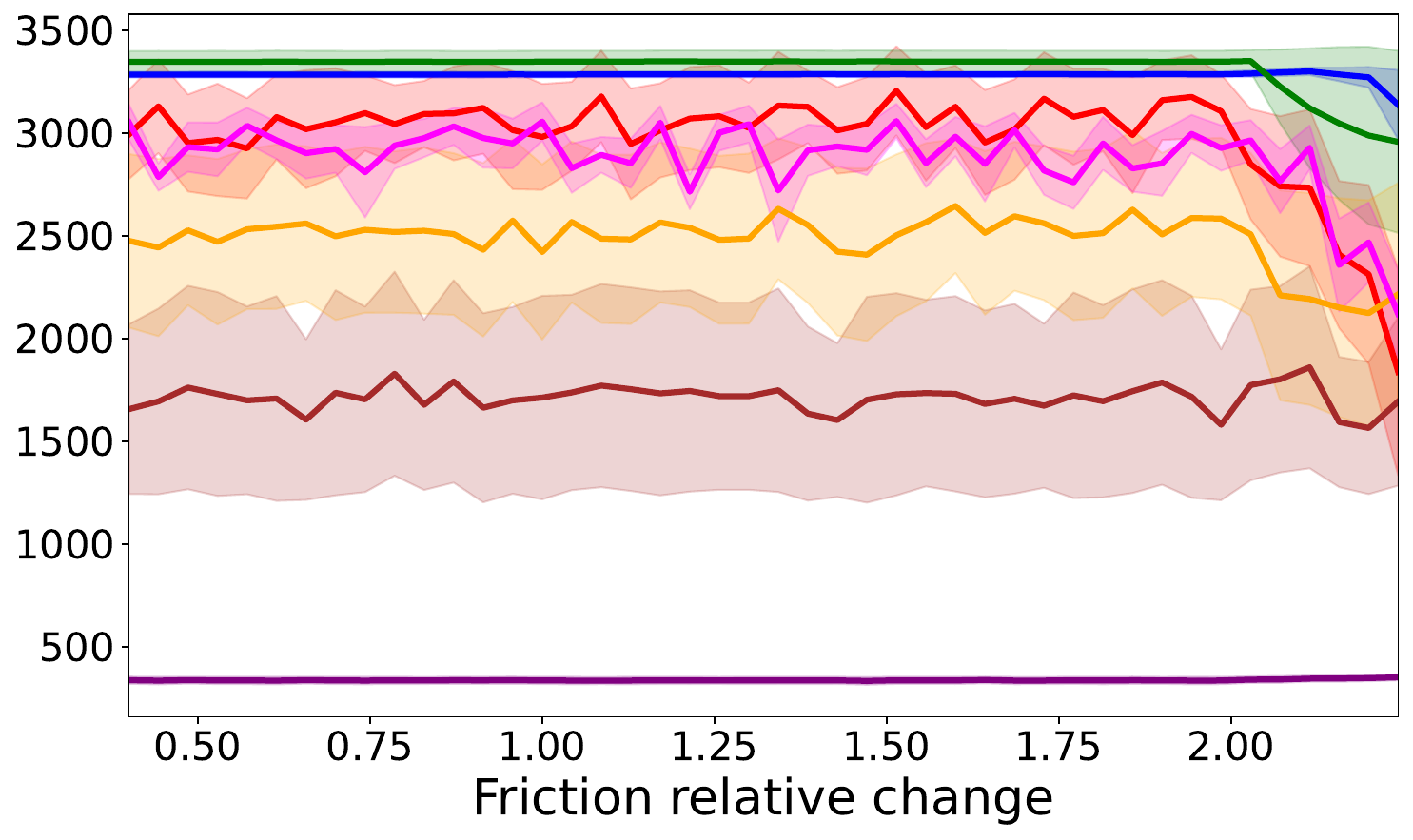}
    \subcaption{Hopper – Friction}\label{fig:Hopper_friction3m}
  \end{subfigure}\hfill
  \begin{subfigure}[b]{0.32\linewidth}
    \centering
    \includegraphics[width=\linewidth]{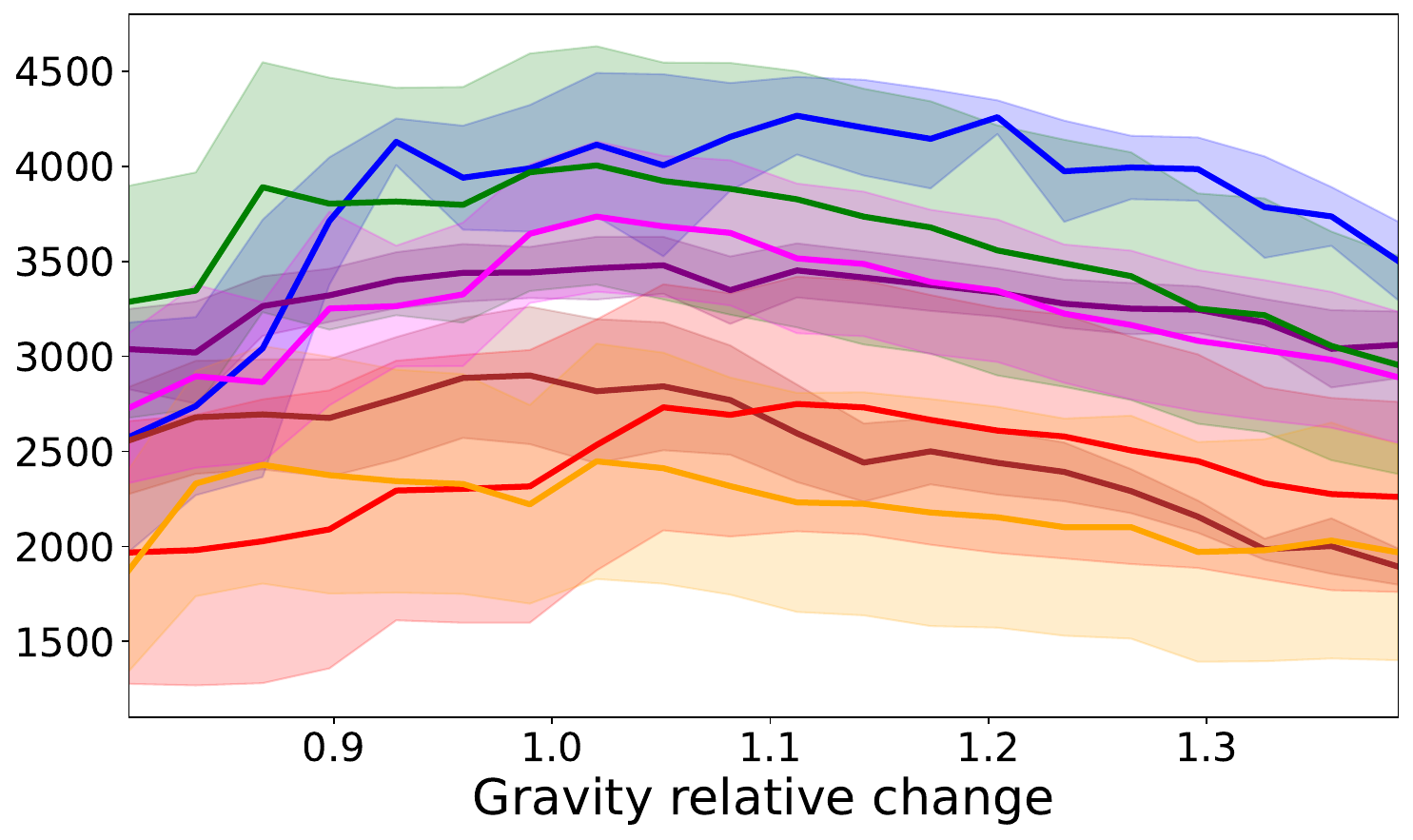}
    \subcaption{Cheetah - Gravity}\label{fig:cheetah_gravity3M}
  \end{subfigure}

  \vspace{0.4\baselineskip}

%============= Row 3 ================
  \begin{subfigure}[b]{0.32\linewidth}
    \centering
    \includegraphics[width=\linewidth]{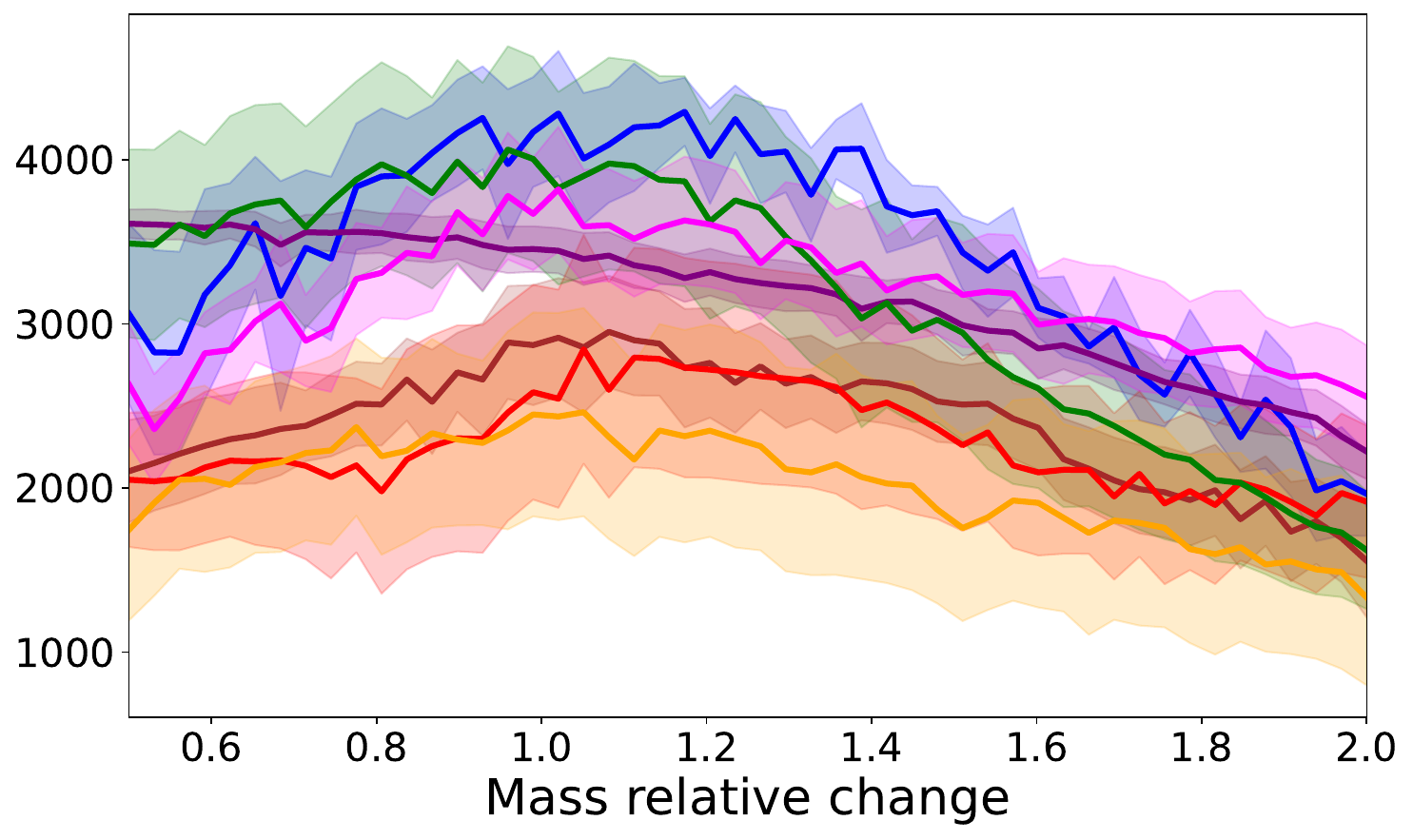}
    \subcaption{Cheetah – Mass}\label{fig:Cheetah_mass3m}
  \end{subfigure}\hfill
  \begin{subfigure}[b]{0.32\linewidth}
    \centering
    \includegraphics[width=\linewidth]{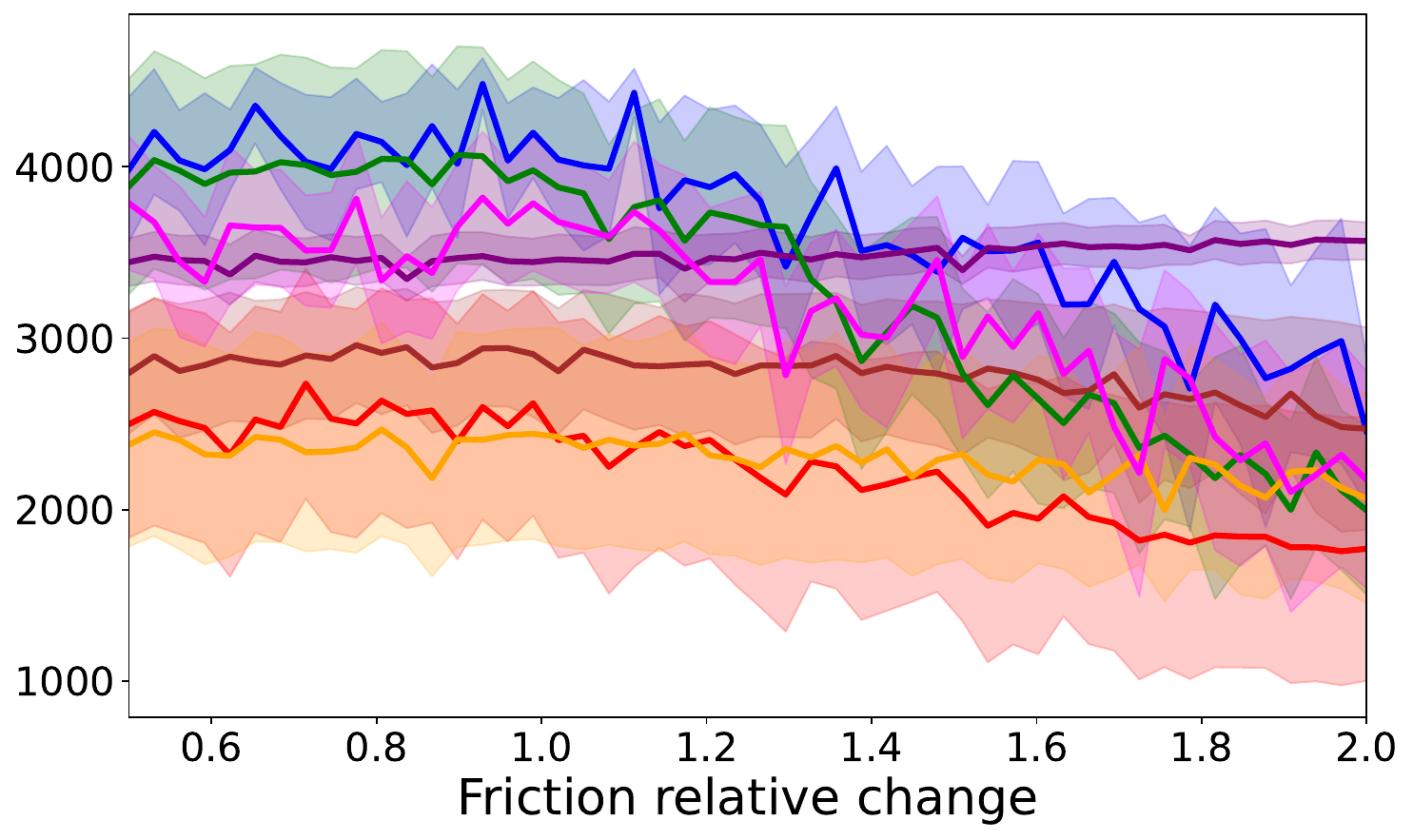}
    \subcaption{Cheetah – Friction}\label{fig:cheetah_friction3m}
  \end{subfigure}\hfill
  \begin{subfigure}[b]{0.32\linewidth}
    \centering
    \includegraphics[width=\linewidth]{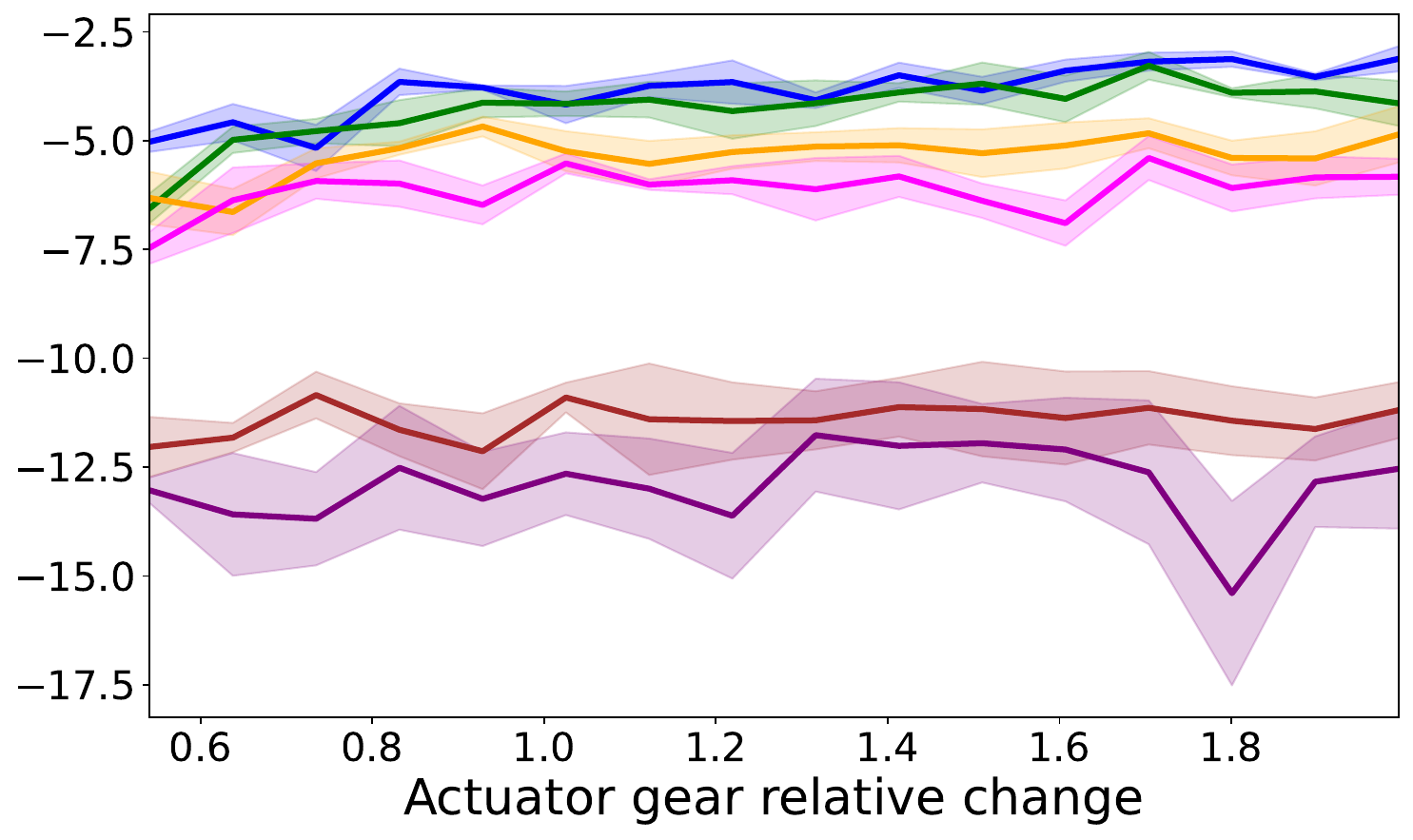}
    \subcaption{Reacher – Actuator Gear}\label{fig:Reacher_gear3m}
  \end{subfigure}

    \vspace{0.4\baselineskip}

%============= Row 4 ================
  \begin{subfigure}[b]{0.32\linewidth}
    \centering
    \includegraphics[width=\linewidth]{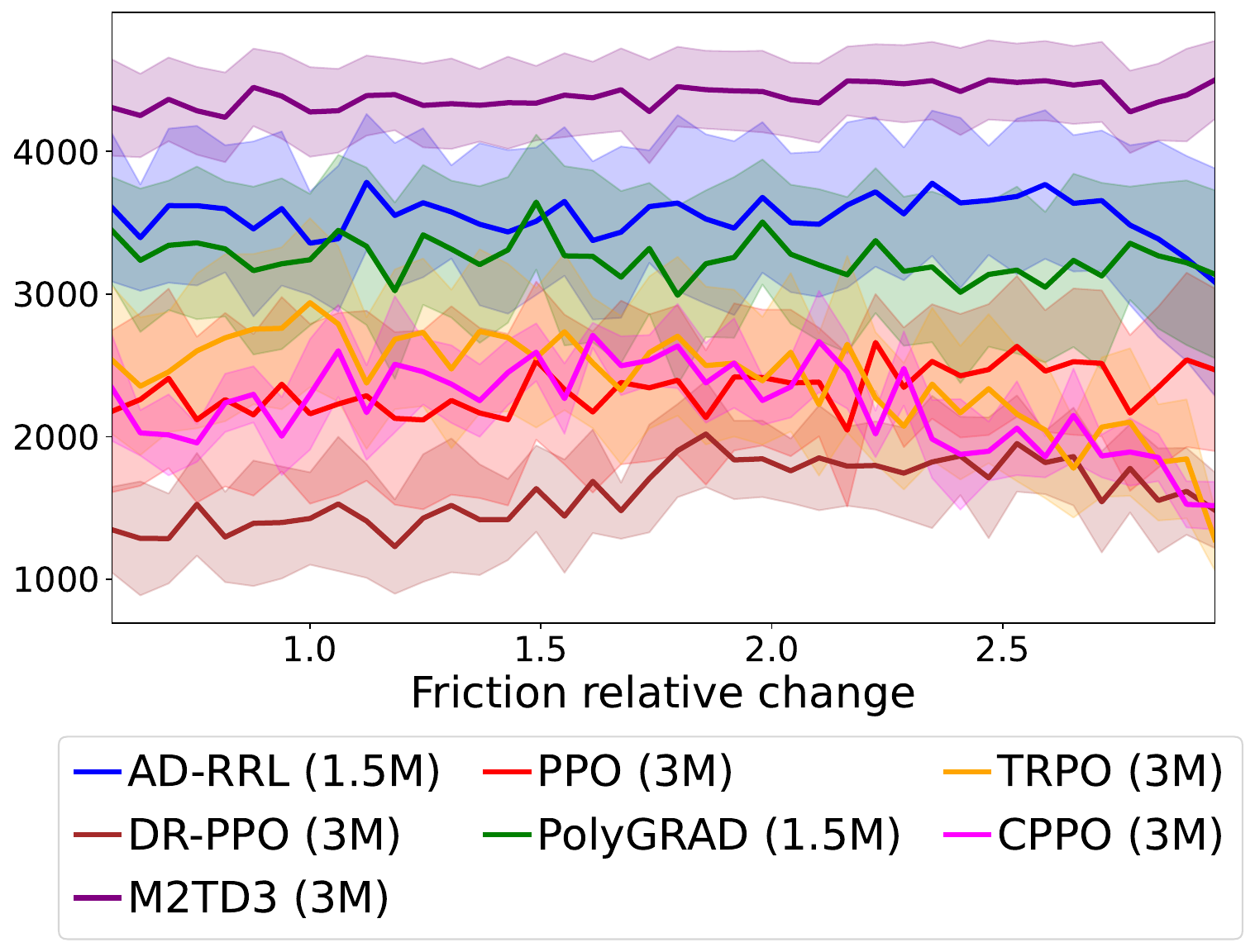}
    \subcaption{Walker – Friction}\label{fig:Walker_friction3M}
  \end{subfigure}\hfill
  \begin{subfigure}[b]{0.32\linewidth}
    \centering
    \includegraphics[width=\linewidth]{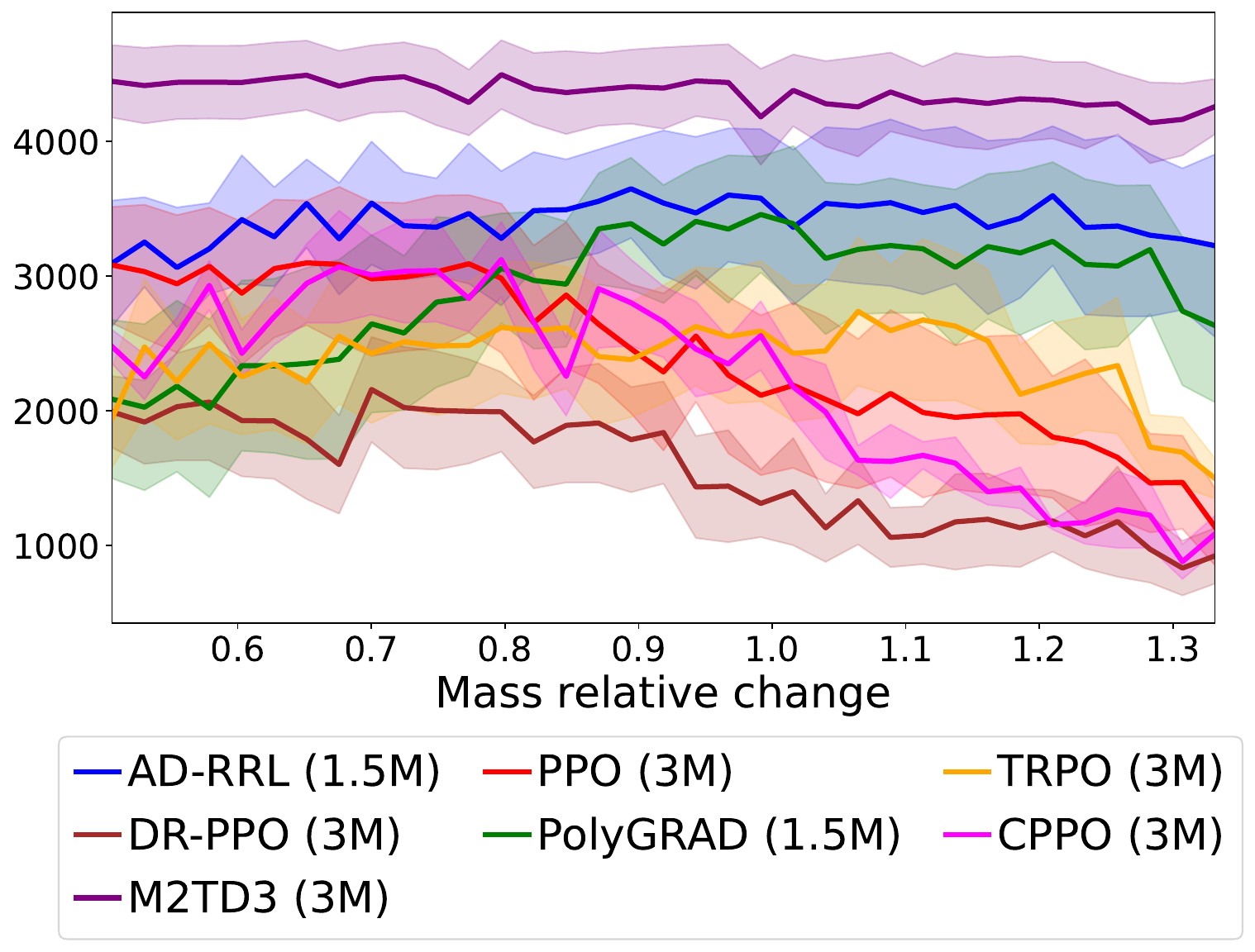}
    \subcaption{Walker – Mass}\label{fig:Walker_mass3M}
  \end{subfigure}\hfill
  \begin{subfigure}[b]{0.32\linewidth}
    \centering
    \includegraphics[width=\linewidth]{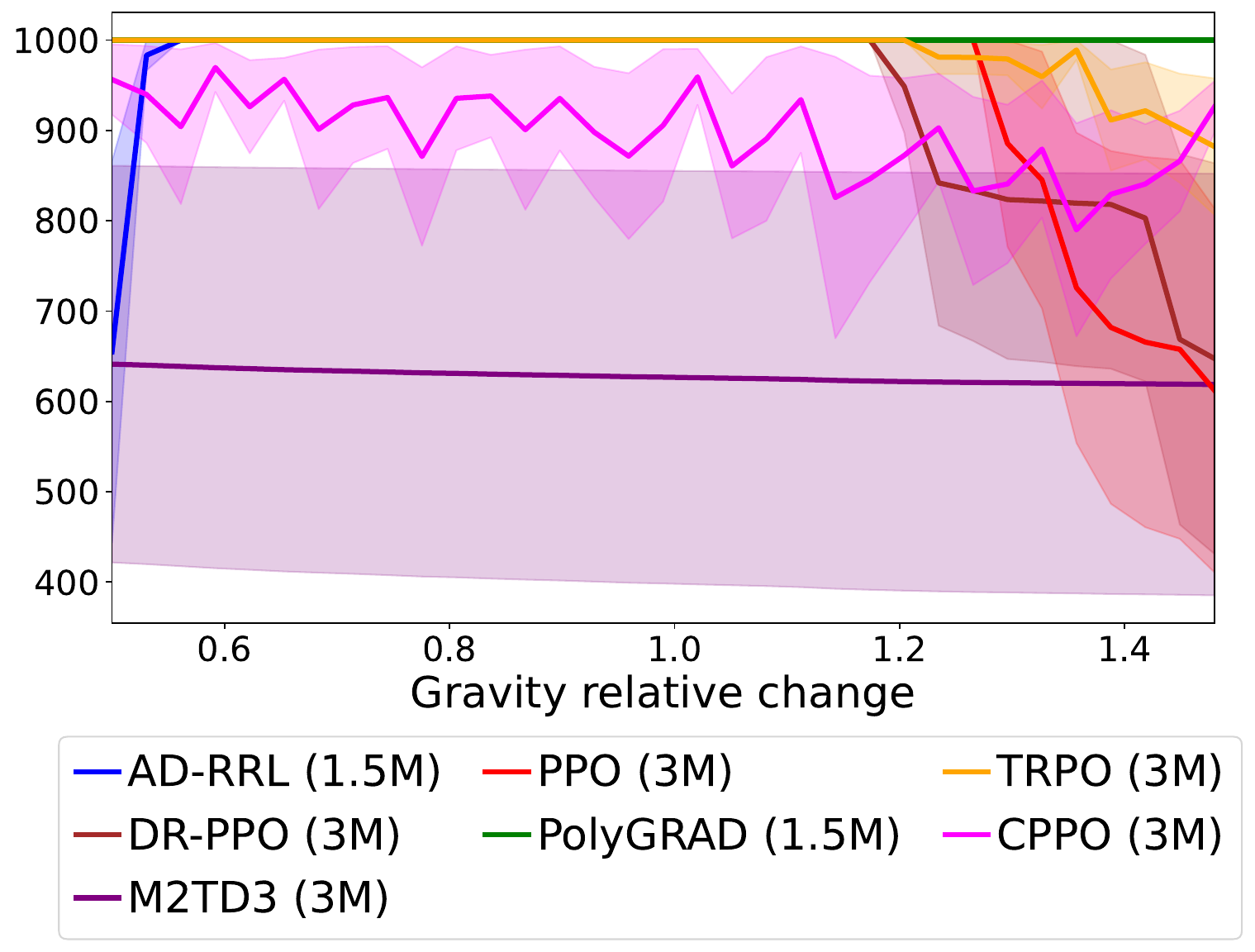}
    \subcaption{InvertedPendulum – Gravity}\label{fig:pendulum_gravity3m}
  \end{subfigure}

  \caption{Average return across variations in selected physics parameters. Shaded regions indicate $\pm$ one standard error over 5 runs. AD-RRL and PolyGRAD are trained on $1.5$M samples, while the remaining Model-Free Baselines are trained on $3$M samples.}
  \label{fig:robust_plots3M}
\end{figure}

In this section, we compare AD-RRL trained on $1.5$M environment samples with model-free baselines trained on $3$M samples. This setup allows model-free methods to compensate for their lower sample efficiency by leveraging their greater computational efficiency. In \cref{fig:training_curves3m} we present the learning curves on the nominal environment for the Model-Free algorithms. We can observe that AD-RRL still outperforms or (in the worst case) matches the performance of the Model-Free baselines in all of the environments. This is also reflected when testing the robustness of the algorithms, as shown in \cref{fig:robust_plots3M}. AD-RRL still achieves on par or higher returns than robust and non-robust Model-Free baselines. When comparing with \cref{fig:robust_plots}, we can see that some baselines perform worse than their $1.5$M samples versions (e.g., DR-PPO on the Walker and Hopper environments; TRPO on Hopper). This is expected, since an increased amount of samples might improve the performance on the nominal training environment, but will also lead to overfitting, making the agents more vulnerable to parameter changes in the test environment. The only exception appears to be M2TD3, which is able to outperform AD-RRL on the Walker environment at test time, when varying the physics parameters, despite achieving the same performance in the nominal environment. However, M2TD3 seems to be more prone to overfitting in easier environments, as seen in \cref{fig:training_pendulum3m}, where the baseline cannot reach the optimal solution but instead suffers a substantial performance decline in the latest training stages. Finally, M2TD3 remains stuck to a suboptimal policy in the Hopper environment, similarly to what we observed when training it on $1.5$M samples. The doubled amount of samples does not seem to benefit the learning process of this baseline. Overall, even when model-free baselines are given twice as many samples, AD-RRL remains on average better or on par across the proposed tasks.

\section{Limitations and future work}

\textbf{Computation time.} Guided diffusion requires dozens of reverse–diffusion steps for every synthetic trajectory and an extra gradient evaluation at each step. Consequently AD--RRL is slower than its precursor PolyGRAD, and both model–based methods need a higher training time (wall-clock) than the
model-free baselines, even though they are more sample efficient. Improving the training time for diffusion models (e.g., fine-tuning the network size or the number of denoising steps) sounds like a natural next step.

\textbf{Smooth–dynamics assumption.} Our derivation employs a Gaussian approximation and the computation of gradients $\nabla_{\!\tau}Z(\tau)$. Both of these presuppose reasonably smooth rewards and state transitions. Some tasks might break this assumption, causing inaccurate guidance. Extending adversarial diffusion to domains with non-smooth dynamics is left for future work.

\textbf{Scope of the evaluation.} Our evaluation focused on simulated control tasks. A natural next step is a Sim2Real study. That is, AD--RRL is trained entirely in simulation and then deployed on real hardware, measuring how much the adversarial-diffusion training
reduces the Sim2Real performance drop.

\section{Societal impact statement}

Our contribution is methodological: we propose a technique for making model-based RL more robust to dynamics misspecification. Robustness is typically beneficial—e.g., safer robot control or fewer failures in medical-decision support—yet any improvement in sample efficiency or policy quality can also lower the barrier to deploying RL in high-stakes settings. In domains such as healthcare, finance, or autonomous driving, deployment must therefore be accompanied by domain-specific safety checks, bias audits, and human oversight. Our work does not introduce new data‐collection practices, nor does it touch sensitive attributes, but it \emph{could} be combined with decision pipelines that do. We encourage future users of this method to evaluate downstream ethical, legal, and safety implications before real-world deployment.

%%%%%%%%%%%%%%%%%%%%%%%%%%%%%%%%%%%%%%%%%%%%%%%%%%%%%%%%%%%%

\newpage
\section*{NeurIPS Paper Checklist}
\begin{enumerate}

\item {\bf Claims}
    \item[] Question: Do the main claims made in the abstract and introduction accurately reflect the paper's contributions and scope?
    \item[] Answer: \answerYes{} % Replace by \answerYes{}, \answerNo{}, or \answerNA{}.
    \item[] Justification: We specified all the claims and assumptions in the abstract and introduction. The claims are backed up by lemmata, propositions and empirical results.

\item {\bf Limitations}
    \item[] Question: Does the paper discuss the limitations of the work performed by the authors?
    \item[] Answer: \answerYes{} % Replace by \answerYes{}, \answerNo{}, or \answerNA{}.
    \item[] Justification: We discuss the limitations of our algorithm in the Experiments and Conclusions and Future work sections.

\item {\bf Theory assumptions and proofs}
    \item[] Question: For each theoretical result, does the paper provide the full set of assumptions and a complete (and correct) proof?
    \item[] Answer: \answerYes{} % Replace by \answerYes{}, \answerNo{}, or \answerNA{}.
    \item[] Justification: We provide the full set of assumptions in the main body of lemmata and propositions, while the complete proofs are provided in the appendix (and referenced in the main body).

    \item {\bf Experimental result reproducibility}
    \item[] Question: Does the paper fully disclose all the information needed to reproduce the main experimental results of the paper to the extent that it affects the main claims and/or conclusions of the paper (regardless of whether the code and data are provided or not)?
    \item[] Answer: \answerYes{} % Replace by \answerYes{}, \answerNo{}, or \answerNA{}.
    \item[] Justification: In the appendix we provide the details regarding hyperparameters for our algorithm and baselines. We also provide pseudocode for our algorithm.

\item {\bf Open access to data and code}
    \item[] Question: Does the paper provide open access to the data and code, with sufficient instructions to faithfully reproduce the main experimental results, as described in supplemental material?
    \item[] Answer: \answerNA{} %\answerYes{} % Replace by \answerYes{}, \answerNo{}, or \answerNA{}.
    \item[] Justification: The code cannot be submitted in the workshop OpenReview form. %We provide the code with our submission, with the instructions to run it.

\item {\bf Experimental setting/details}
    \item[] Question: Does the paper specify all the training and test details (e.g., data splits, hyperparameters, how they were chosen, type of optimizer, etc.) necessary to understand the results?
    \item[] Answer: \answerYes{} % Replace by \answerYes{}, \answerNo{}, or \answerNA{}.
    \item[] Justification: We provide hyperparameters in the appendix section.

\item {\bf Experiment statistical significance}
    \item[] Question: Does the paper report error bars suitably and correctly defined or other appropriate information about the statistical significance of the experiments?
    \item[] Answer: \answerYes{} % Replace by \answerYes{}, \answerNo{}, or \answerNA{}.
    \item[] Justification: Result tables and plots always include the standard error across all the simulations.

\item {\bf Experiments compute resources}
    \item[] Question: For each experiment, does the paper provide sufficient information on the computer resources (type of compute workers, memory, time of execution) needed to reproduce the experiments?
    \item[] Answer: \answerYes{} % Replace by \answerYes{}, \answerNo{}, or \answerNA{}.
    \item[] Justification: In the appendix we provide detailed information about the computing resources used to perform the experiments.
    
\item {\bf Code of ethics}
    \item[] Question: Does the research conducted in the paper conform, in every respect, with the NeurIPS Code of Ethics \url{https://neurips.cc/public/EthicsGuidelines}?
    \item[] Answer: \answerYes{} % Replace by \answerYes{}, \answerNo{}, or \answerNA{}.
    \item[] Justification: We reviewed the NeurIPS Code of Ethics and we do not use sensitive data.

\item {\bf Broader impacts}
    \item[] Question: Does the paper discuss both potential positive societal impacts and negative societal impacts of the work performed?
    \item[] Answer: \answerYes{} % Replace by \answerYes{}, \answerNo{}, or \answerNA{}.
    \item[] Justification: We discuss potential negative societal impacts in the appendix.
    
\item {\bf Safeguards}
    \item[] Question: Does the paper describe safeguards that have been put in place for responsible release of data or models that have a high risk for misuse (e.g., pretrained language models, image generators, or scraped datasets)?
    \item[] Answer: \answerNA{} % Replace by \answerYes{}, \answerNo{}, or \answerNA{}.
    \item[] Justification: NA

\item {\bf Licenses for existing assets}
    \item[] Question: Are the creators or original owners of assets (e.g., code, data, models), used in the paper, properly credited and are the license and terms of use explicitly mentioned and properly respected?
    \item[] Answer: \answerYes{} % Replace by \answerYes{}, \answerNo{}, or \answerNA{}.
    \item[] Justification: We credit all the authors of software used in the paper (like the experiments baselines).

\item {\bf New assets}
    \item[] Question: Are new assets introduced in the paper well documented and is the documentation provided alongside the assets?
    \item[] Answer: \answerYes{} % Replace by \answerYes{}, \answerNo{}, or \answerNA{}.
    \item[] Justification: We provide details about our model (hyperparameters and pseudocode) and the code itself.

\item {\bf Crowdsourcing and research with human subjects}
    \item[] Question: For crowdsourcing experiments and research with human subjects, does the paper include the full text of instructions given to participants and screenshots, if applicable, as well as details about compensation (if any)? 
    \item[] Answer: \answerNA{} % Replace by \answerYes{}, \answerNo{}, or \answerNA{}.
    \item[] Justification: NA

\item {\bf Institutional review board (IRB) approvals or equivalent for research with human subjects}
    \item[] Question: Does the paper describe potential risks incurred by study participants, whether such risks were disclosed to the subjects, and whether Institutional Review Board (IRB) approvals (or an equivalent approval/review based on the requirements of your country or institution) were obtained?
    \item[] Answer: \answerNA{} % Replace by \answerYes{}, \answerNo{}, or \answerNA{}.
    \item[] Justification: NA

\item {\bf Declaration of LLM usage}
    \item[] Question: Does the paper describe the usage of LLMs if it is an important, original, or non-standard component of the core methods in this research? Note that if the LLM is used only for writing, editing, or formatting purposes and does not impact the core methodology, scientific rigorousness, or originality of the research, declaration is not required.
    %this research? 
    \item[] Answer: \answerNA{} % Replace by \answerYes{}, \answerNo{}, or \answerNA{}.
    \item[] Justification: NA

\end{enumerate}

%%%%%%%%%%%%%%%%%%%%%%%%%%%%%%%%%%%%%%%%%%%%%%%%%%%%%%%%%%%%

\end{document}